\documentclass[11pt]{article}

\usepackage{microtype}
\usepackage{graphicx}
\usepackage{subfigure} 
\usepackage{booktabs} 
\usepackage{natbib}
\usepackage{epsfig,epsf,fancybox}
\usepackage{amsmath}
\usepackage{mathrsfs}
\usepackage{amssymb}
\usepackage{color}
\usepackage{multirow}
\usepackage{paralist}
\usepackage{verbatim}
\usepackage{algorithm}
\usepackage{algorithmic}
\usepackage{galois}
\usepackage{boxedminipage}
\usepackage{accents}
\usepackage{stmaryrd}
\usepackage[bottom]{footmisc}
\usepackage{microtype}
\usepackage{graphicx}
\usepackage{subfigure}
\usepackage{booktabs} 
\usepackage{amsmath}
\usepackage{amssymb}
\usepackage{amsthm}

\usepackage{natbib}

\usepackage{url}
\usepackage[colorlinks, linkcolor={blue!70!gray}, citecolor={blue!70!gray}, urlcolor={blue!70!gray}]{hyperref}
\usepackage{enumitem}

\usepackage[normalem]{ulem}
\usepackage{xcolor}

\newcommand{\scrcmd}{\textsuperscript}
\newcommand{\scr}[1]{\scrcmd{#1}}

\def\R{{\mathbb{R}}}

\newcommand{\E}{\mathbb{E}}

\def\by{{\boldsymbol y}}
\def\bb{{\boldsymbol b}}
\def\ba{{\boldsymbol a}}
\def\bI{{\boldsymbol I}}
\def\bv{{\boldsymbol v}}
\def\bx{{\boldsymbol x}}
\def\bpi{{\boldsymbol \pi}}
\def\bdelta{{\boldsymbol \delta}}
\def\btheta{{\boldsymbol \theta}}
\def\mbf{{\boldsymbol f}}

\def\cN{{\mathcal N}}

\def\cC{{\mathcal C}}
\def\cT{{\mathcal T}}

\def\sR{{\textsf R}}
\def\AR{{\textsf{ARisk}}}
\def\AB{{\textsf{ABias}}}
\def\AV{{\textsf{AVar}}}

\newcommand{\<}{\langle}
\renewcommand{\>}{\rangle}
\newcommand{\ep}{\varepsilon}
\newcommand{\eps}{\varepsilon}

\newcommand{\htn}{\hat{\btheta}_n}

\newcommand{\hpi}{\hat{\pi}}
\newcommand{\ts}{\btheta_\star}
\newcommand{\Var}{\text{Var}}

\newtheorem{proposition}{Proposition}

\usepackage{subfigure} 
\usepackage{pifont}
\newcommand{\cmark}{\ding{51}}
\newcommand{\xmark}{\ding{55}}

\begin{document}


\begin{center}

{\bf{\LARGE{Understanding Generalization in Adversarial Training \\ \vspace{2mm}  via the Bias-Variance Decomposition}}}

\vspace*{.3in}
{\large{
\begin{tabular}{c}
Yaodong Yu$^{\dagger,\star}$ \,
Zitong Yang$^{\dagger,\star}$ \,
Edgar Dobriban$^\ddagger$ \,
Jacob Steinhardt$^{\diamond,\dagger}$ \,
Yi Ma$^{\dagger}$ \\
\end{tabular}
}}
\vspace*{.15in}

\begin{abstract}
Adversarially trained models exhibit a large generalization gap: they can interpolate the training set even for large perturbation radii, but at the cost of large test error on clean samples. To investigate this gap, we decompose the test risk into its bias and variance components and study their behavior as a function of adversarial training perturbation radii ($\varepsilon$). We find that the bias increases monotonically with $\varepsilon$ and is the dominant term in the risk. Meanwhile, the variance is unimodal as a function of $\varepsilon$, peaking near the interpolation threshold for the training set. This characteristic behavior occurs robustly across different datasets and also for other robust training procedures such as randomized smoothing. It thus provides a test for proposed explanations of the generalization gap. We find that some existing explanations fail this test--for instance, by predicting a monotonically increasing variance curve. This underscores the power of bias-variance decompositions in modern settings–by providing two measurements instead of one, they can rule out more explanations than test accuracy alone. We also show that bias and variance can provide useful guidance for scalably reducing the generalization gap, highlighting pre-training and unlabeled data as promising routes.
\end{abstract}

\let\thefootnote\relax\footnotetext{\hspace*{-6.0mm}$^\star$\,Yaodong Yu and Zitong Yang contributed equally to this work.}
\let\thefootnote\relax\footnotetext{\hspace*{-6.0mm}$^\dagger$\,Department of Electrical Engineering and Computer Sciences, University of California, Berkeley.}
\let\thefootnote\relax\footnotetext{\hspace*{-6.0mm}$^\diamond$\,Department of Statistics, University of California, Berkeley.}
\let\thefootnote\relax\footnotetext{\hspace*{-6.0mm}$^\ddagger$\,Department of Statistics, University of Pennsylvania.}
\let\thefootnote\relax\footnotetext{\hspace*{-4.0mm}Email: \texttt{yyu@eecs.berkeley.edu}, \texttt{zitong@berkeley.edu}, \texttt{dobriban@wharton.upenn.edu}, \texttt{jsteinhardt@berkeley.edu}, \texttt{yima@eecs.berkeley.edu}.}
\end{center}

\section{Introduction}\label{sec:intro}
\begin{figure*}[t]
  \begin{center}
    \includegraphics[width=.43\textwidth]{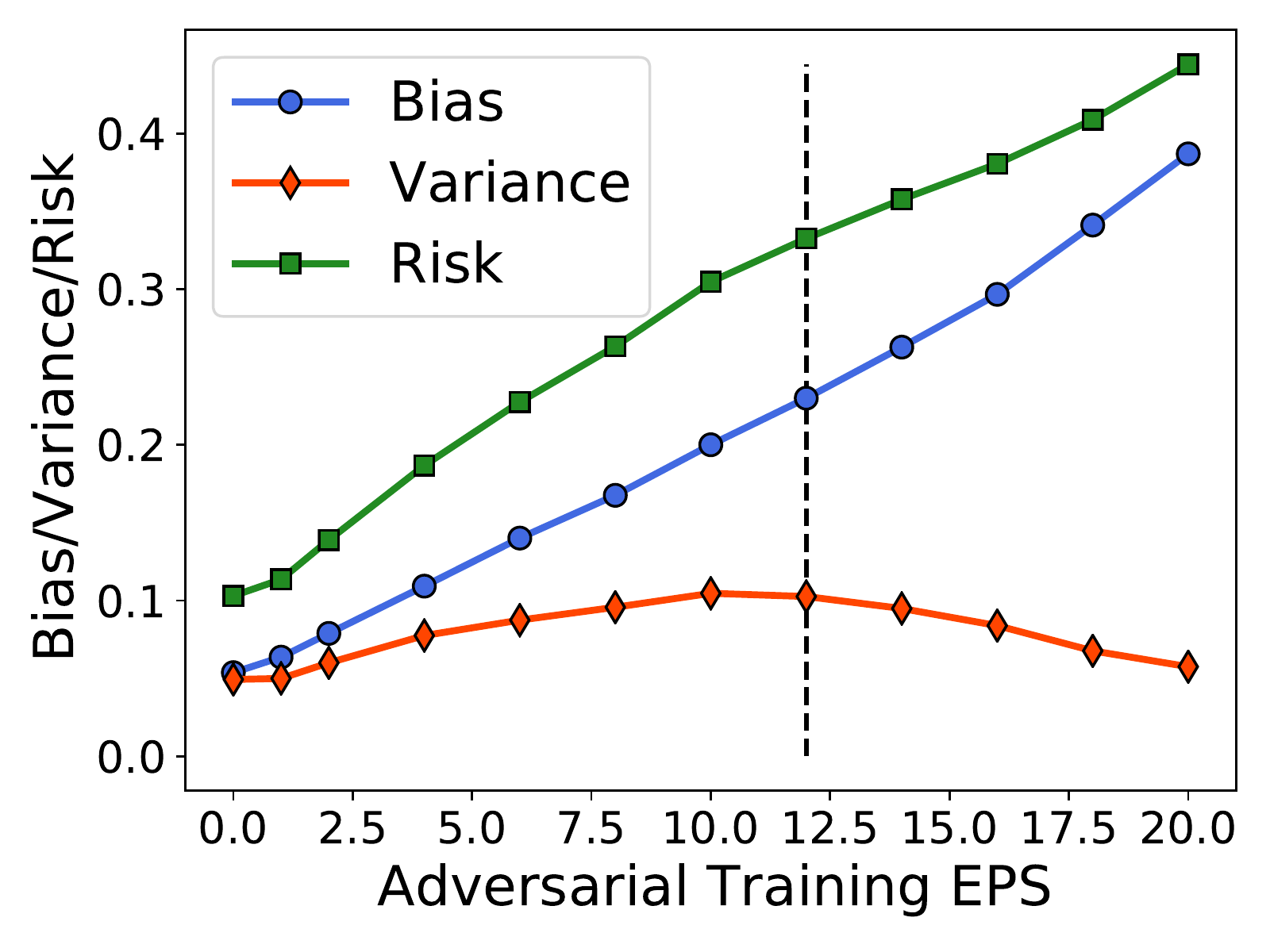}
    \includegraphics[width=.43\textwidth]{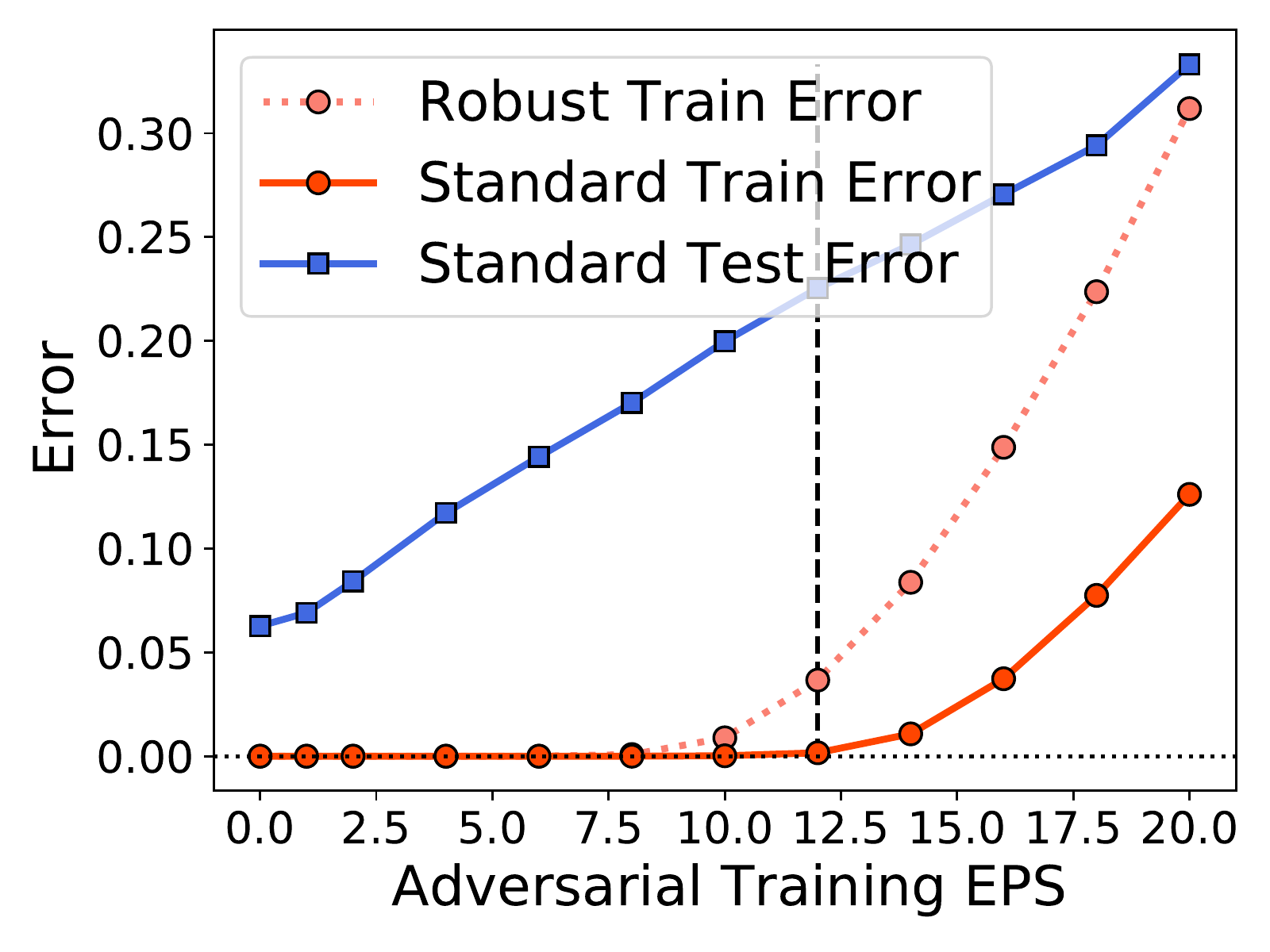}
    \vskip -0.15in
    \caption{\footnotesize{Measuring the performance for $\ell_{\infty}$-adversarial training (with increasing perturbation size) on the CIFAR10 dataset.
    \textbf{Standard error} means the error rate on clean samples, and \textbf{robust error} means the error rate on adversarially perturbed samples. 
    The \textit{vertical dashed line} corresponds to the robust training error of the adversarially trained model reaching $2\%$ (i.e., robust interpolation threshold).
    (\textbf{Left}) Evaluating the bias, variance, and risk for the $\ell_{\infty}$-adversarially trained model (WideResNet-28-10).
    (\textbf{Right}) Evaluating robust training error, and standard training/test error on the same model.} 
    }
  \label{fig:bvr-mainline}
  \end{center}
  \vskip -0.35in
\end{figure*}

Adversarial training enhances the robustness of modern machine learning methods at the cost of decreased accuracy on the clean test samples~\citep{goodfellow2014explaining, madry2017towards, sinha2017certifying}. Though the model can fit the training data perfectly in adversarial training, the generalization error on clean test data increases compared with non-adversarially trained models. 
As illustrated in the right panel of Figure \ref{fig:bvr-mainline}, adversarially trained models on CIFAR10 can achieve nearly zero robust training error, yet their test error still grows large as the perturbation radius $\ep$ increases, even for clean test examples. 
To improve the robustness and accuracy of adversarially trained models, we thus seek to 
understand the cause for this increased ``generalization gap'' between errors on the training and test datasets.

To better understand this generalization gap, we turn to a standard tool of statistical learning theory, the bias-variance decomposition~\citep{markov,lehmann1983theory,casella2021statistical,hastie2009elements, geman1992neural}. A large bias means the model predicts poorly on average, while a large variance means the predictions are very spread out. Adversarial training can be thought of as a 
form of regularization, and so we might expect it to increase bias and decrease variance. On the 
other hand, adversarial training intuitively makes the decision boundary more jagged and complex~\citep{yang2020robustness}, suggesting increased variance is the main cause. 
To arbitrate these different perspectives, we empirically investigate bias and variance 
across a variety of datasets as a function of the adversarial training radius $\ep$.

Interestingly, we find that both perspectives are incomplete--the variance is neither increasing 
nor decreasing with $\ep$. Instead, we robustly observe the following behavior across several 
datasets. First, the bias is \emph{monotonically increasing} in $\ep$ and is the dominant term of the risk. 
Second, the variance is \emph{unimodal} in $\ep$: The variance increases up to some peak and then 
decreases. Moreover, the peak consistently occurs near the 
\emph{robust interpolation threshold}--the minimum $\ep$ for which the model can no longer robustly interpolate the training set.

The unimodal variance observation is surprising given that it contradicts both of the perspectives above, and its co-occurence with the interpolation threshold suggests a deeper phenomenon. 
We believe these observations are an inversion of the ``double descent'' phenomenon where larger 
models sometimes generalize better \citep{Belkin15849, Nakkiran2020Deep}, and in particular echo observations in \citet{YYBV2020}, which finds the variance of a (regularly-trained) neural network to be unimodal as a function of width, with the peak also occurring at the interpolation threshold. We discuss this in more detail in Section \ref{sec:discussion}.

Beyond their intrinsic interest, we draw two additional implications from the observed characteristic properties of bias and variance. 
First, since they can be robustly observed across many adversarial learning tasks, they can be used as criteria to check the validity of simplified 
conceptual models of adversarial training. 
If a simplified model fails to predict the correct behavior of bias and variance, then it has limitations for understanding adversarial training. We apply this test to several such 
models and find the following:
\begin{itemize}[topsep=3pt,parsep=1pt,partopsep=0.0pt]
    \item The \textbf{non-linearity} of the predictor is not generally important: the monotonic bias and unimodal variance properties already (approximately) hold for logistic regression.
    \item The \textbf{adaptivity} of the attack is not important: the properties hold for randomized smoothing, which adds independent Gaussian noise to each minibatch.
    \item The \textbf{dynamic nature} of the attack \emph{is} important: if we add \emph{fixed} Gaussian noise to each minibatch (rather than new noise at each epoch) the properties cease to hold.
    \item The \textbf{high dimensionality} of the data is important: the properties fail to hold on certain key low-dimensional models, while being observable in the same models in high dimensions.
\end{itemize}
This last point in particular casts doubt on popular intuitions regarding the generalization gap in adversarial training, which are based on the "jaggedness" of the decision boundary in low dimensions (Section~\ref{sec:toy_2d}). 
The observations above also shed light into the mechanism underlying the adversarial generalization gap: it is primarily a property of dynamic training noise in high-dimensional settings.

As a second implication, since the variance is dominated by the bias and decreases for large 
$\eps$, scalable solutions to adversarial robustness \emph{must primarily address the bias}. We therefore 
investigate several algorithms for improving adversarial robustness, to determine which ones decrease the bias. 
The common approach of early stopping~\citep{rice2020overfitting} decreases variance, but slightly increases bias, so we cannot solely rely on it. Increasing the width also reduces variance, but not bias. However, we find that adding unlabeled data~\citep{carmon2019unlabeled, uesato2019labels, najafi2019robustness, zhai2019adversarially} and using pre-trained models~\citep{hendrycks2019using} both decrease the bias, suggesting these may lead to scalable opportunities to achieve further robustness.

\subsection{Related work}
\paragraph{Robustness-accuracy tradeoff.} There is a large literature on understanding the trade-off between adversarial robustness and standard accuracy for adversarially robust classification problems. 
Classical works study robustness of linear models, showing equivalences between adversarially robust regression and lasso~\citep{xu2009robust, xu2009robustness}. 
Recent works identify the fundamental robustness-accuracy trade-off in simplified theoretical models, where samples from different classes are close and there is no robust and accurate classifier~\citep{fawzi2018analysis, zhang2019theoretically, dobriban2020provable}. 
\citet{nakkiran2019adversarial} argue theoretically that the tradeoff is due to the simplicity of the classifiers, suggesting that more complex classifiers are required for adversarial robustness. 
\citet{tsipras2018robustness} and \citet{ilyas2019adversarial} argue that adversarially robust models are less accurate because they cannot rely on predictive, yet non-robust, features of the data. 
Other works characterize the robust error and trade-offs for linear models~\citep{dan2020sharp, javanmard2020precise-COLT, javanmard2020precise, dobriban2020provable,megyeri2019adversarial}. \citet{raghunathan2020understanding} argue that unlabeled data can be used to mitigate the trade-off and improve model robustness. \citet{yang2020adversarial} empirically study the trade-off via local Lipschitzness. Other works study robustness for non-parametric methods~\citep{wang2018analyzing,bhattacharjee2020non,yang2020robustness}.

\vspace{-0.12in}
\paragraph{Adversarially robust generalization.} \citet{schmidt2018adversarially} consider a Gaussian classification model and show that adversarial training requires more data for generalization.  
Several previous works study the generalization and sample complexity via Rademacher complexity and VC dimension~\citep{cullina2018pac, attias2019improved, yin2019rademacher, khim2018adversarial, montasser2019vc}. 
\citet{bubeck2018adversarial, bubeck2019adversarial} argue that learning robust classifiers may be computationally hard. 
Other works study statistical properties of adversarial logistic regression~\citep{javanmard2020precise, dan2020sharp, dobriban2020provable, taheri2020asymptotic}. Several works use the concentration of measure to study adversarial examples in high dimensions \citep{gilmer2018trb:arxiv:v3,shafahi2018are,mahloujifar2019empirically}. Similarly, some works develop connections to optimal transport \citep{dohmatob2019limitations,bhagoji2019lower,pydi2019adversarial}.

\vspace*{-0.12in}
\paragraph{Adversarial defenses and attacks.} A large body of literature is devoted to improving adversarial robustness~\citep{goodfellow2014explaining, kurakin2016adversarial, madry2017towards, hein2017formal, wong2017provable, sinha2017certifying, cohen2019certified, raghunathan2018certified, zhang2019theoretically, wu2020adversarial} and developing adversarial attacks~\citep{carlini2017towards, papernot2017practical, athalye2018obfuscated, kang2019testing, aboutalebi2020vulnerability, croce2020reliable}.  
Incorporating unlabeled data has been shown to be an effective approach for improving model adversarial robustness~\citep{carmon2019unlabeled, uesato2019labels, najafi2019robustness, zhai2019adversarially}.  
\citet{shah2020pitfalls} argue that the ``Simplicity Bias'' in deep neural networks could be an explanation for the existence of universal adversarial examples. 
\citet{gowal2020uncovering} show that by tuning the model size, activation function, and model weight averaging, adversarial training can achieve state-of-the-art $\ell_{\infty}$/$\ell_{2}$ adversarial robustness on the CIFAR10 dataset. \citet{croce2020robustbench} recently benchmark adversarial $\ell_{\infty}$/$\ell_{2}$ robustness based on AutoAttack~\citep{croce2020reliable}.

\section{Preliminary}\label{sec:preliminary}
We review the bias-variance decomposition  that will be used in later sections. We mainly consider the decomposition of the squared $\ell_{2}$ loss; extensions to the cross-entropy loss can be found in Appendices~\ref{sec:appendix-bv-prelim} and~\ref{sec:appendix-additional-exp-results}.

\vspace{-0.12in}
\paragraph{Standard training and adversarial training.} For standard supervised $k$-class classification, we are given $n$ i.i.d. training samples $\cT = \{(\bx_i, \by_i)\}_{i=1}^n$, 
a parameterized  class of prediction functions $f_{\btheta}(\cdot)$---where $f_{\btheta}(\cdot), \by_i \in \mathbb{R}^{k}$--- and a loss function $\ell(\cdot, \cdot)$ .
The goal of \textit{standard training}  is to find parameters $\widehat{\btheta}(\cT)$ that approximately solve the following problem:
\vspace*{-0.5em}
\begin{equation}\label{eq:standard-training}
\min_{\btheta}  \frac{1}{n} \sum_{i=1}^n \ell\left(f_\btheta(\bx_i), \by_i \right).
\end{equation}
Certain models learned through Eq.~\eqref{eq:standard-training} are vulnerable to adversarial perturbations, especially some deep neural nets~\citep{szegedy2013intriguing,biggio2013evasion}.
\textit{Adversarial training}~(AT) enhances the robustness of the model by solving the following  robust optimization problem, for a set of perturbations $\Delta$: 
\vspace{-0.05in}
\begin{equation}\label{eq:adv-training-formulation}
\min_{\btheta} \frac{1}{n} \sum_{i=1}^n \max_{\bdelta_i \in \Delta}  \ell\left(f_\btheta(\bx_i + \bdelta_i), \by_i \right).
\end{equation}
In practice, the inner maximization is approximately solved by projected gradient descent (PGD), and the outer minimization is approximately solved by stochastic gradient descent (SGD)~\citep{madry2017towards}. 
We follow this adversarial training procedure in this paper. A typical choice for the perturbation set is the $\ell_{p}$ norm ball $\Delta = \{\bdelta : \|\bdelta\|_{p} \leq \varepsilon\}$.

\vspace{-0.1in}
\paragraph{Bias-variance decomposition.} Given a test point $(\bx, \by)$, an accurately trained model satisfies $\by\approx f_{\widehat{\btheta}}(\bx)$. Therefore, for a good model, the prediction $f_{\widehat{\btheta}}(\bx)$ should not depend too much on the training set $\cT$.
This is quantified by the variance:
\begin{equation}\label{eq:variance_def}
    \textsf{Var} = \E_\bx \Var_{\cT} \left[ f_{\widehat{\btheta}(\cT)}(\bx) \right],
\end{equation}
which captures the variation in the prediction under different initialization of the training set. 
The variance depends on the training algorithm $\smash{\cT\to \widehat\btheta(\cT)}$ (and thus implicitly also on the sample size) and the data distribution.
A large variance implies that the trained model $f_{\widehat{\btheta}}$ is sensitive to a particular realization of the training dataset $\cT$, which is unfavorable. 
The bias term is
\[
    \textsf{Bias} = \E_{\widehat{\btheta}, \bx, \by}\left\|\by- f_{\htn(\cT)}(\bx) \right\|_2^2 - \textsf{Var} =\E_{\bx, \by}\left\|\by- \E_{\widehat{\btheta}}f_{\htn(\cT)}(\bx) \right\|_2^2,
\]
which captures how good the model performs on average.

\vspace{-0.12in}
\paragraph{Estimation of bias and variance.} To estimate the bias and variance of the models, we first partition the training dataset $\mathcal{T}$ into two disjoint parts $\mathcal{T} = \mathcal{T}_{1} \cup \mathcal{T}_{2}$\footnote{Note that this changes the training sample size from $n$ to $n/2$.}.  
Then we rely on the bias-variance decomposition for the mean squared error $\|\by - f_{\widehat{\btheta}}(\bx)\|^{2}$ (see previous paragraphs), where $\by$ is the one-hot encoding label vector and $f_{\widehat{\btheta}}(\bx)$ is the output of the softmax layer. 
More details for estimating the bias and the variance can be found in Appendix~\ref{sec:appendix-bv-prelim}.

\section{Measuring Bias and Variance for Deep Neural Networks}\label{sec:bv-analysis-dl}
We study the bias-variance behavior of deep neural networks adversarially trained on the CIFAR10, CIFAR100, and ImageNet10 (a subset of 10 classes from ImageNet~\citep{deng2009imagenet}) dataset. We find that \textit{the variance is unimodal and the bias is monotonically increasing}~(Figure~\ref{fig:bvr-mainline} and \ref{fig:bvr-cifar-imagenet}) as a function of perturbation size. We find that the same phenomenon holds for randomized smoothing (another 
common robust training method), but not for training on noisy Gaussian-perturbed data (where the perturbations are fixed during training). Our code is available at \url{https://github.com/yaodongyu/BiasVariance-AdversarialTraining}.

\subsection{Adversarially Trained Models}\label{sec:adv-training-on-2d-cifar10}
We first study the bias-variance decomposition for $\ell_{\infty}$ adversarially trained models with increasing perturbation size $\varepsilon$ on image classification tasks.

\begin{figure*}[t]
\begin{center}
\subfigure[CIFAR100.]{
\includegraphics[width=.31\textwidth]{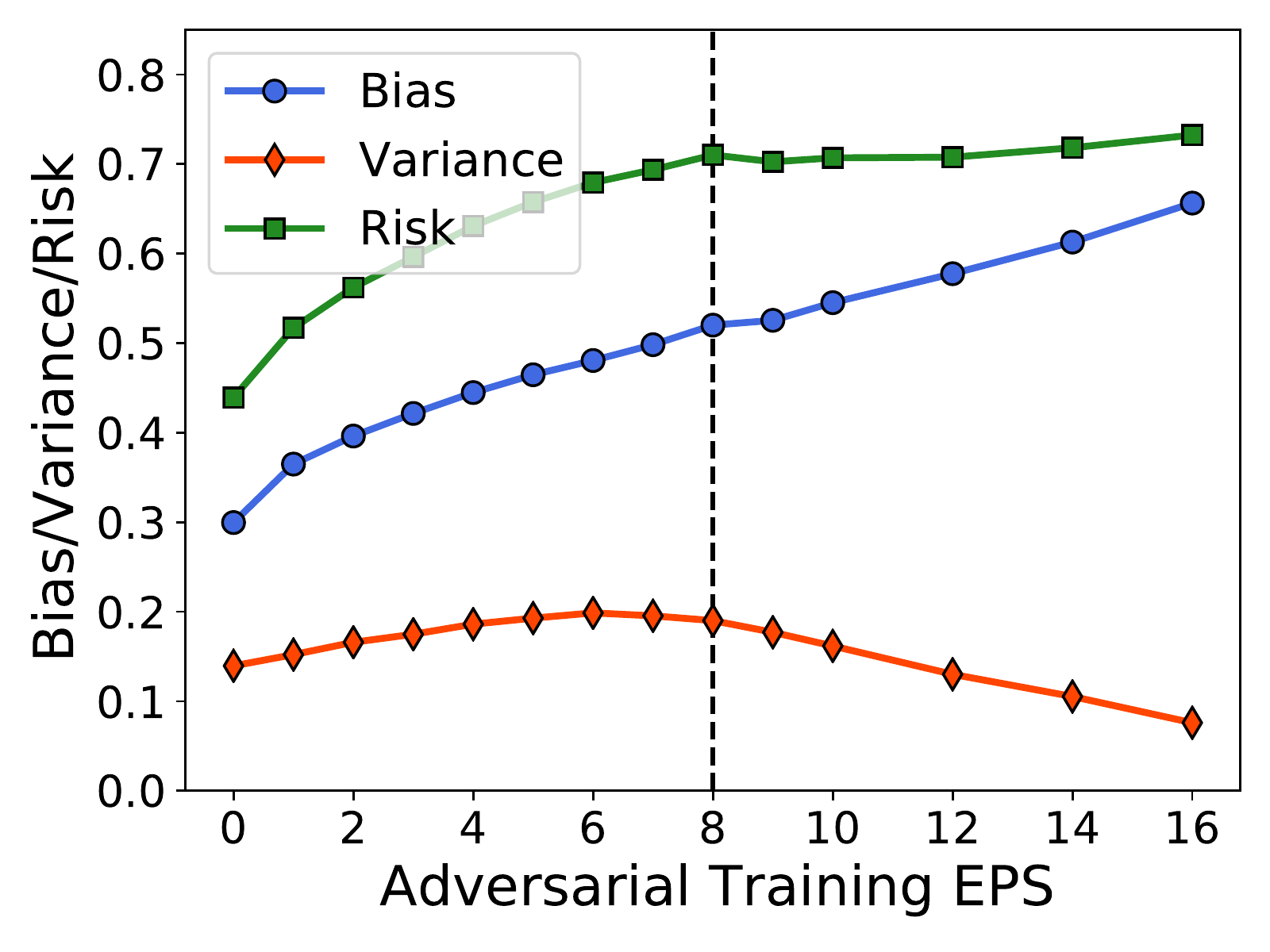}
\label{fig:bvr-cifar100-linf-main-text}
}
\subfigure[ImageNet10.]{
\includegraphics[width=.31\textwidth]{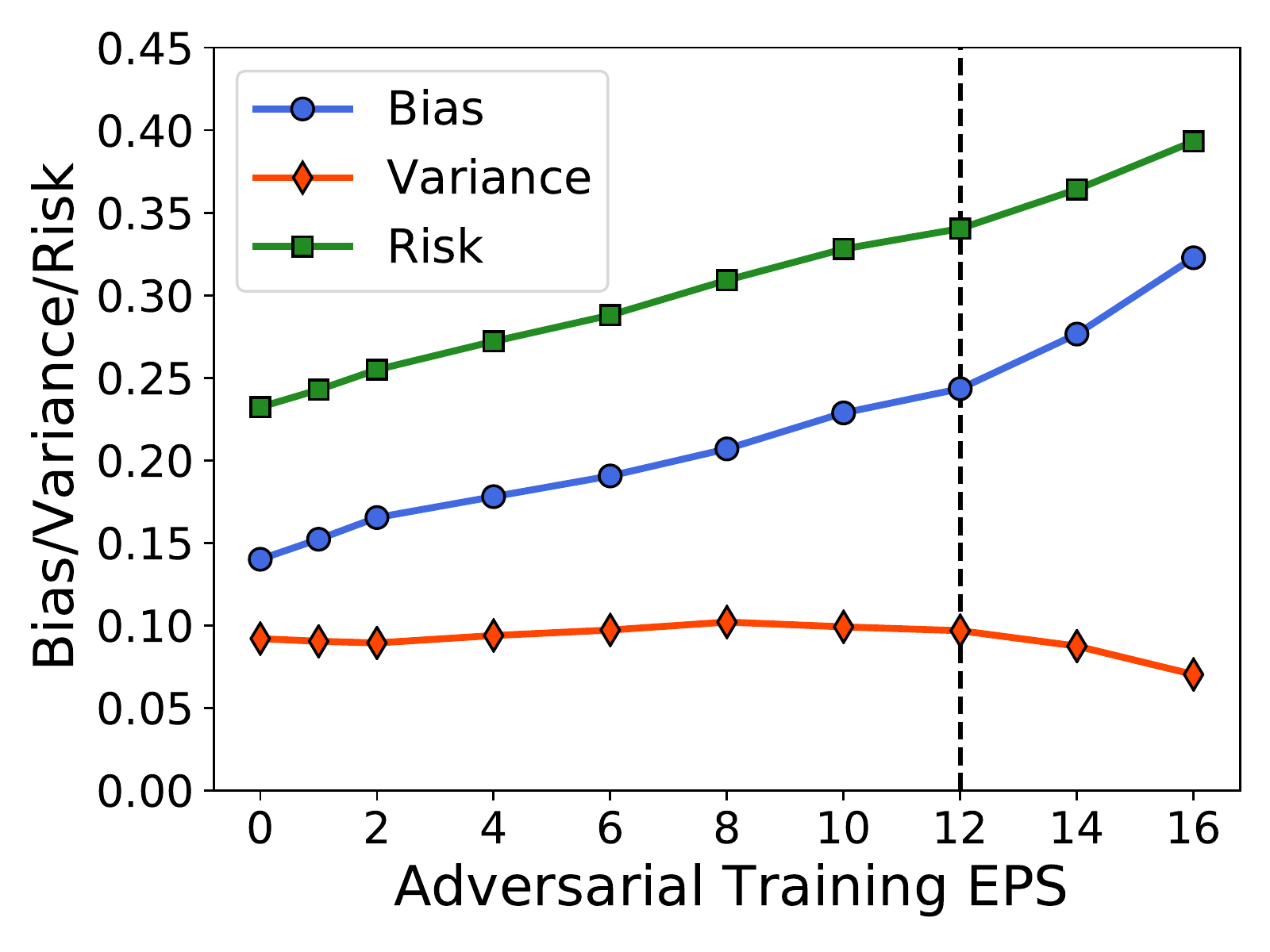}
\label{fig:bvr-imagenet10-linf-main-text}
}
\subfigure[CIFAR10 (\textit{early stopping}).]{
\includegraphics[width=.31\textwidth]{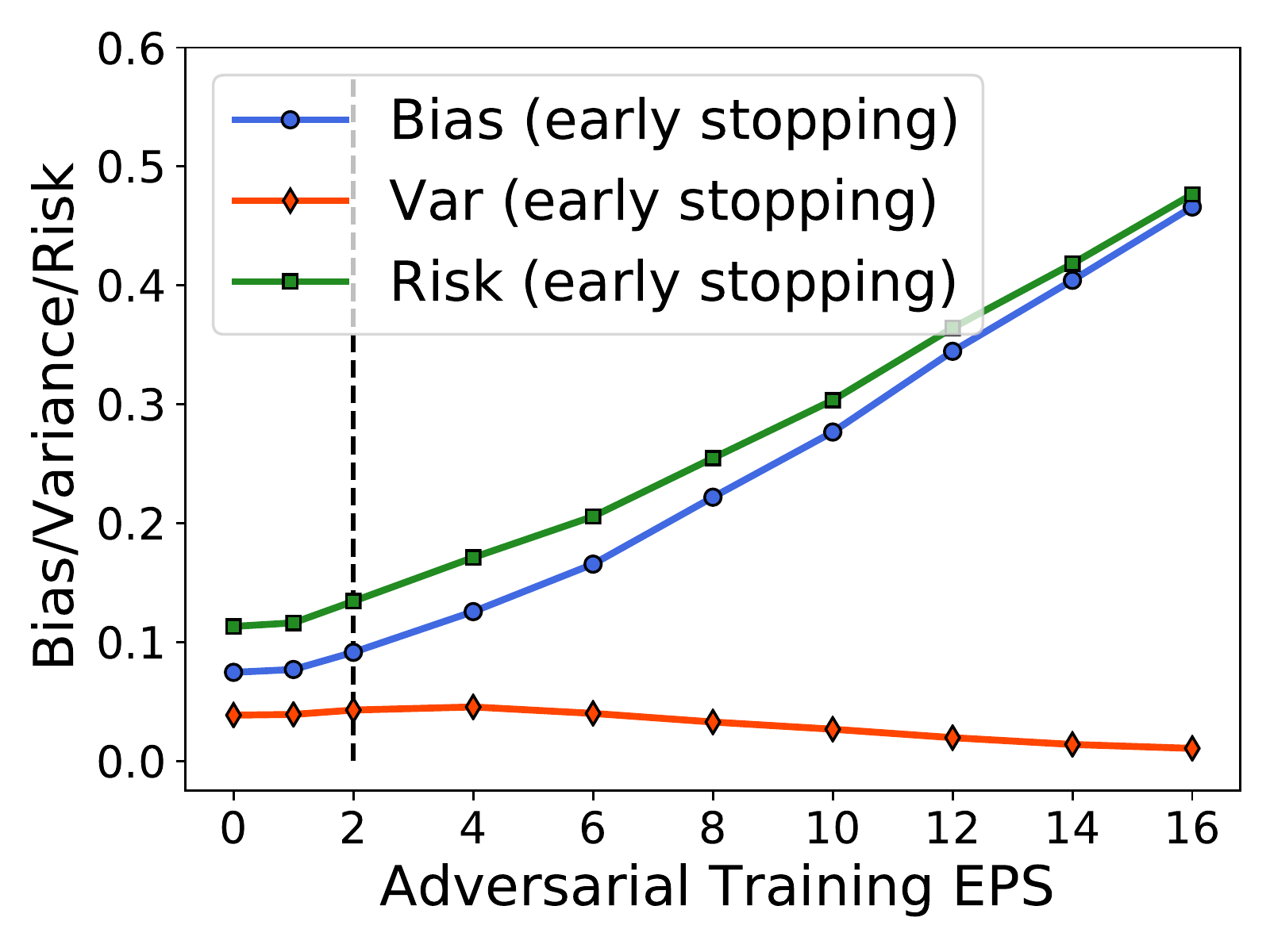}
\label{fig:adv-bvr-compare-earlystopping-main-text}
}
\vskip -0.1in
\caption{Measuring bias-variance for $\ell_{\infty}$-adversarial training (with increasing perturbation size) on the CIFAR10\,/\,CIFAR100\,/\,ImageNet10 dataset.
The {\textit{vertical dashed line}} corresponds to the robust training error of the adversarially trained model being larger than $2\%$ (i.e., robust interpolation threshold).
(\textbf{a}) CIFAR100 dataset. (\textbf{b}) ImageNet10 dataset. (\textbf{c}) CIFAR10 dataset (evaluated on early stopped models).}
\label{fig:bvr-cifar-imagenet}
\end{center}
\vskip -0.3in
\end{figure*}

\vspace*{-0.75em}
\paragraph{Experimental setup.} We train our models using stochastic gradient descent (SGD) with momentum 0.9, where the learning rate is 0.1, mini-batch size is {128}, and the weight decay parameter is 0.0005. We train the models for 200 epochs and apply stage-wise learning rate decay during training, where we decay the learning rate by a factor of 0.1 at epochs 100 and 150. We use the Wide-ResNet architecture (WRN-28-10)~\citep{zagoruyko2016wide} on CIFAR10~\citep{krizhevsky2009learning}, PreActResNet18~\citep{he2016identity} on CIFAR100, and ResNet18~\citep{he2016deep} on ImageNet10.

We consider a perturbation size $\varepsilon/255$ with $\varepsilon$ varying from $0$ to $16$ on CIFAR100 and ImageNet10, and use larger $\varepsilon$ on CIFAR10 ($0\leq\varepsilon\leq 20$). For the inner maximization in adversarial training, we apply a 10-step PGD attack with perturbation step size $\eta = 0.25\cdot(\varepsilon/255)$. 
In addition to models trained to full convergence, we also study models that are stopped early, at the 100-th epoch. Early stopping has been shown to improve adversarial robustness~\citep{zhang2019theoretically, rice2020overfitting, gowal2020uncovering}. More detailed descriptions of the experimental setup can be found in Appendix~\ref{sec:appendix-additional-exp-results}.

\vspace*{-0.75em}
\paragraph{Bias-variance behavior.} From Figure~\ref{fig:bvr-mainline}~(CIFAR10), Figure~\ref{fig:bvr-cifar100-linf-main-text}~(CIFAR100), and Figure~\ref{fig:bvr-imagenet10-linf-main-text}~(ImageNet10), we observe several consistent properties of adversarial training:
\begin{enumerate}
    \item[P1.] The bias increases monotonically when we increase the perturbation size $\varepsilon$.
    \item[P2.] The variance first increases, then decreases, with $\varepsilon$.
    \item[P3.] The variance peak occurs near the point where the robust training error first starts to rise. We call this point the \textbf{robust interpolation threshold} and quantify it as the smallest $\varepsilon$ with larger than $2\%$ robust training error.
    \item[P4.] The bias is the dominant term in the risk.
\end{enumerate}

Properties P1-P3 are reminiscent of previous findings related to double descent~\citep{YYBV2020,lin2020causes, Belkin15849, Nakkiran2020Deep}.
There, the bias and variance exhibit similar behavior (for \emph{standard, non-robust} training) when increasing another parameter, model width. 
Increasing $\varepsilon$ makes the data more complex and decreasing width make the model less powerful. Both make the learning task harder to interpolate and lead to a bias/varaince behavior that satisfy properties P1-P3. The connection here suggests a more general  relation between the bias/variance behavior and the ``hardness of interpolation''.

\begin{table}[t]
\begin{center}
\caption{Summary of bias-variance behavior of different training methods on various datasets. P1-P4 are 4 consistent properties of adversarial training (see Section~\ref{sec:adv-training-on-2d-cifar10} for detailed descriptions).  ({\color{green!50!gray}\large\cmark}) -- property is satisfied, ({\color{red!50!gray}\large\xmark}) -- property is not satisfied, and (${\approx}$) -- property is approximately satisfied.}
\begin{tabular}{llccccc}
\toprule
TRAINING SETTING & DATASET & P1 & P2 & P3 & P4 & REFERENCE\\
\midrule
Adversarial Training (AT) & Image & {\color{green!50!gray}\large\cmark} & {\color{green!50!gray}\large\cmark} & {\color{green!50!gray}\large\cmark} & {\color{green!50!gray}\large\cmark} & Section \ref{sec:adv-training-on-2d-cifar10}\\
Early Stopping-AT & Image & {\color{green!50!gray}\large\cmark} & {\color{green!50!gray}\large\cmark} & {\color{green!50!gray}\large\cmark} & {\color{green!50!gray}\large\cmark} & Section \ref{sec:adv-training-on-2d-cifar10}\\
\midrule
Randomized Smoothing & Image  & {\color{green!50!gray}\large\cmark} & {\color{green!50!gray}\large\cmark} & {\color{green!50!gray}\large\cmark} & {\color{green!50!gray}\large\cmark} & Section \ref{sec:training-with-gaussian-noise}\\
(\textit{Fixed-})Gaussian Noise & Image  & {\color{green!50!gray}\large\cmark} & {\color{red!50!gray}\large\xmark} & {\color{red!50!gray}\large\xmark} & {\color{green!50!gray}\large\cmark} & Section \ref{sec:training-with-gaussian-noise}\\
\midrule
Adversarial Training & Box Dataset ($d \leq 5$) & {\color{green!50!gray}\large\cmark} & {\color{red!50!gray}\large\xmark} & {\color{red!50!gray}\large\xmark} & {\color{green!50!gray}\large\cmark}& Section \ref{sec:toy_2d}\\
Adversarial Training & Box Dataset ($d \geq 10$) & {\color{green!50!gray}\large\cmark} & {\color{green!50!gray}\large\cmark} & {\color{green!50!gray}\large\cmark} & {\color{green!50!gray}\large\cmark} & Section \ref{sec:toy_2d}\\
\midrule
Logistic Regression & Mixture of Gaussian & $\approx$ & {\color{green!50!gray}\large\cmark} & {\color{green!50!gray}\large\cmark} & {\color{green!50!gray}\large\cmark} & Section \ref{sec:toy_logistic}\\
Logistic Regression & Planted Robust Feature & $\approx$ & {\color{green!50!gray}\large\cmark} & {\color{green!50!gray}\large\cmark} & {\color{green!50!gray}\large\cmark}& Section \ref{sec:toy_logistic}\\
\bottomrule
\end{tabular}
\label{table:robust-check}
\end{center}
\vskip -0.1in
\end{table}

\vspace*{-0.75em}
\paragraph{Early stopping.} Figure~\ref{fig:adv-bvr-compare-earlystopping-main-text} depicts the results of early stopping on CIFAR10. Compared with the fully trained model in Figure~\ref{fig:bvr-mainline}, the peak of the variance curve occurs earlier, but still coincides with the robust interpolation threshold. Indeed, each of properties P1-P4 still hold in this setting.

The early stopped model has lower risk than the fully trained model when $\varepsilon$ is small. However, the risk is lower due to decreased variance (the bias in fact increases slightly). Since the variance is already low (per P4), decreasing it further has limited effect. We therefore also need approaches for 
reducing the bias--we will study this further in Section~\ref{sec:performance-analysis}.

\vspace{-0.12in}
\paragraph{Robustness checks.} Beyond the settings above, we observe properties P1-P4 for 
$\ell_2$-adversarial training and for a different bias-variance decomposition based on cross-entropy (Appendix~\ref{sec:appendix-additional-exp-results}).
We summarize the bias-variance behavior of various training methods in the first two rows of Table~\ref{table:robust-check}. 
In addition to standard adversarial training on the image classification tasks considered, we find that properties P1-P4 consistently  hold in several other settings.

\subsection{Training with Gaussian Noise}\label{sec:training-with-gaussian-noise}
In this subsection, we study the bias-variance decomposition for two other training approaches: randomized smoothing and training on Gaussian-perturbed data.
Randomized smoothing adds random Gaussian noise to each mini-batch used for the SGD updates, and has been used to obtain certifiably robust models~\citep{lecuyer2019certified, cohen2019certified}. 
In contrast, training on Gaussian-perturbed data adds Gaussian noise to training images \emph{once} at the beginning of training, which can be viewed as smoothing of the distribution of the training dataset. 
For both methods, we perturb the training images with random Gaussian noise with variance $\sigma^{2} \in [0.0, 1.0]$ and clip pixels to $[0.0, 1.0]$.
We discuss full details of the experimental setup in Appendix~\ref{sec:gauss-noise-setup}.

\begin{figure*}[t]
\begin{center}
\subfigure[Randomized smoothing training.]{
\includegraphics[width=.237\textwidth]{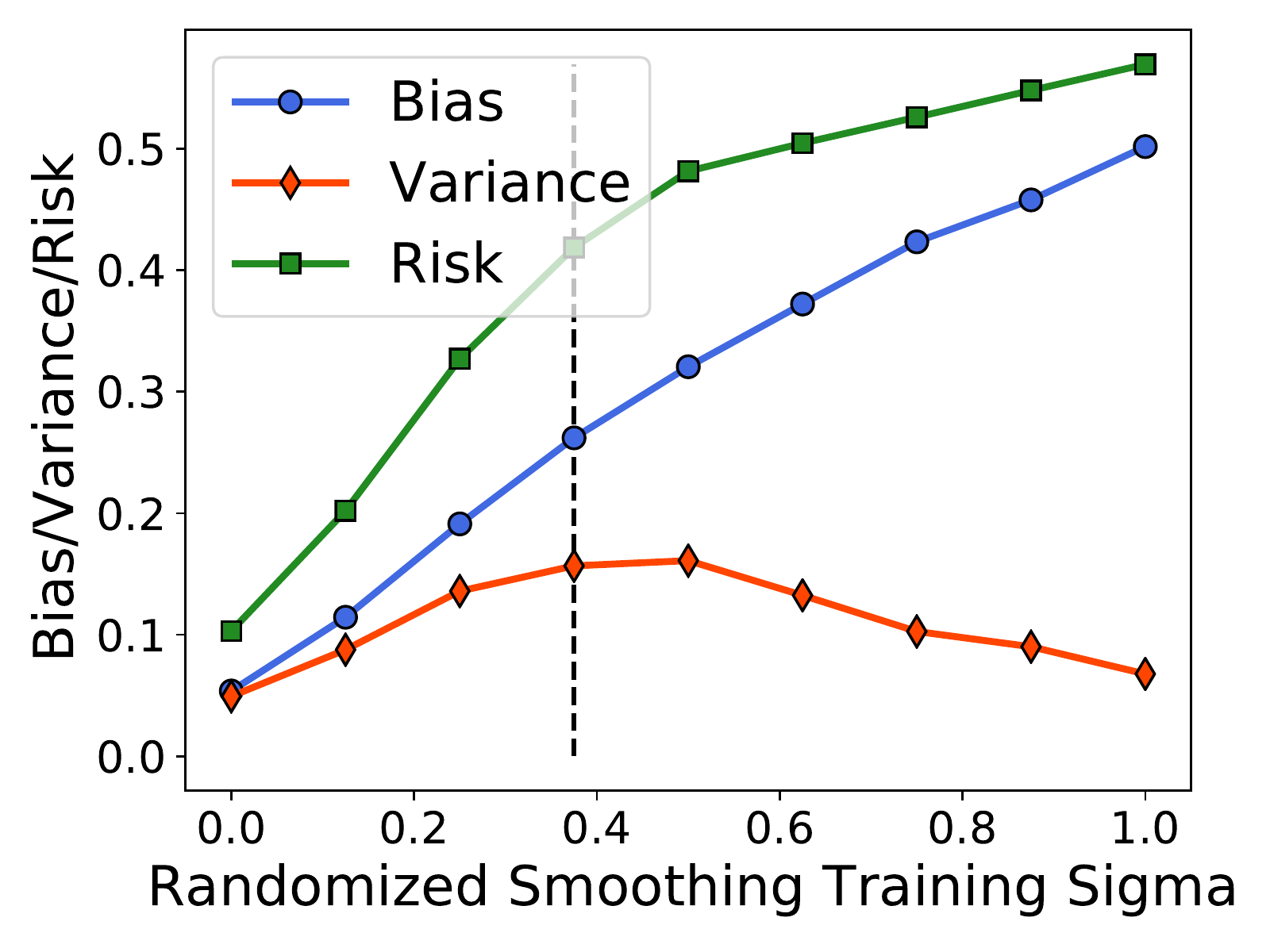}
\includegraphics[width=.237\textwidth]{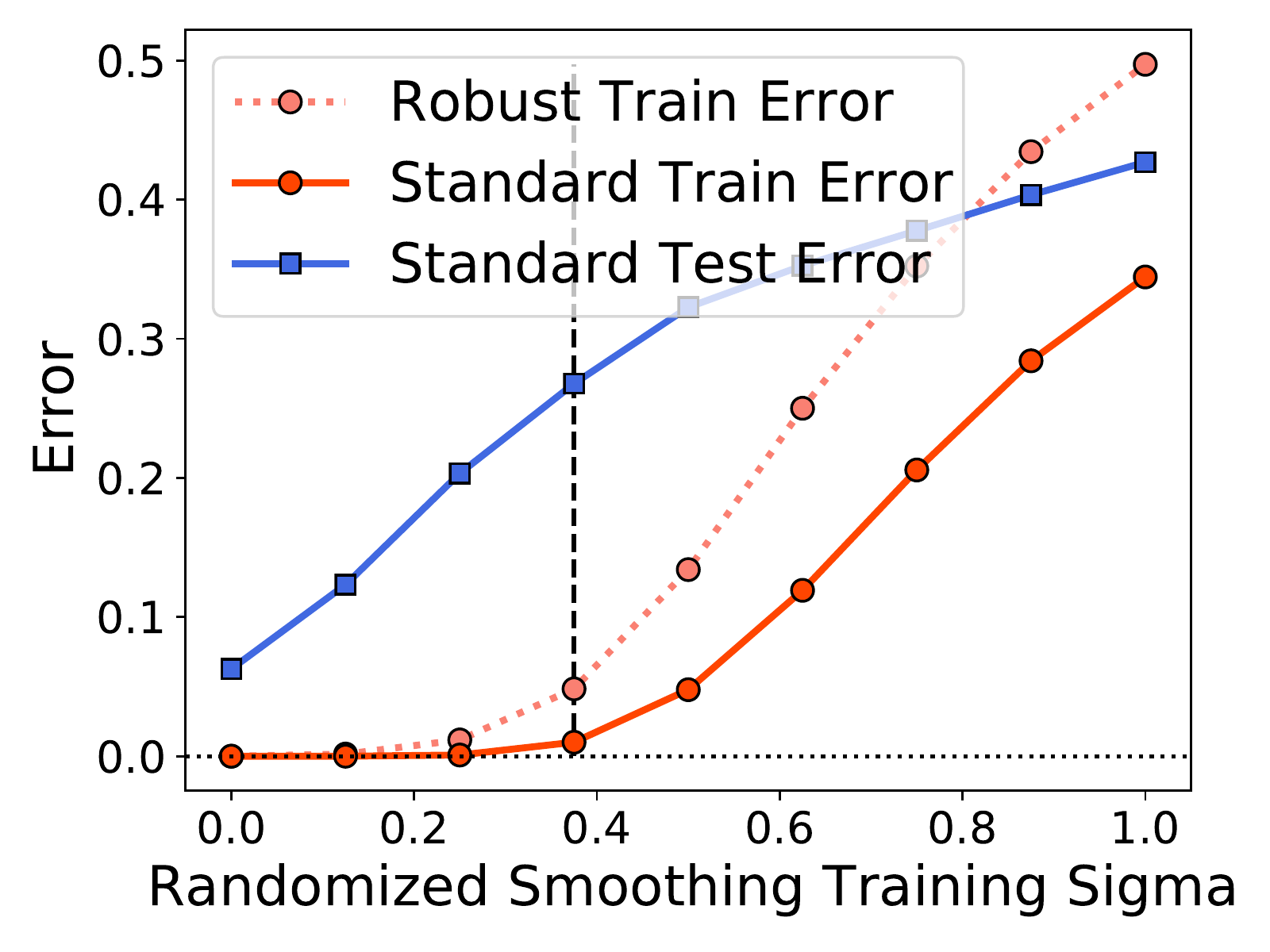}
\label{fig:rs-bvr}
}
\subfigure[Training on Gaussian-perturbed data.]{
\includegraphics[width=.237\textwidth]{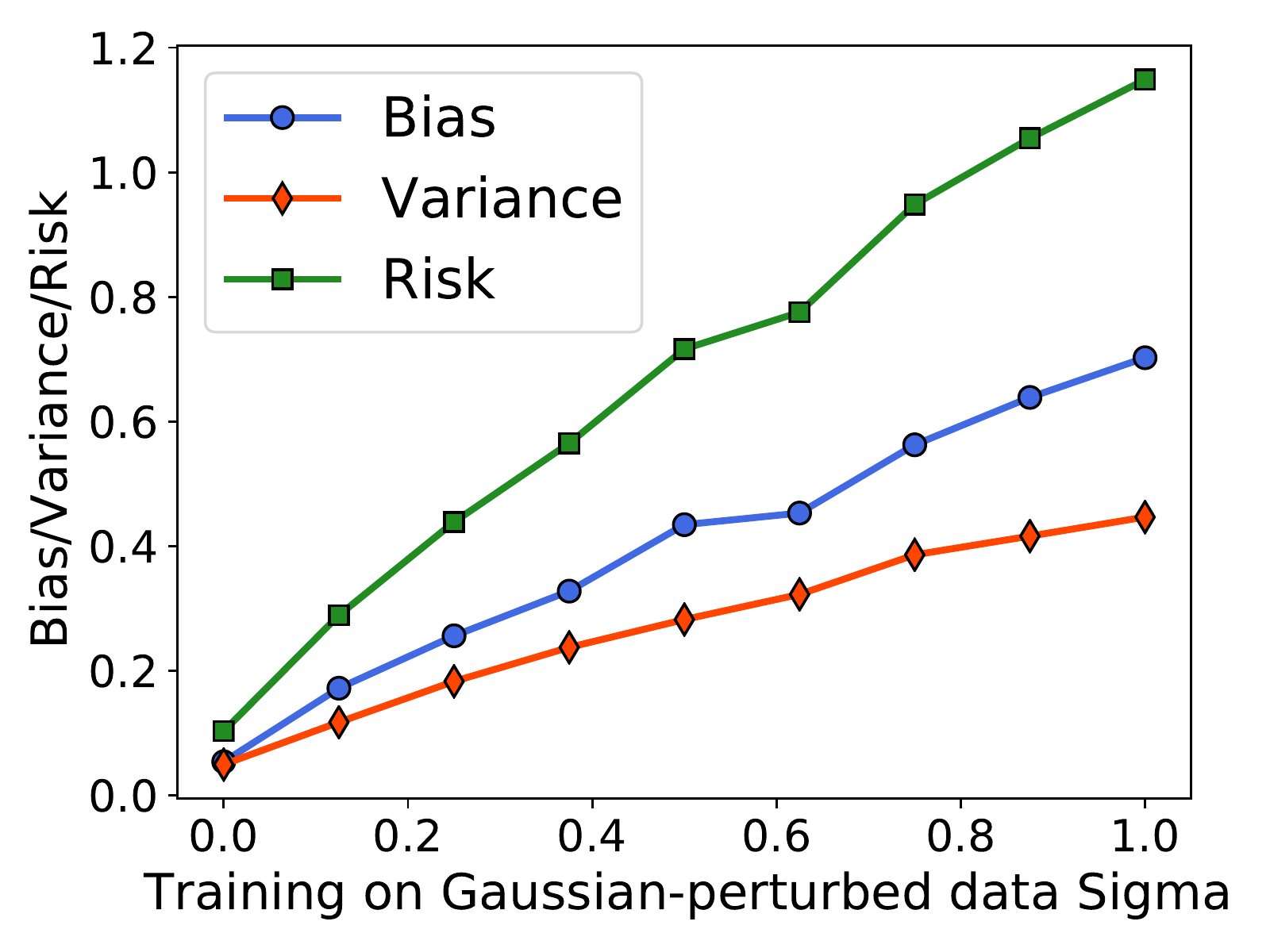}
\includegraphics[width=.237\textwidth]{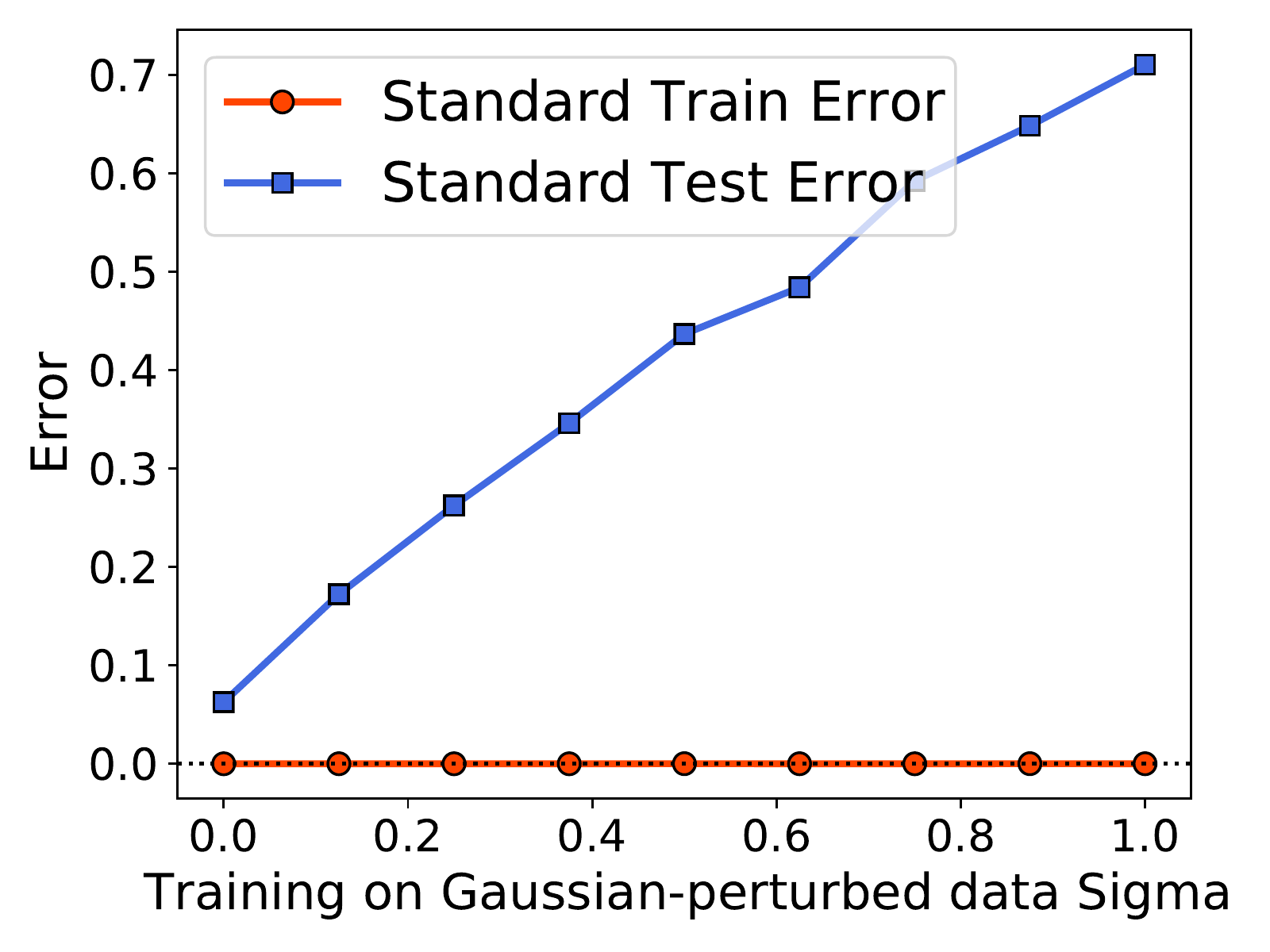}
\label{fig:perturbonce-bvr}
}
\vskip -0.1in
\caption{Measuring bias/variance/risk and train/test error for \textit{randomized smoothing training} and \textit{training on Gaussian-perturbed data} on the CIFAR10 dataset using WRN-28-10, varying $\sigma^{2}$. 
(\textbf{a}) Results for randomized smoothing training, where the dashed line indicates the robust interpolation threshold. 
(\textbf{b}) Results for training on Gaussian-perturbed data.}
\label{fig:Gaussian-bvr}
\end{center}
\vskip -0.25in
\end{figure*}

\vspace*{-0.75em}
\paragraph{Comparison to adversarial training.} 
As shown in Figure~\ref{fig:rs-bvr}, we find that {models trained with randomized smoothing  behave similarly to adversarially trained models}: properties P1-P4 continue to hold. 
However, for Gaussian-perturbed data the story is different. In Figure~\ref{fig:perturbonce-bvr}, we see that \textit{both the bias and the variance are monotonically increasing} as $\sigma^{2}$ increases. Thus, while P1 (monotonic bias) holds, properties P2 and P3 (unimodal variance) do not.
This suggests that,  in comparison with randomized smoothing, training on Gaussian-perturbed data may not be a good proxy for understanding adversarial training. We summarize the result in the second two rows of Table \ref{table:robust-check}.

\section{Testing Conceptual Models for Adversarial Training}\label{sec:testing_toy}
In the previous section, we saw that Gaussian-perturbed data fails to produce the bias-variance properties P1-P3 and thus is not a good model for understanding adversarial training. 
In this section, we take this idea further, by applying the same test to several other models. 
We find that we can observe much of the same qualitative behavior already for a linear model adversarially trained on a mixture of Gaussians, as for an adversarially trained neural net on image data. 
On the other hand, high dimensionality is important--in low dimensions neither of the variance properties (P2 or P3) hold, which casts doubt on some common intuitions around the adversarial generalization gap.

\subsection{2D Box Example}\label{sec:toy_2d}
\begin{figure*}[t]
  \begin{center}
    \subfigure[\label{fig:adv-bvr-2d-example-decision-boundary}Visualization of decision boundaries in 2D.]{
    \includegraphics[width=.2\textwidth]{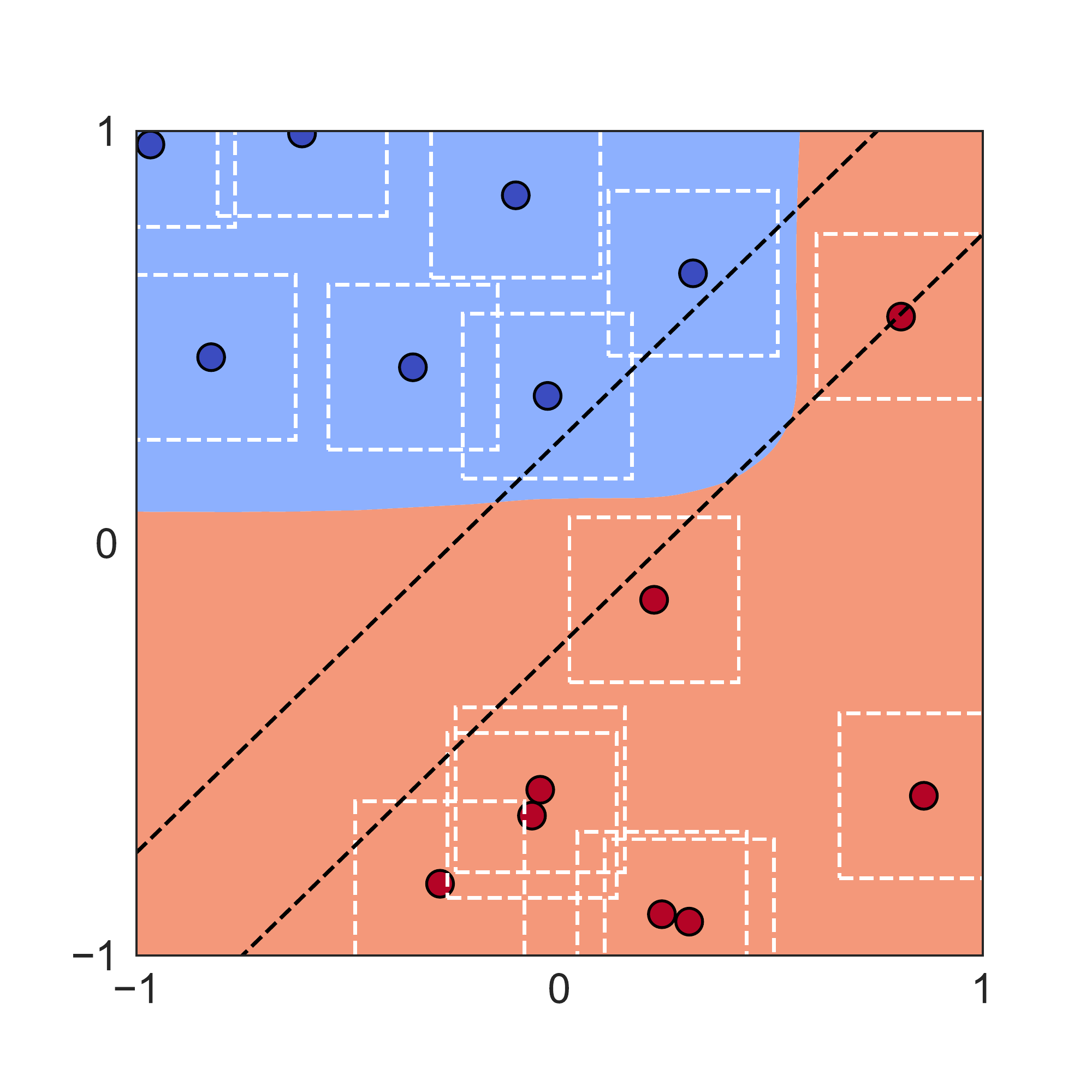}
    \includegraphics[width=.2\textwidth]{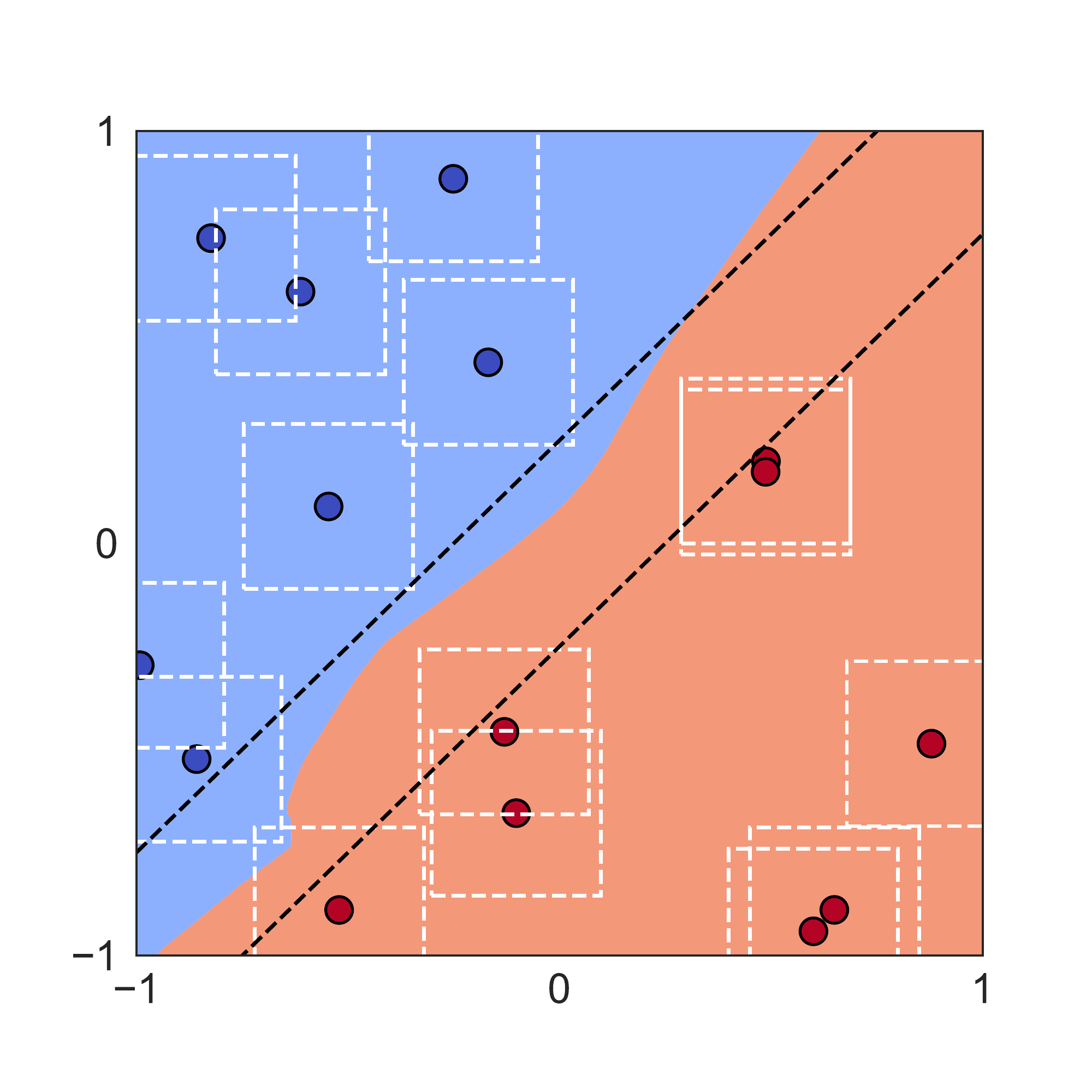}
    \label{fig:2d_vis}
    }
    \subfigure[\label{fig:adv-bvr-mainline-2d}Bias/variance/risk (\textit{box example}, $d=2, 20$).]{
    \includegraphics[width=.27\textwidth]{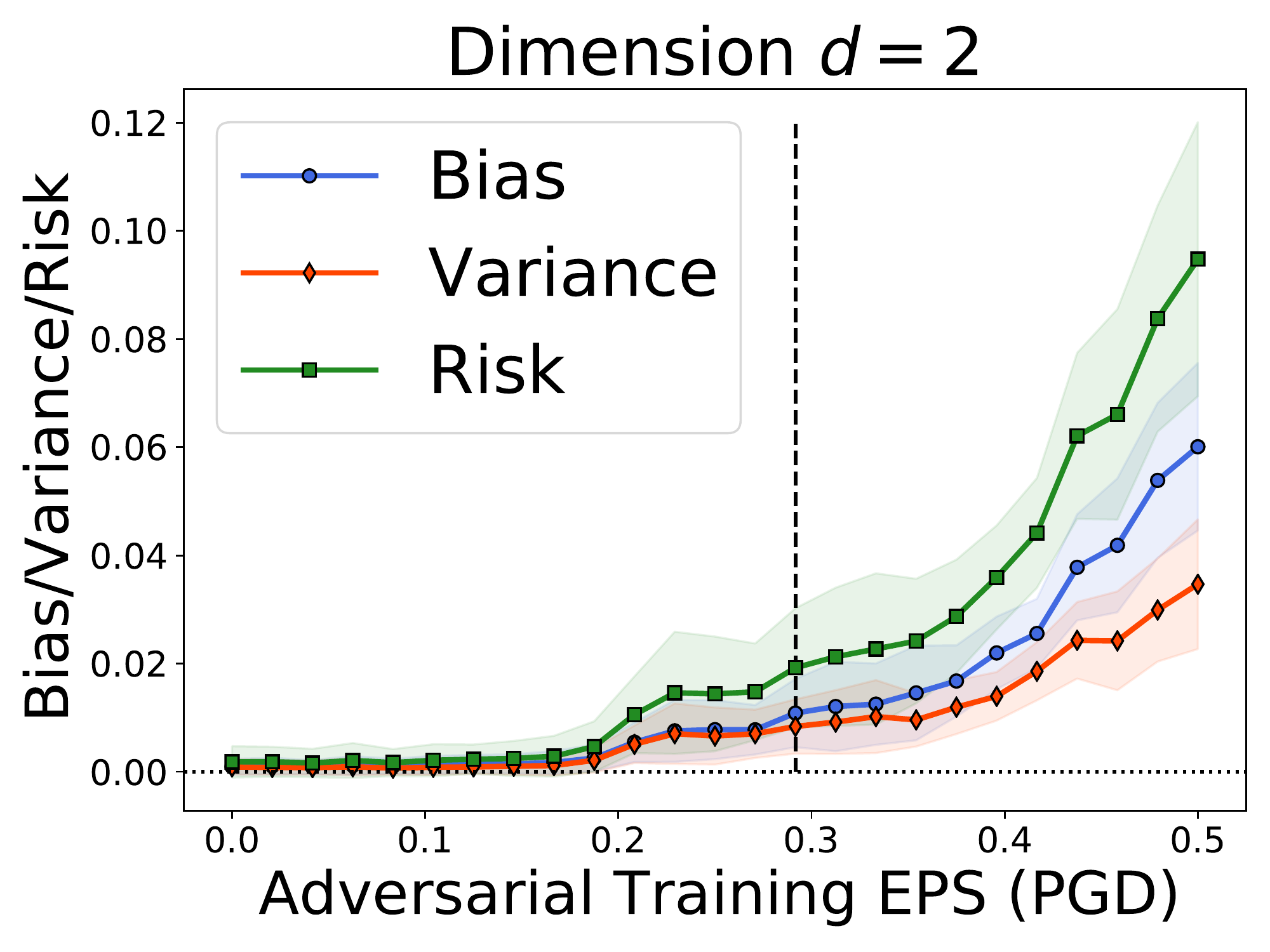}
    \includegraphics[width=.27\textwidth]{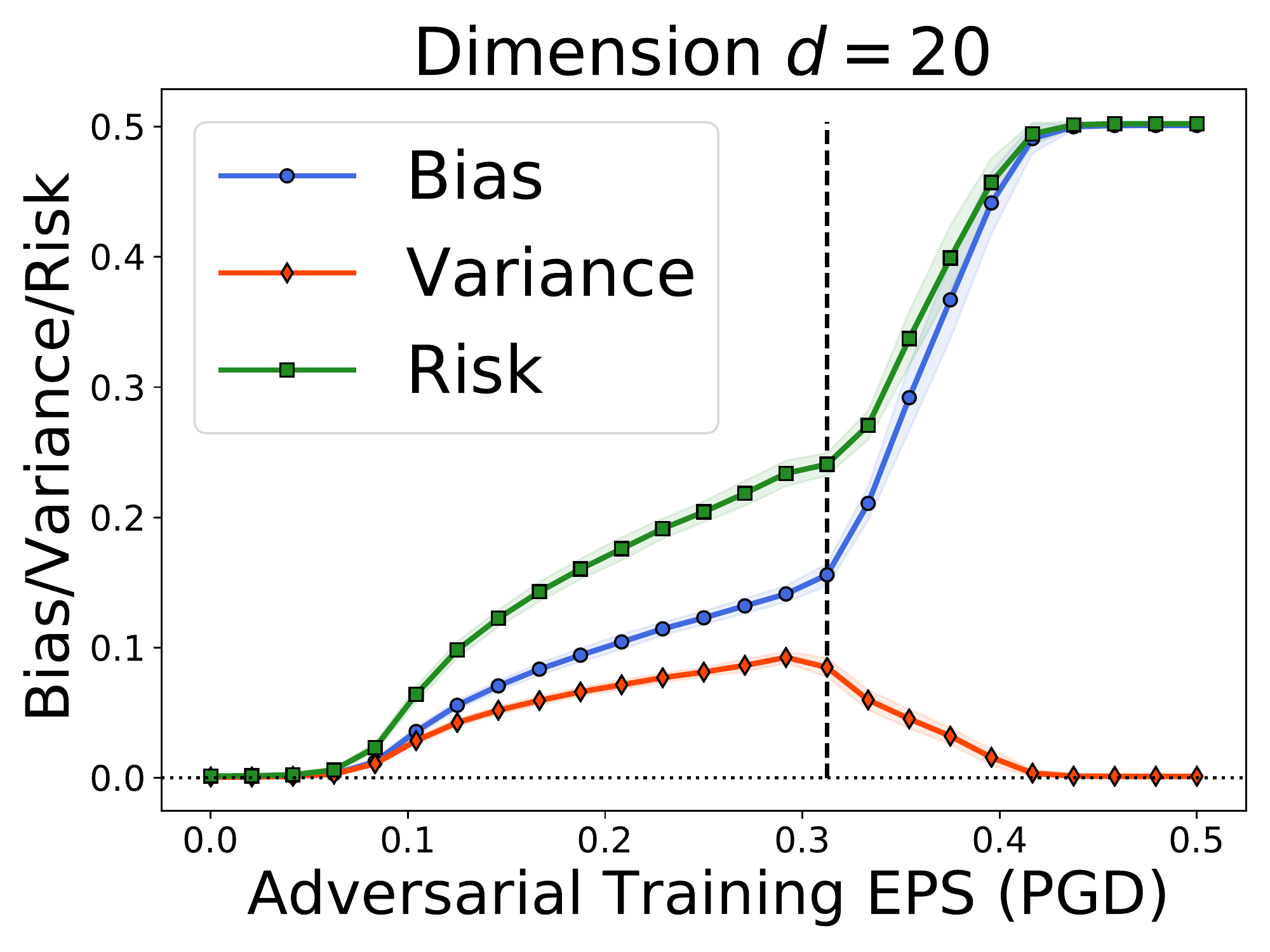}
    }
    \vskip -0.1in
    \caption{\textbf{(a)} Visualization of decision boundaries of $\ell_{\infty}$ adversarially trained models on \textit{2D box example}. The training datasets are randomly sampled from the same data distribution.  \textbf{(b)} Evaluating the bias, variance, and risk for the $\ell_{\infty}$-adversarial training (with increasing perturbation size) on the \textit{box} dataset with dimension $d=2$ (\textbf{left}) and $d=20$ (\textbf{right}), and the dashed line indicates the robust interpolation threshold. 
    }
  \end{center}
  \vskip -0.3in
\end{figure*}

In this subsection, we study the bias-variance decomposition for the ``2D box example''. 
As illustrated in the synthetic binary classification problem of Figure \ref{fig:2d_vis}, ``2D box example'' visualizes the adversary by adding small $\ell_\infty$ boxes around each training sample. Adversarial training then strives to find a decision boundary avoiding the boxes. This model explains the robustness-accuracy gap via the jagged decision boundary. 
This intuition is commonly presented in the literature \citep{madry2017towards, wong2017provable, zhang2019theoretically, yang2020adversarial} as an important "mental picture" to understand adversarial training. 
As we will see, this explanation is limited, and such a low dimensional model fails to capture the complexities of adversarial training.

\vspace{-0.12in}
\paragraph{2D box example experimental setup.}  To formally study this 2D box model, we consider a synthetic binary classification problem on $[-1, 1]^2$, where the input $\bx$ is uniformly sampled from the region 
$\{\bx\,: \, |x_1 - x_2|\geq \gamma/\sqrt{2}\},$
and the label for $\bx$ is 
$y=\text{sign}(x_1-x_2)$. Here $\gamma \in(0, 1)$ is a margin parameter that controls the distance between two classes. 
We set the margin $\gamma=1/4$, the number of training sample $n=20$, and the number of test samples as $10,000$. 
We perform $\ell_{\infty}$ adversarial training using a fully connected network (three hidden layers with 100 nodes each). 
We vary the adversarial perturbation $\varepsilon$ from $0.0$ to $0.5$, we set the perturbation step size for 10-step-PGD as $\eta = \varepsilon \cdot 0.4$. We use the Adam~\citep{kingma2014adam} optimizer for 2,000 epochs with a learning rate of 0.001.

\vspace{-0.12in}
\paragraph{Adversarial training on CIFAR10 v.s. 2D box example.} 
In the left of Figure~\ref{fig:adv-bvr-mainline-2d} (2D box example), we see that the bias is monotonically increasing and behaves similarly to experiments on real image data, so properties P1 and P4 hold. 
However, \textit{the variance of the 2D box example is different}: namely, monotonically increasing instead of unimodal, so P2 and P4 do not hold. 
This implies that the conceptual ``2D box example'' is limited for understanding adversarial training.

\vspace{0.1in}
We offer an intuitive explanation for the monotonically increasing variance for the 2D box example. In this model, larger $\varepsilon$ makes the decision boundary more jagged and complex. In Figure~\ref{fig:2d_vis}, we present two random draws of this 2D box example. Although the decision boundaries for different random draws successfully avoid almost all the boxes, the two boundaries are very different, which leads to increased variance. 

\vspace{-0.1in}
\paragraph{Higher dimensional box example.} The inconsistent behavior of the variance is partly due to the low dimensionality of the model. We can generalize of the current 2D setting to higher dimensions ($d\gg 2$) by considering data $\bx$ sampled uniformly from
$\{\bx\in [-1, 1]^d\,:\, |\langle \bx, \mathbf{1}/\sqrt{d}\rangle|\geq \gamma \}$, and $y = \text{sign}(\langle \bx, \boldsymbol 1\rangle)$. We set the sample size as $n=10\cdot d$ and keep the other parameters fixed. 
On the left of Figure \ref{fig:adv-bvr-mainline-2d}, we observe a unimodal variance for a higher dimensional box example, as for the image data. As summarized in the third two rows of Table \ref{table:robust-check}, P1-P4 all hold. The distinction highlights the importance of high-dimensionality of a good conceptual model for adversarial training.
We also include results other dimensions in Appendix~\ref{sec:appendix-toy-model}.

\begin{figure*}[t]
  \begin{center}
    \subfigure[\label{fig:logistic_bvr}Bias/variance on MOG.]{
    \includegraphics[width=.22\textwidth]{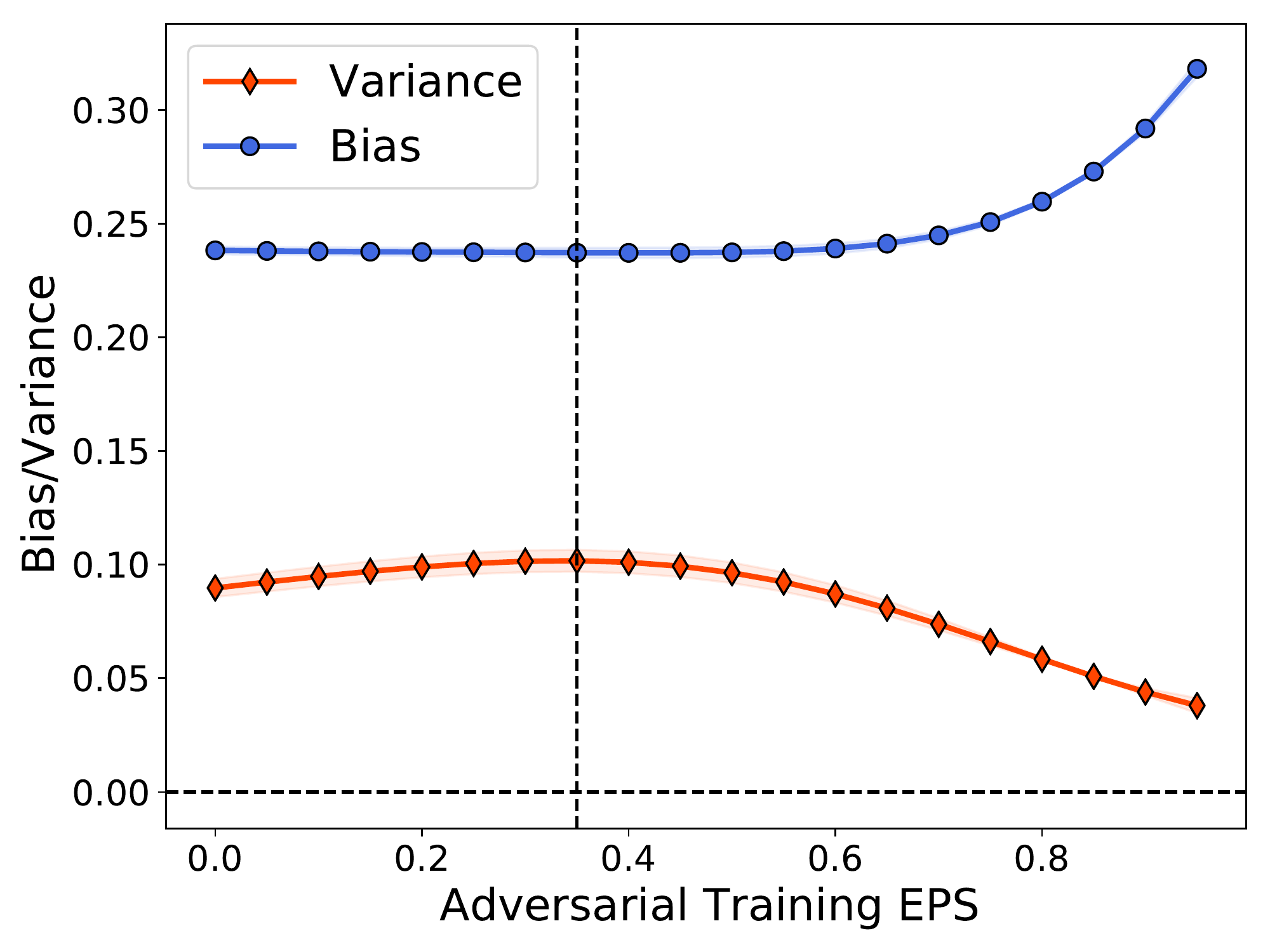}
    }
    \subfigure[\label{fig:logistic_err}Error on MOG.]{
    \includegraphics[width=.22\textwidth]{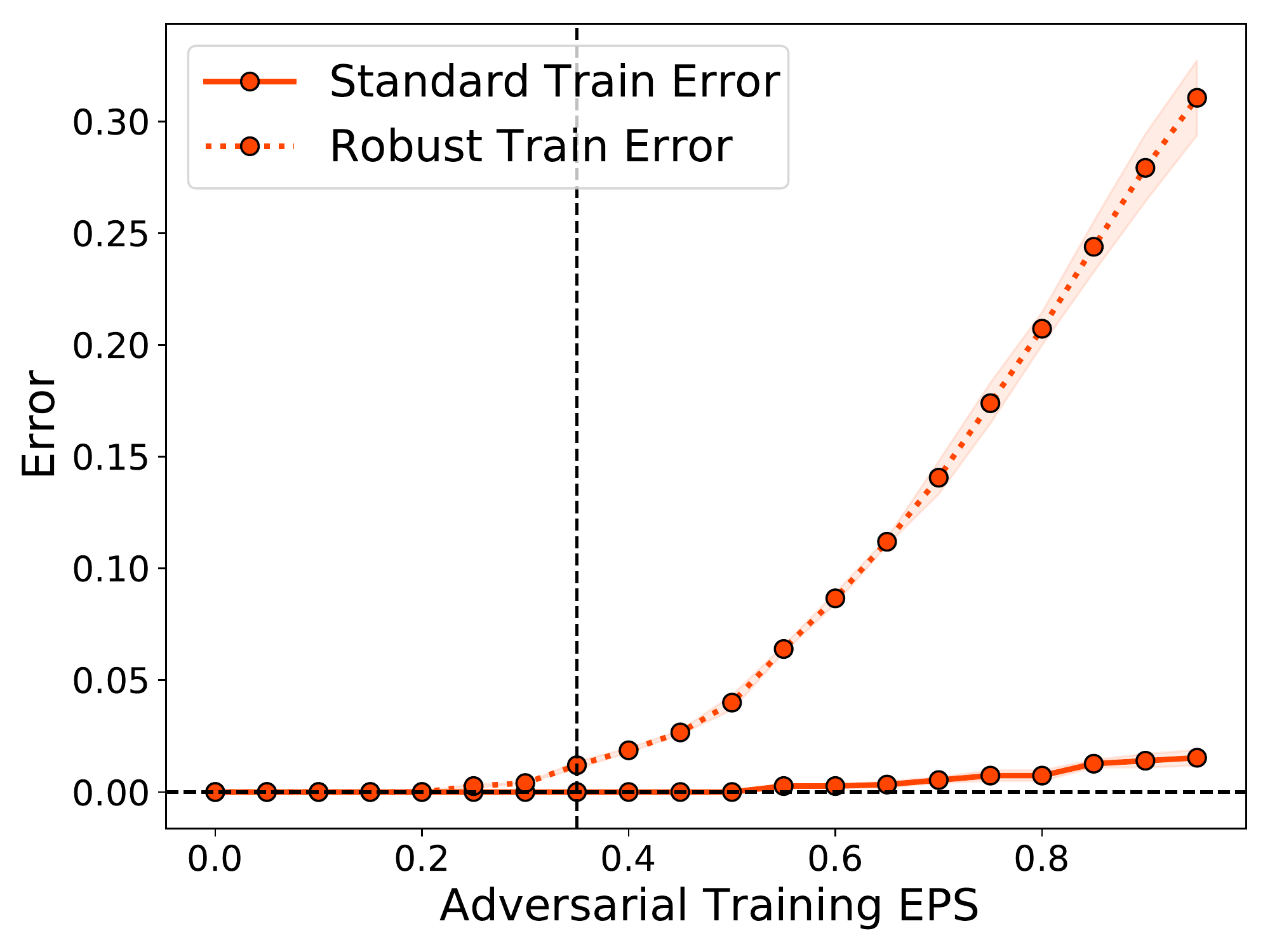}
    }
    \subfigure[\label{fig:logistic_bv_rob}Bias/variance on robust feature distribution.]{
    \includegraphics[width=.22\textwidth]{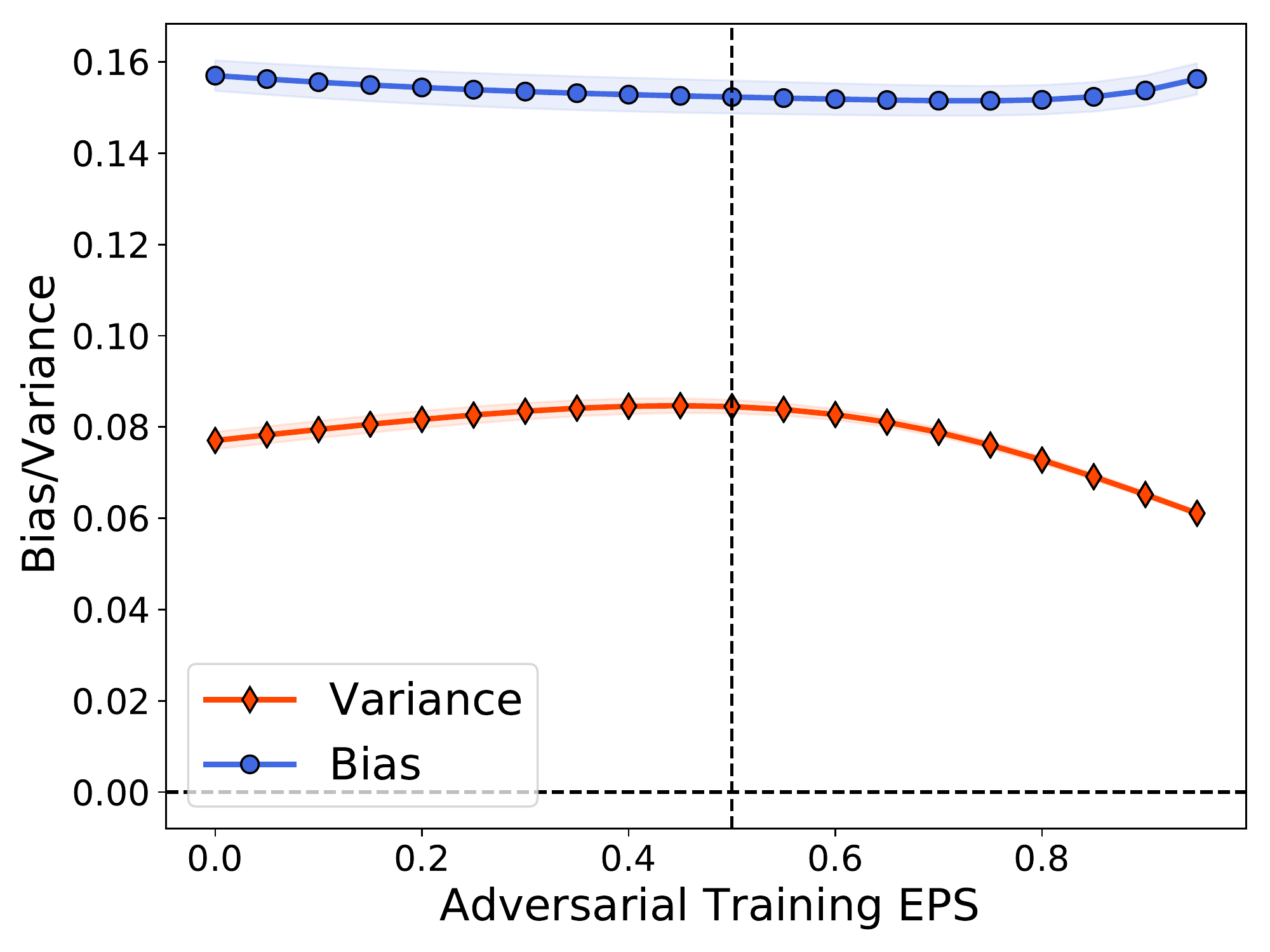}
    }
    \subfigure[\label{fig:logistic_err_rob}Error on robust feature distribution.]{
    \includegraphics[width=.22\textwidth]{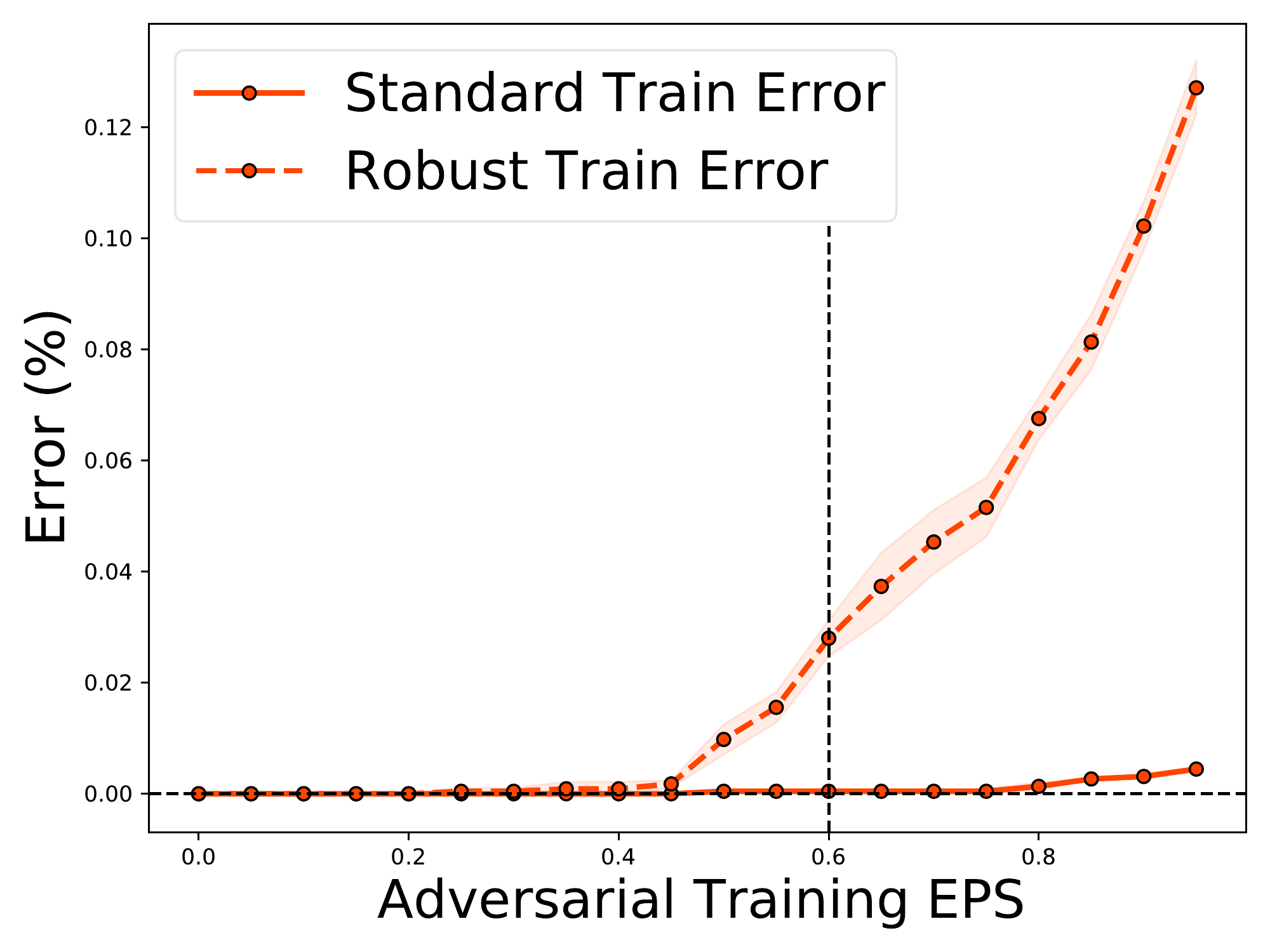}
    }
    \vspace{-0.15in}
    \caption{
    \textbf{(a)(b):} Bias, variance and training error for adversarial logistic regression with mixture of Gaussian (MoG) data. 
    \textbf{(c)(d):} Bias, variance and training error for adversarial logistic regression with the robust feature distribution described in Eq.~\eqref{eq:rob_fea_dist}. For all four figures, the dashed line indicates robust interpolation threshold.
    }
  \end{center}
  \vspace{-0.2in}
\end{figure*}

\subsection{Logistic Regression}\label{sec:toy_logistic}
Our investigation of the 2D box example leads us to consider an intrinsically high dimensional model. In this section, we consider a standard logistic regression setup with adversarial training. 

\vspace{-0.12in}
\paragraph{Adversarial logistic regression.} We use the standard cross-entropy loss for logistic regression
\[
  \sR^{(n)}(\btheta) = \frac{1}{n} \sum_{i=1}^n \ell(y_i\<\bx_i, \btheta\>),
\]
where $\ell(z) = \log(1+e^{-z})$. The inner maximization problem (using $\ell_2$ perturbation of magnitude $\varepsilon$) in the adversarial loss can be solved exactly (and is often used for theoretical analysis~\citep{javanmard2020precise-COLT, javanmard2020precise,dobriban2020provable}):
\begin{align}\label{eqn:logistic_adv_loss}
  \AR^{(n)}(\btheta)= \frac{1}{n} \sum_{i=1}^n \max_{\|\bdelta_i\|_2\leq\varepsilon}\ell(y_i\<\bx_i, \btheta\>) = \frac{1}{n} \sum_{i=1}^n  \log\left(1+e^{-y_i\<\bx_i, \btheta\> +\ep\|\btheta\|_2}\right).
\end{align}
Adversarial training amounts to solving the convex program
$\htn  = \arg\min_{\btheta} \AR^{(n)}(\btheta).$
Given a test data-point $(\bx, y)$, the risk is
$\sR(\htn) =  \E_{\bx, y}\, \ell(y\<\htn, \bx\>)$. 
Instead of the bias-variance decomposition for the squared loss used in other sections, we use a bias-variance decomposition for logistic loss introduced in \ref{sec:logistic_bv}, which is more natural for logistic regression.  We will study two distributions for the underlying data $(\bx, y)$.

\vspace{-0.05in}
\paragraph{Mixture of Gaussians.} As studied in previous works~\citep{dobriban2020provable,javanmard2020precise}, we assume that, for $i=1,\dots,n$, each pair of training samples $\bx_i\in\R^d$ and $y_i\in\{-1, 1\}$ has distribution
\begin{equation}\label{eqn:logistic_generative_model}
  y \overset{\text{u.a.r}}{\sim} \{-1, +1\}, \quad \bx_i|y_i \sim \cN(y_i\bv, \sigma^2\bI_d),
\end{equation}
where $\bv = [1, 1,\dots,1]/\sqrt{d}\in\R^d$ specifies the center of the cluster.
For this experiment, we set the number of training samples $n=100$, the data dimension $d=100$, and the cluster radius (variance of the Gaussian mixture) $\sigma=0.7$. 

\vspace{-0.12in}
\paragraph{Planted robust feature distribution.} To investigate whether the observed bias-variance behavior in adversarial logistic regression is tied to the the specific mixture-of-Gaussians, we also consider a different distribution proposed in \citet{tsipras2018robustness}. For a data-label pair $(\bx, y)$, we let
\begin{equation}\label{eq:rob_fea_dist}
y \overset{\text{u.a.r}}{\sim} \{-1, +1\},~~
\bx_1 = \begin{cases}
        +y, &\text{w.p. }0.95\\
        -y, &\text{w.p. }0.05
      \end{cases},~~
(\bx_2,\dots,\bx_{d}) \overset{\text{i.i.d}}{\sim} \cN(y\bv, \bI_{d-1}),
\end{equation}
where $\bv = [1, 1,\dots,1]/\sqrt{d}\in\R^{d-1}$. In this model , an additional robust feature is added to the first coordinate of the of the feature vector $\bx_1$ and the rest of the coordinates are sampled from a mixture of Gaussians. 
The first coordinate $\bx_1$ can be interpreted as the ``robust feature'' since it is hard to be perturbed. 
Meanwhile, the robust feature is less accurate because itself can only provide $95\%$ accuracy. More details discussions on the trade-off between robustness and accuracy can be found in \citet{tsipras2018robustness}. 
We set the dimension $d=100$ and the number of training samples $n=150$.

\vspace{-0.17in}
\paragraph{Results.} Results for both models are depicted in Figure~\ref{fig:logistic_bvr}. In both cases we see that properties P2 and P3 hold--the variance increases and then decreases with $\varepsilon$, and the peak occurs at the robust interpolation threshold. P4 also holds--the bias dominates the variance. However, P1 does not quite hold--the bias initially decreases at the start of training before eventually increasing. We believe the slight decrease in bias is attributable to the exceeding simplicity of probabilistic model for the data. For both distributions, the linear classifiers are already optimal in the population. Therefore slightly increasing $\varepsilon$ does not lead to a qualitatively incorrect decision boundary as it could for deep neural networks on images. However for large enough $\varepsilon$ the bias still increases as expected. We summarize the results in the last two rows of Table \ref{table:robust-check}.

\begin{figure*}[t]
\begin{center}
\subfigure[Standard bias-variance.]{
\includegraphics[width=.31\textwidth]{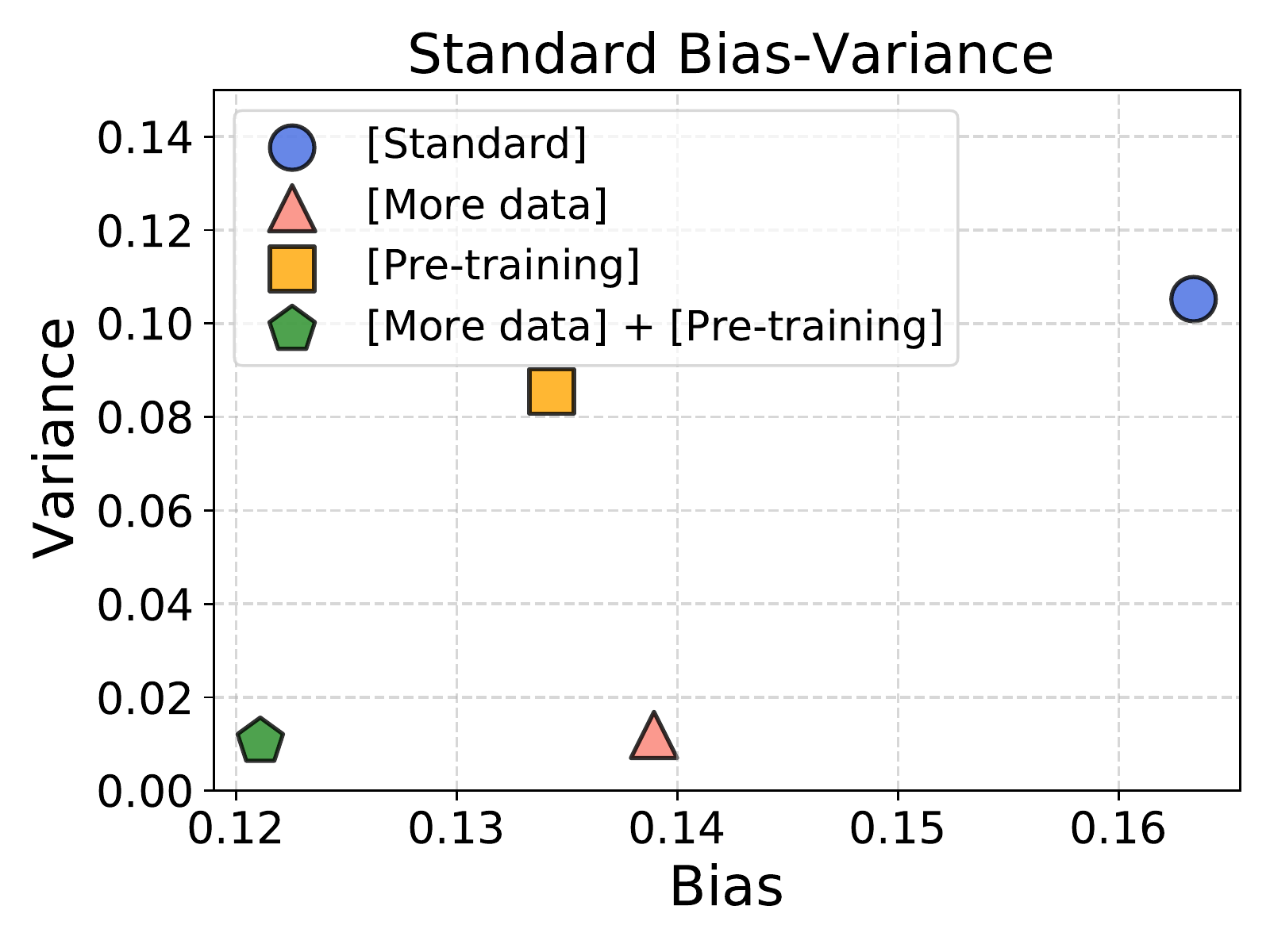}
\label{fig:analysis-moredata-pretrain-bv}
}
\subfigure[Adversarial bias-variance.]{
\includegraphics[width=.31\textwidth]{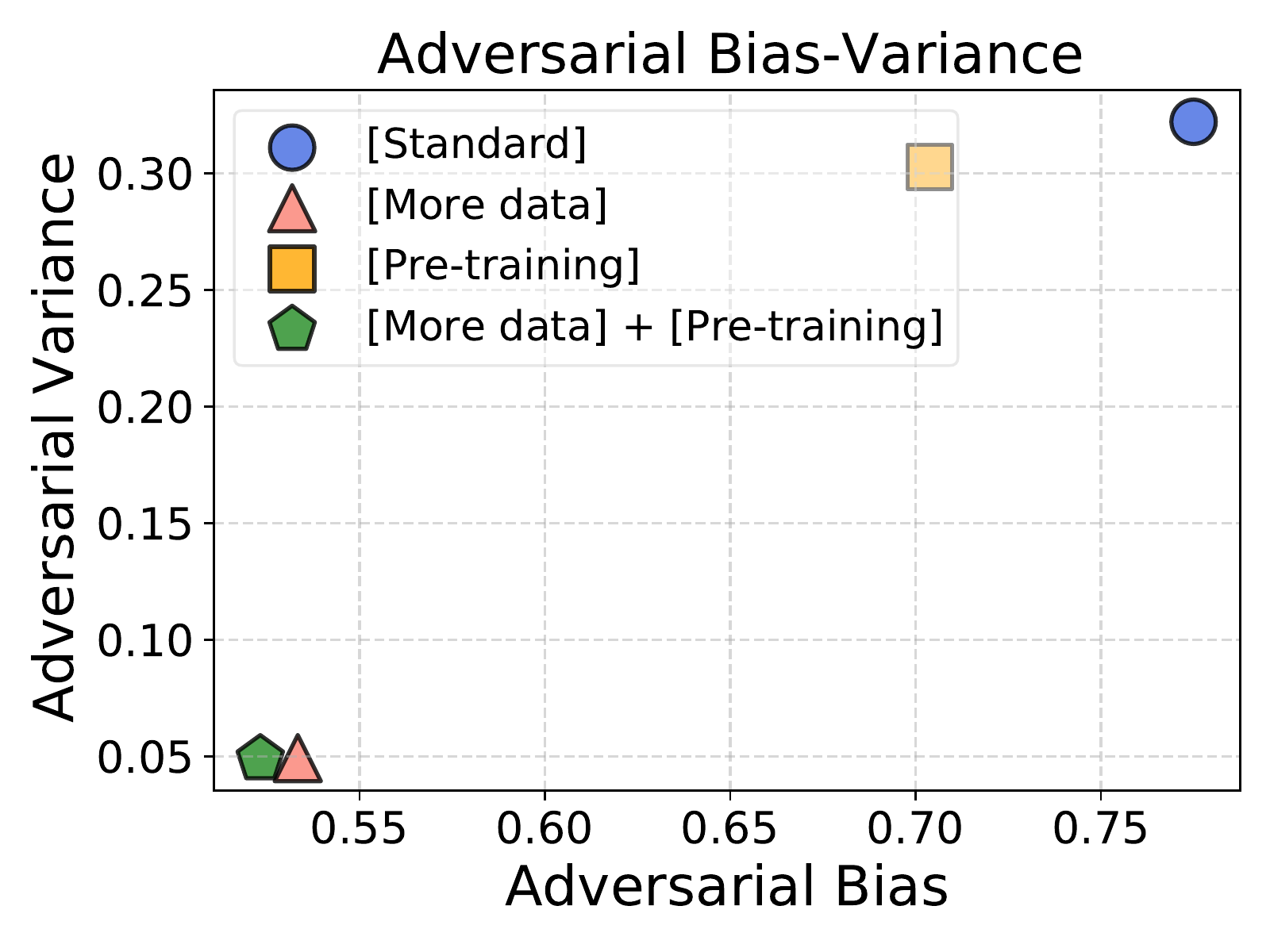}
\label{fig:analysis-moredata-pretrain-adv-bv}
}
\subfigure[Standard and robust error.]{
\includegraphics[width=.31\textwidth]{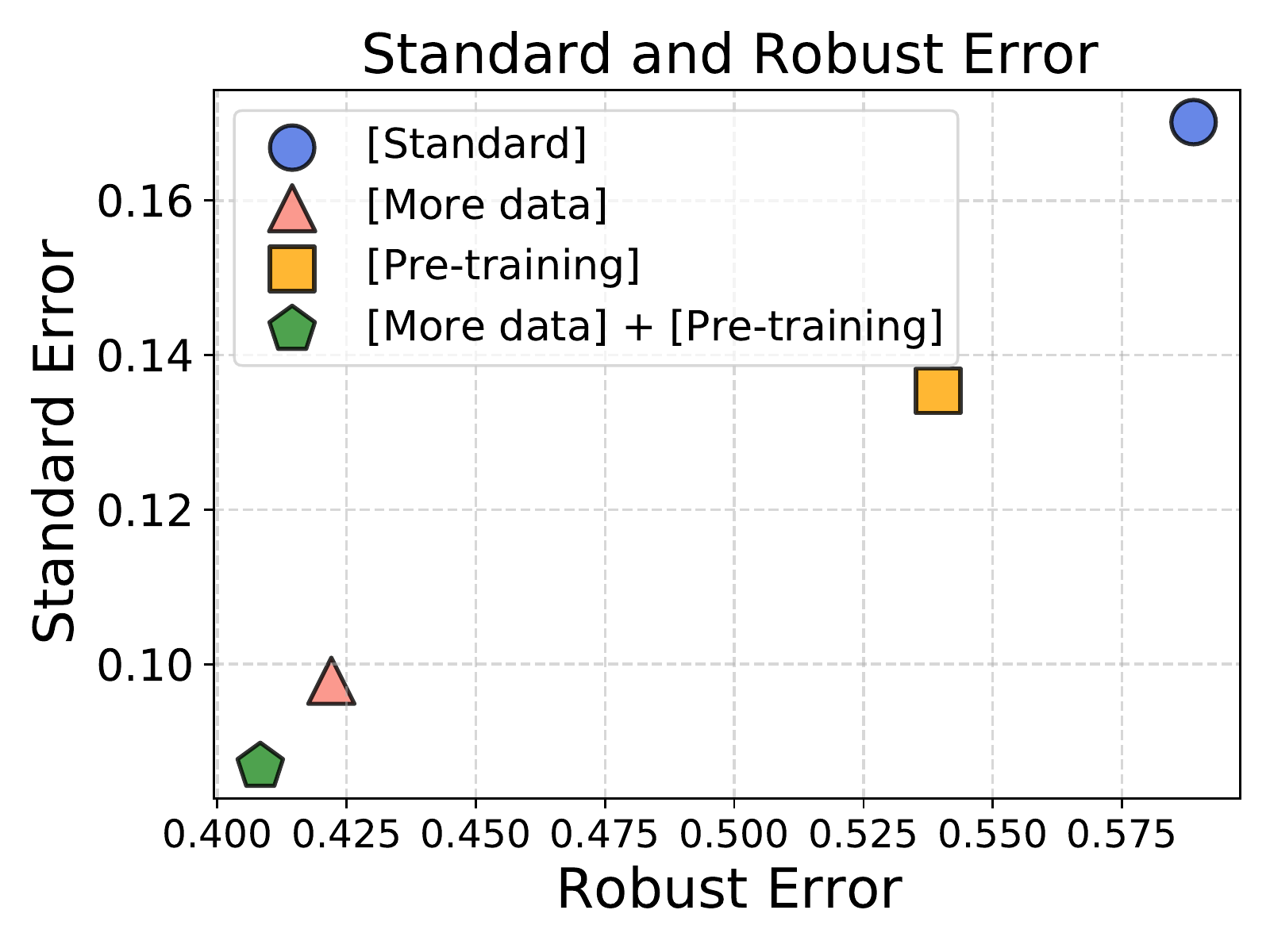}
\label{fig:analysis-moredata-pretrain-error}
}
\vskip -0.1in
\caption{Results of adversarial training on CIFAR10 ($\ell_{\infty}, \ep=8/255$) 
for four different methods: {\scriptsize \textsf{[Standard]}} -- standard adversarial training; {\scriptsize\textsf{[More data]}} -- using additional unlabeled data;  {\scriptsize\textsf{[Pre-training]}} -- applying adversarial pre-trained model;  {\scriptsize\textsf{[More data]+[Pre-training]}} -- using both additional unlabeled data and adversarial pre-training. Points toward lower left indicate better performance in all three figures.}
\label{fig:analysis-moredata-pretrain}
\end{center}
\vskip -0.2in
\end{figure*}

\begin{figure*}[ht]
  \begin{center}
    \includegraphics[width=.43\textwidth]{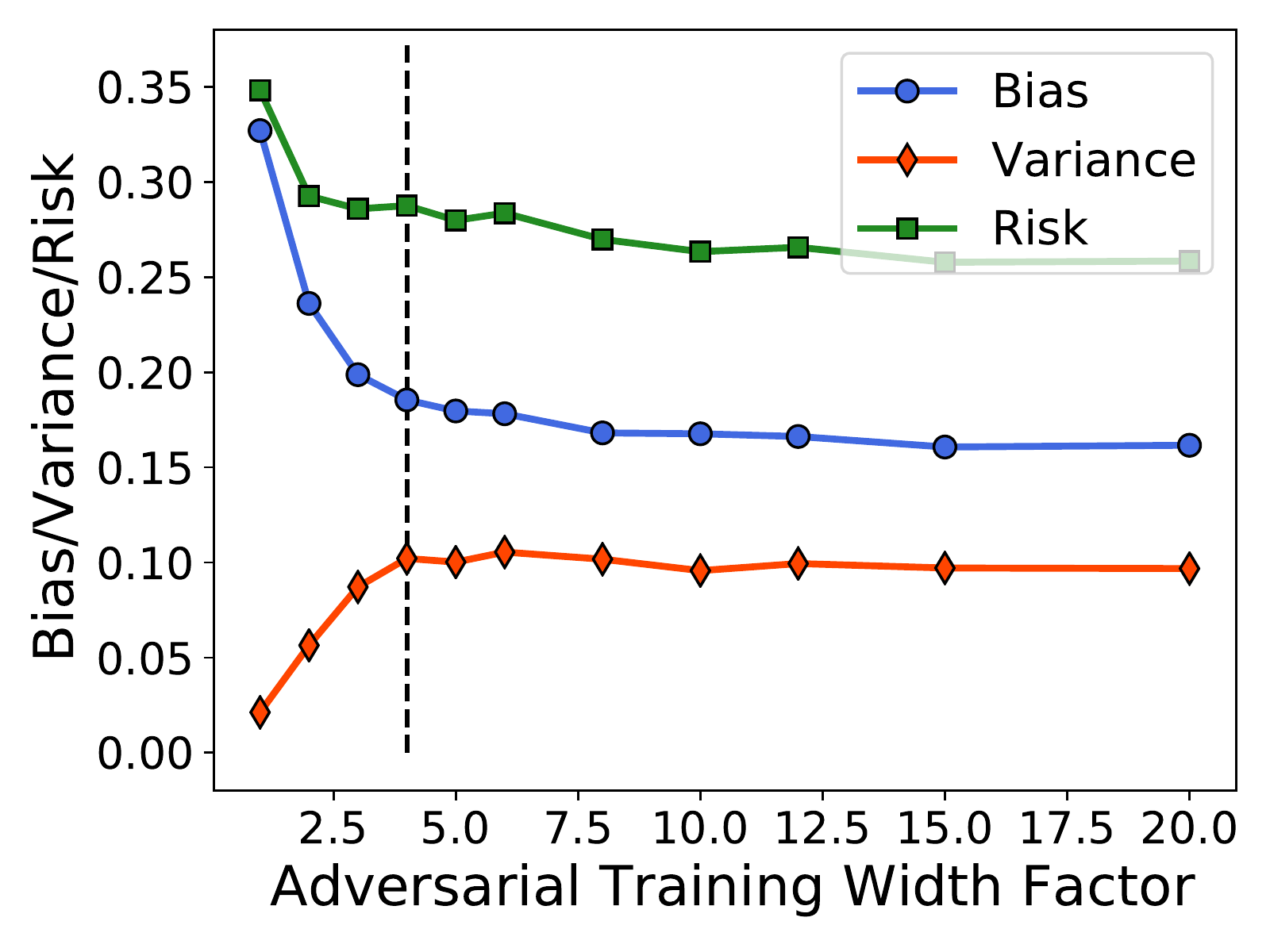}
    \includegraphics[width=.43\textwidth]{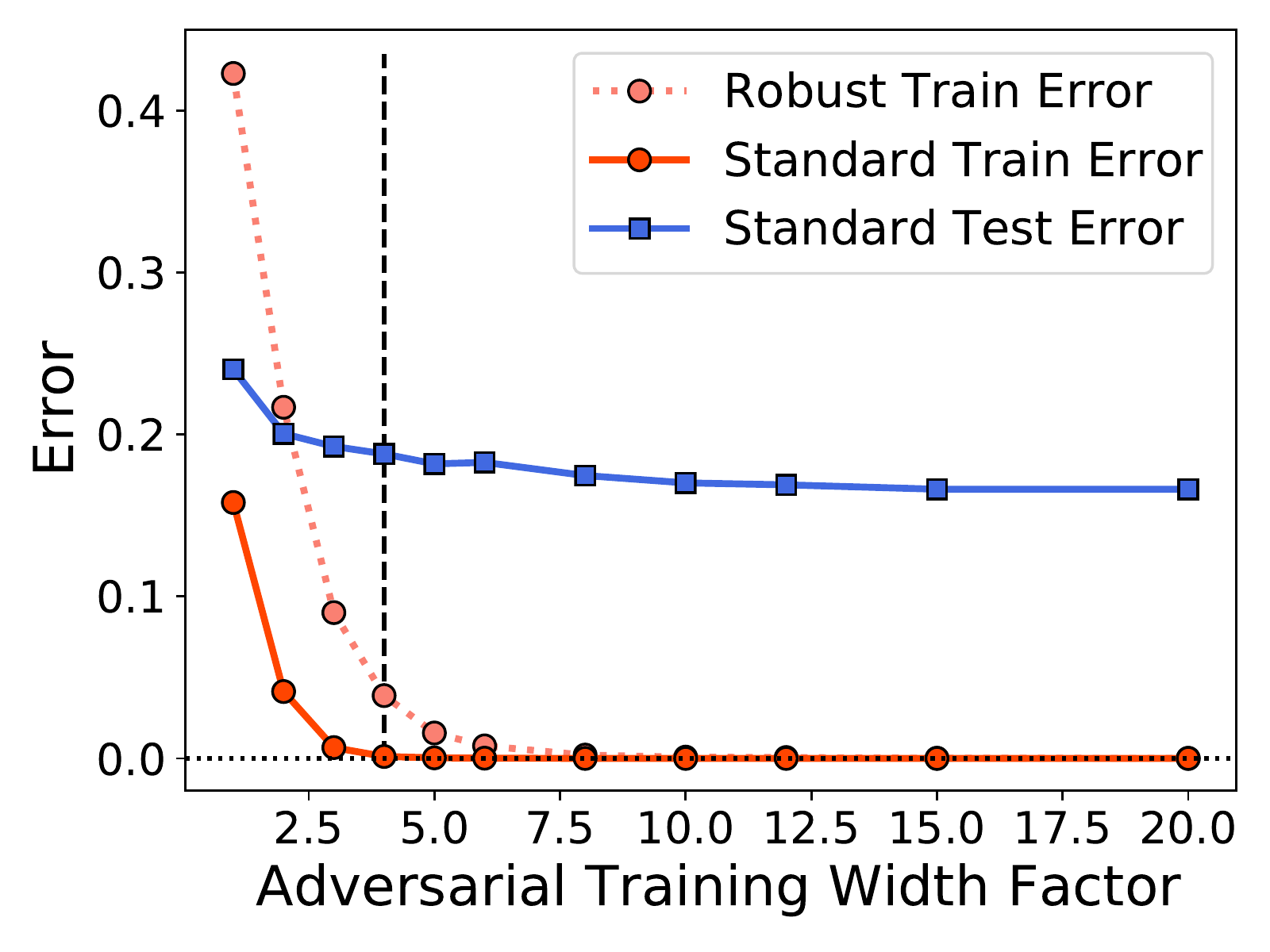}
    \vskip -0.2in
    \caption{Measuring performance for $(\ell_{\infty}, \varepsilon = 8 / 255)$-adversarial training (with increasing width factor for WideResNet-28-[\textit{width}]) on \textit{CIFAR10} dataset.
    (\textbf{left}) Evaluating bias, variance, and risk.
    (\textbf{right}) Evaluating robust training error and standard training/test error. The dashed line indicates the robust interpolation threshold. 
    }
  \label{fig:width-adv-linf-bvr-last-epoch-cifar10-main-text}
  \end{center}
  \vskip -0.3in
\end{figure*}

\section{Analyzing Model Performance via the Bias-Variance Decomposition}\label{sec:performance-analysis}
In this section, we apply the bias-variance decomposition to analyze the performance of adversarially trained models. 
We measure the bias and variance of techniques that improve the performance of adversarial training, including using \textit{additional unlabeled data}~\citep{carmon2019unlabeled} and \textit{adversarial pre-training}~\citep{hendrycks2019using}. We also study the bias-variance decomposition for $\ell_{\infty}$-adversarially trained models with different width factors. 
Our motivation is to understand whether these techniques can reduce the bias, since that is the dominant term in the risk, and the previously considered early stopping technique primarily reduces variance.

\vspace{-0.15in}
\paragraph{Analyzing techniques that improve adversarial training.} Additional unlabeled data and adversarial pre-training can improve model performance for both standard and robust accuracy~\citep{carmon2019unlabeled, hendrycks2019using, Salman2019ProvablyRD, gowal2020uncovering}. We consider the same Wide-ResNet architecture (WRN-28-10) as before, trained at $\ep=8/255$ in $\ell_{\infty}$-norm. We investigate four cases:  (1).~Standard adversarial training; (2).~Using additional unlabeled data; (3).~Using adversarially pre-trained models for initialization; (4).~Using both additional unlabeled data and adversarial pre-training. We measure the standard bias and variance, their adversarial counterparts (see Section~\ref{sec:adv-bv-definition-results}), and the standard and robust test errors in each case. The results are summarized in Figure~\ref{fig:analysis-moredata-pretrain}.

In Figure~\ref{fig:analysis-moredata-pretrain-bv}, we see that additional unlabeled data significantly decreases both the bias and variance, while pre-training mainly decreases the bias. Using both a pre-trained model \emph{and} additional data further decreases the  bias, leading to a smaller clean error (see Figure~\ref{fig:analysis-moredata-pretrain-error}). In terms of the \textit{adversarial} bias-variance decomposition results shown in Figure~\ref{fig:analysis-moredata-pretrain-adv-bv}, we find that adversarial pre-training mainly improves the adversarial bias term, and additional unlabeled data improves both the adversarial bias and adversarial variance.

\vspace{-0.14in}
\paragraph{Model width.} We also consider increasing the width factor of the network, training WRN-28-$[\textit{width}]$ for $\textit{width}\in\{1, \cdots, 20\}$. 
From Figure~\ref{fig:width-adv-linf-bvr-last-epoch-cifar10-main-text}, we find that making models wider only marginally improves the bias beyond a certain width factor. This suggest that increasing width might not be a scalable solution to reduce the generalization gap in adversarial training.

Aside from width, previous works demonstrate that making models deeper can significantly improve generalization in adversarial training~\citep{xie2019feature, gowal2020uncovering}. We hypothesize that using deeper models might be more effective than increasing width, and that they would decrease bias. This would align with 
results for regularly trained models \citep{YYBV2020}, for which deeper models decreased bias at the cost of variance.

\vspace{-0.05in}
\section{Discussion and Future Work}\label{sec:discussion}
\vspace{-0.05in}
In this paper, we studied the generalization gap in adversarial training by measuring the bias and variance of adversarially trained models. 
Across many training settings, we robustly observed that the bias increases monotonically (P1) with the adversarial perturbation radii $\varepsilon$ and is the dominant term of the risk (P4). The variance is unimodal (P2) with its peak near the robust interpolation threshold (P3). 
Beyond their intrinsic interest, P1-P4 allowed us to test 
several conceptual models for the generalization gap and 
thus obtain new insights into its underlying cause.
We further used bias and variance to study how existing techniques have helped reduce the generalization gap. 

\vspace{-0.12in}
\paragraph{Implications for adversarial training.} Our results provide a jumping-off point for reducing the adversarial generalization gap. 
They show that the gap is mainly due to bias, pinpointing the locus for improvement. 
The large bias may be counterintuitive, as it occurs even when the training error is zero, which usually instead corresponds to low bias and high variance. This suggests that adversarial training 
imposes a strong form of \emph{implicit regularization}~\citep{neyshabur2014search} that causes the bias. A better understanding of this regularization could perhaps help us design better architectures to accommodate adversarial training.

\vspace{-0.12in}
\paragraph{Variance peak and interpolation threshold.} Our variance curves consistently reach their maximum near the point where the robust training error becomes non-zero. This echoes results in \citet{YYBV2020}, who found a parallel phenomenon for regular training with models of increasing width.
Their bias/variance-width plot aligns with our bias/variance-$\varepsilon$ plot. 
In both cases, the variance is unimodal with peak near the interpolation threshold. 
The connection here is perhaps deeper: Increasing width increases model complexity and increasing $\varepsilon$ adds regularization, which effectively reduces model complexity. This explains why we observe bias/variance behavior in reverse (increased bias in $\ep$ vs.~decreased bias in width).
The connection hints to a universal behavior of the model variance in relation to a notion of ``effective model complexity'', and deserves refined investigation in future work.

\section*{Acknowledgements}
We would like to thank Preetum Nakkiran, Aditi Raghunathan, and Dimitris Tsipras for their valuable feedback and comments.


\bibliographystyle{plainnat}
\bibliography{reference}

\newpage
\appendix

\newpage
\section{Adversarial Bias-variance Decomposition}\label{sec:adv-bv-definition-results}
In this section, we propose a \textit{generalized} bias-variance decomposition, the \textit{adversarial bias-variance} decomposition, to help us investigate the behavior of adversarial trained models evaluated on adversarial examples. 
A challenge is that in the adversarial setting, we not only need to take into account the prediction at a particular test point $\bx$, but also at a point $\bx' = \bx + \bdelta$, where $\bdelta \in \Delta$ is the adversarial perturbation.

For a robust model, $f_{\widehat{\btheta}(\cT)}\left(\bx + \bdelta (\bx, \by, \cT) \right)$ should still be a good predictor of the label $y$. Therefore, as in the standard training setting, the prediction $f_{\widehat{\btheta}}\left(\bx + \bdelta (\bx, \by, \cT) \right)$ should not depend too much on $\cT$. We can measure how much the prediction depends on $\cT$ by computing the adversarial variance $\AV = \E_{\bx, \by}\Var_{\cT}[ f_{\widehat{\btheta}(\cT)}(\bx + \bdelta(\bx, \by, \cT) ) ]$. The adversarial bias can be analogously defined as $\AB = \E_{\bx, \by} [\| \by - \E_\cT f_{\widehat{\btheta}(\cT)}(\bx + \bdelta (\bx, \by, \cT))\|^{2}]$. 
One can check that the sum of adversarial bias ($\AB$) and adversarial variance ($\AV$) gives the expected adversarial  risk ($\AR$), i.e., 
\begin{equation}\label{eq:adv-bv-l2}
    \AR = \E_{\cT} \E_{\bx, \by} \big[ \max_{\bdelta\in\Delta} \big\| f_{\widehat{\btheta}}(\bx + \bdelta)- \by \big\|^2 \big] = \AB + \AV,
\end{equation}
where $\bdelta_{\cT}$ is short for $\bdelta(\bx, \by, \cT)$. The advantage of our definition of variance is that it directly measures the variation of the model's prediction at test sample $\bx$ in presence of adversarial perturbation.
An algorithm for estimating the variance given a single training dataset $\cT$ at a particular test point $(\bx, \by)$ is given in Algorithm \ref{alg:l2advV} in Appendix~\ref{sec:appendix-bv-prelim}.

In addition to standard bias-variance decomposition, we also study the adversarial bias-variance decomposition (defined in Eq.~\eqref{eq:adv-bv-l2}) for adversarial training and randomized smoothing training on the CIFAR10 dataset.  
We apply the same training dataset partition procedure as bias-variance decomposition. To approximately find  $\bdelta_{\cT}$ defined in Eq.~\eqref{eq:adv-bv-l2}, we use the $\ell_{\infty}$-PGD adversarial attack for generating the adversarial perturbations, where the number of perturbation steps is 20 and the step size is $\eta = 0.15 \cdot (\varepsilon/255)$.

\begin{figure*}[ht]
  \begin{center}
    \subfigure{
    \includegraphics[width=.31\textwidth]{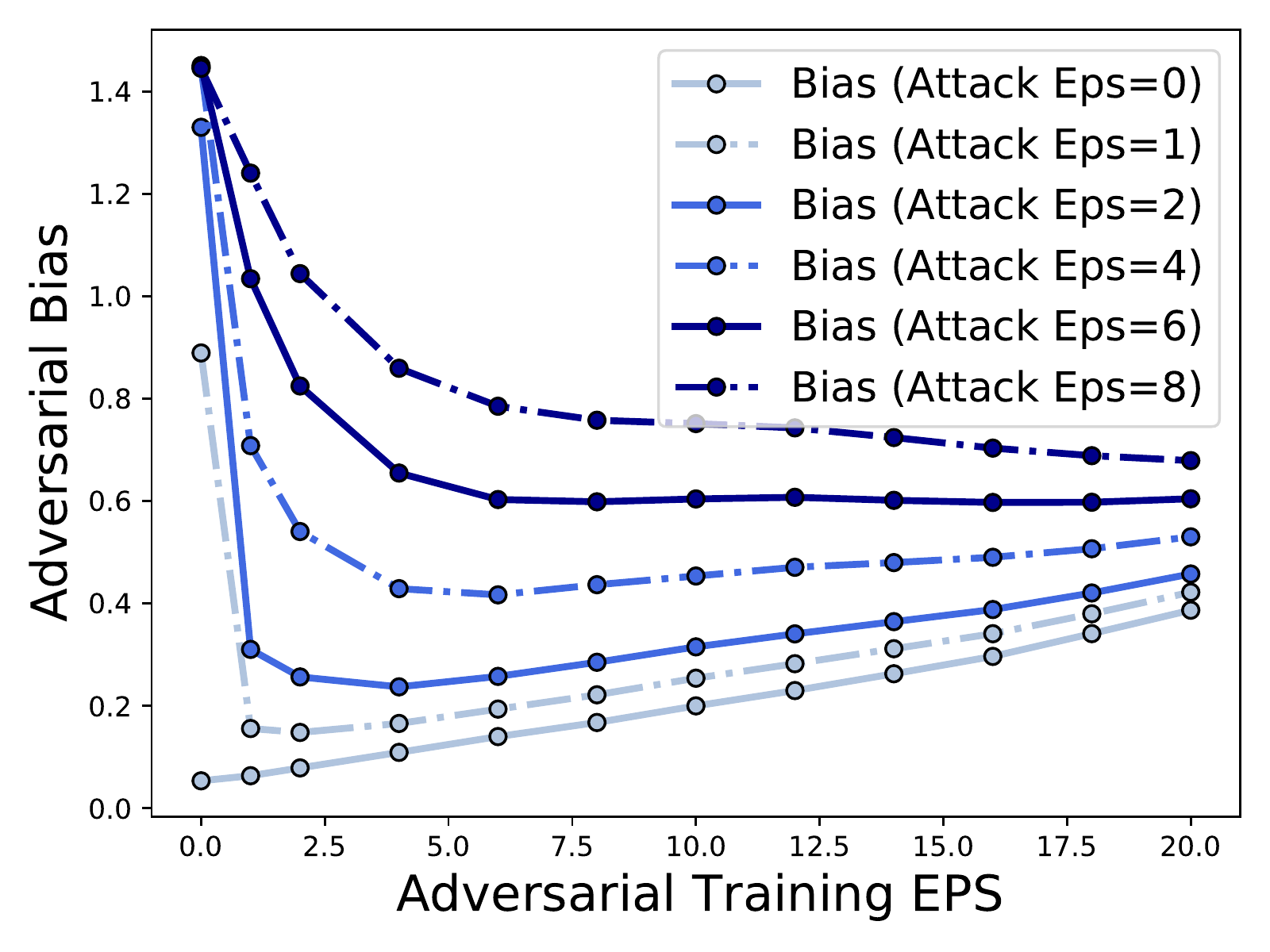}
    }
    \subfigure{
    \includegraphics[width=.31\textwidth]{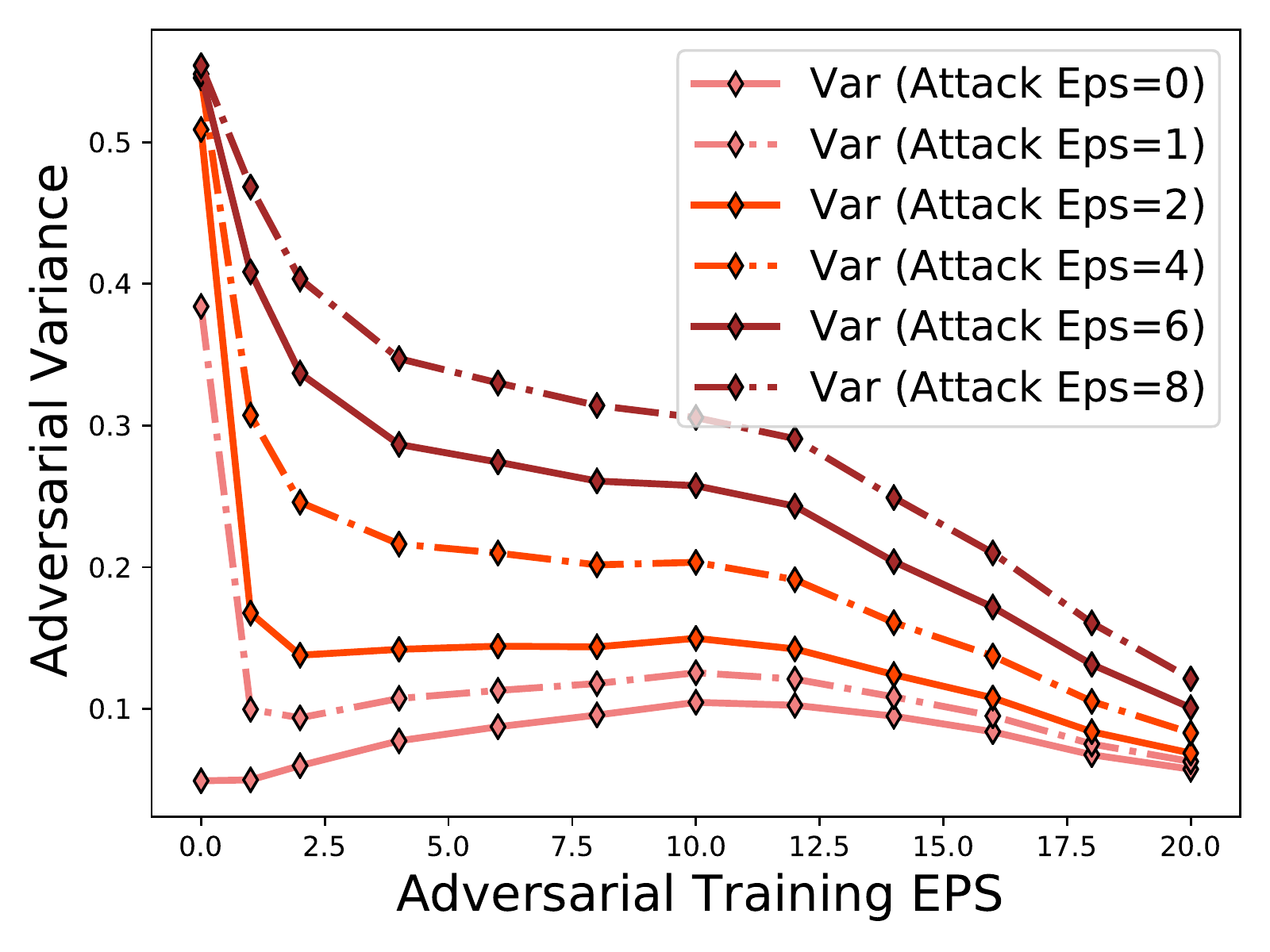}
    }
    \subfigure{
    \includegraphics[width=.31\textwidth]{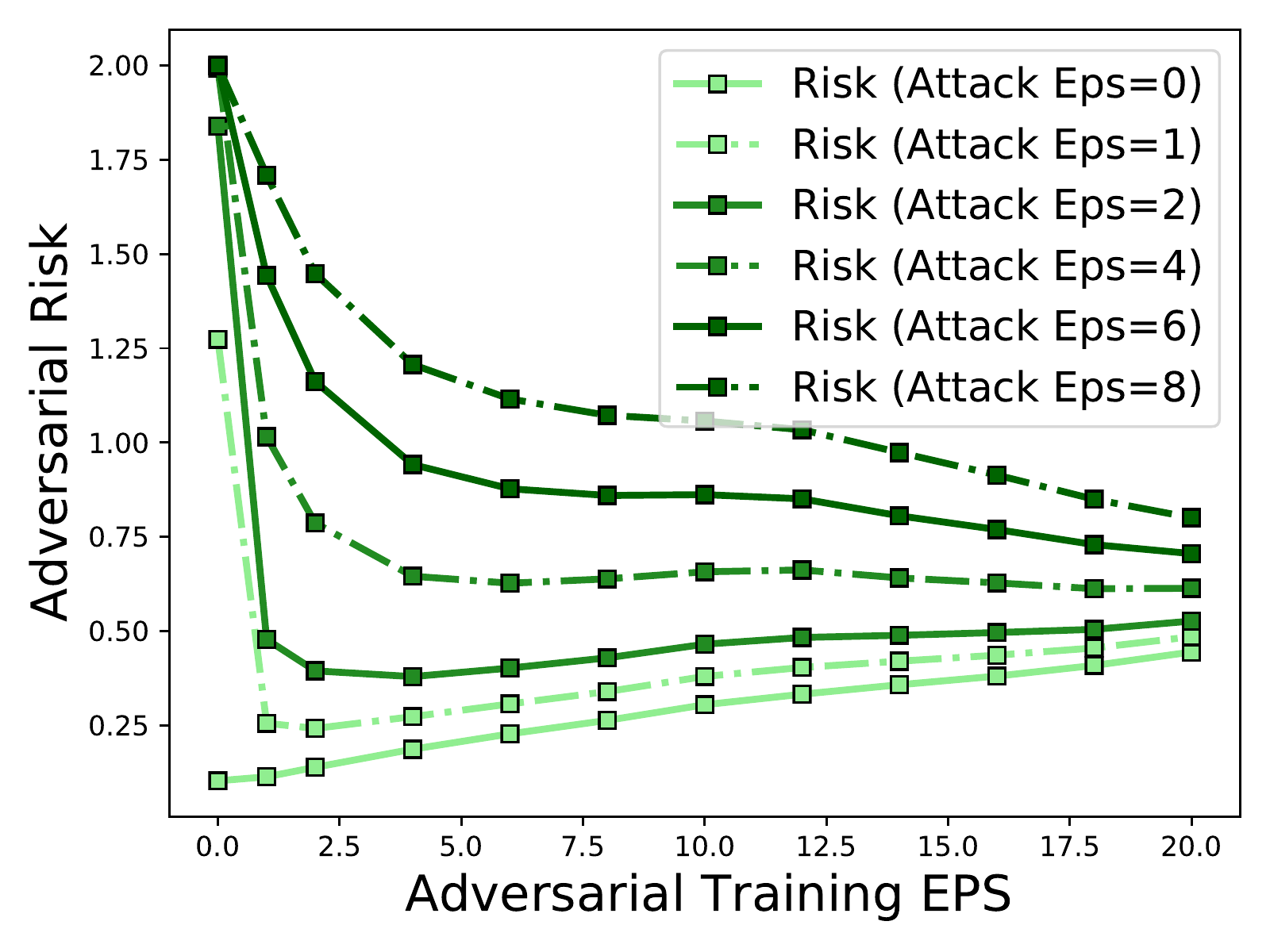}
    }
    \vskip -0.1in
    \caption{Adversarial bias, variance, and risk for $\ell_{\infty}$ \textit{adversarially trained models} on the CIFAR10 dataset using WRN-28-10. Each curve corresponds to the $\varepsilon$-adversarial bias-variance decomposition, and the \textsf{EPS} represents the $\ell_{\infty} = \textsf{EPS}/255.0$ PGD attack. (\textbf{Left}) Adversarial bias. (\textbf{Middle}) Adversarial variance. (\textbf{Right}) Adversarial risk. 
    }
  \label{fig:adv-bvr-mainline}
  \end{center}
  \vskip -0.25in
\end{figure*}

\begin{figure*}[ht]
  \begin{center}
    \subfigure{
    \includegraphics[width=.31\textwidth]{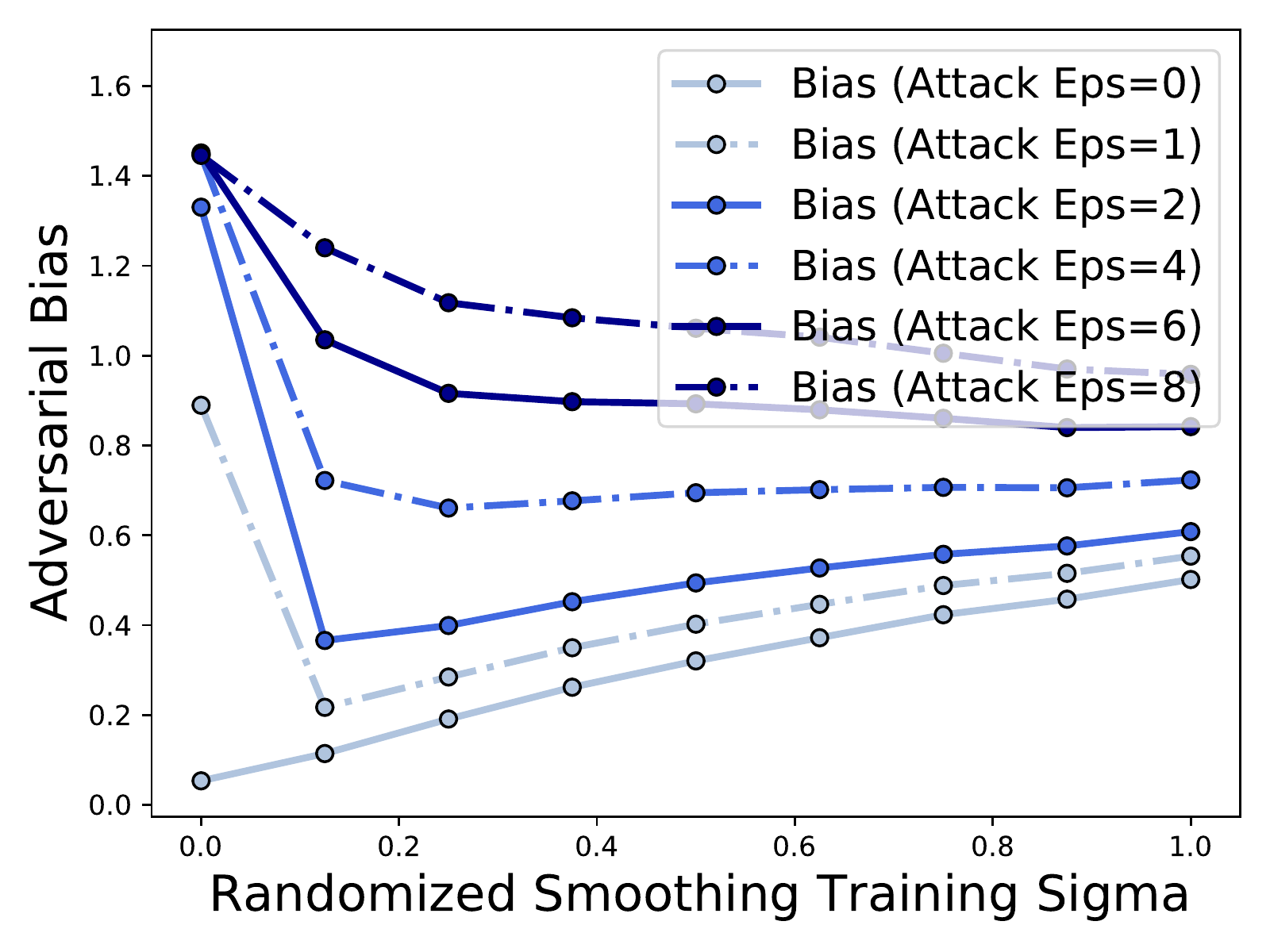}
    }
    \subfigure{
    \includegraphics[width=.31\textwidth]{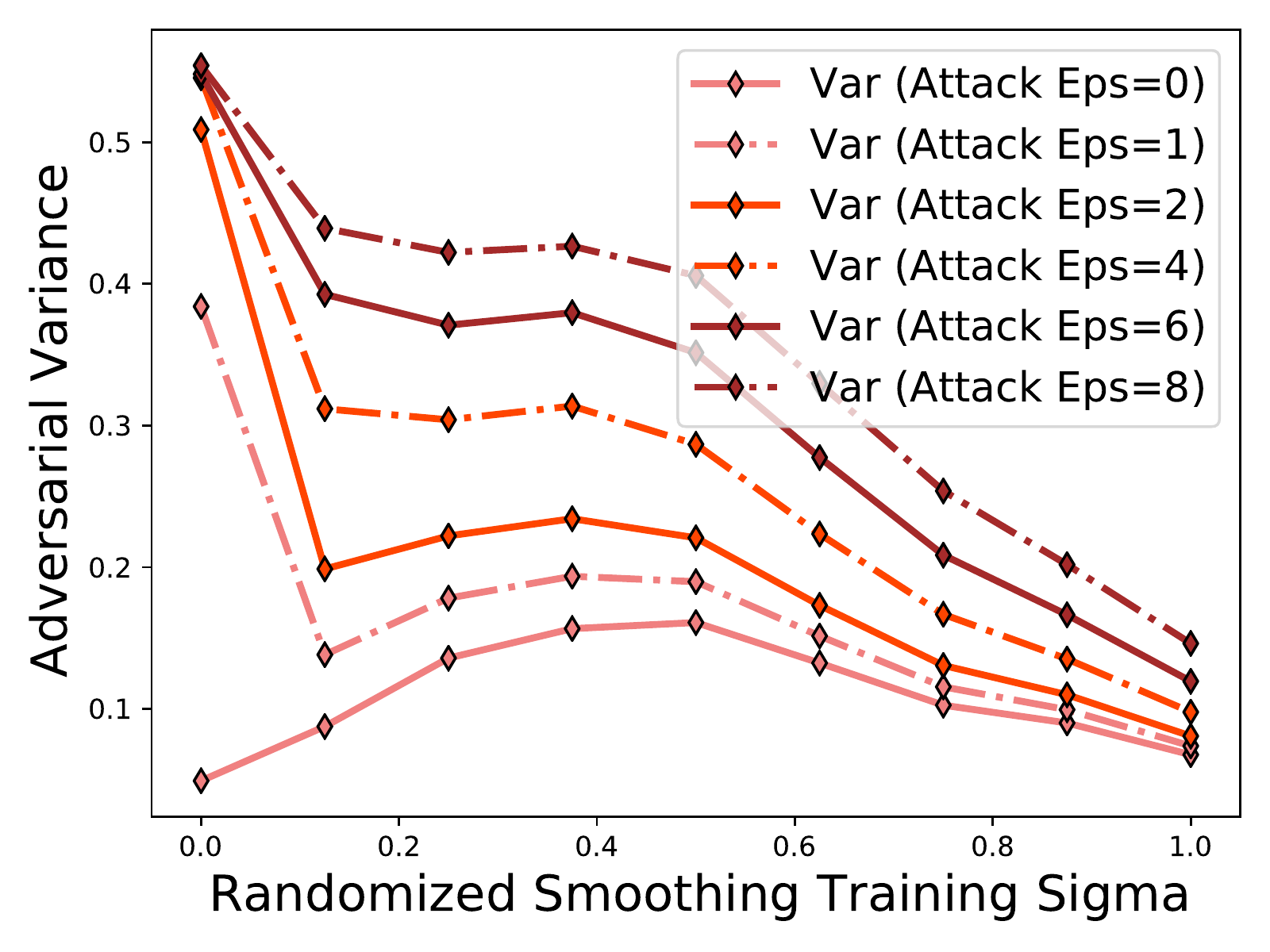}
    }
    \subfigure{
    \includegraphics[width=.31\textwidth]{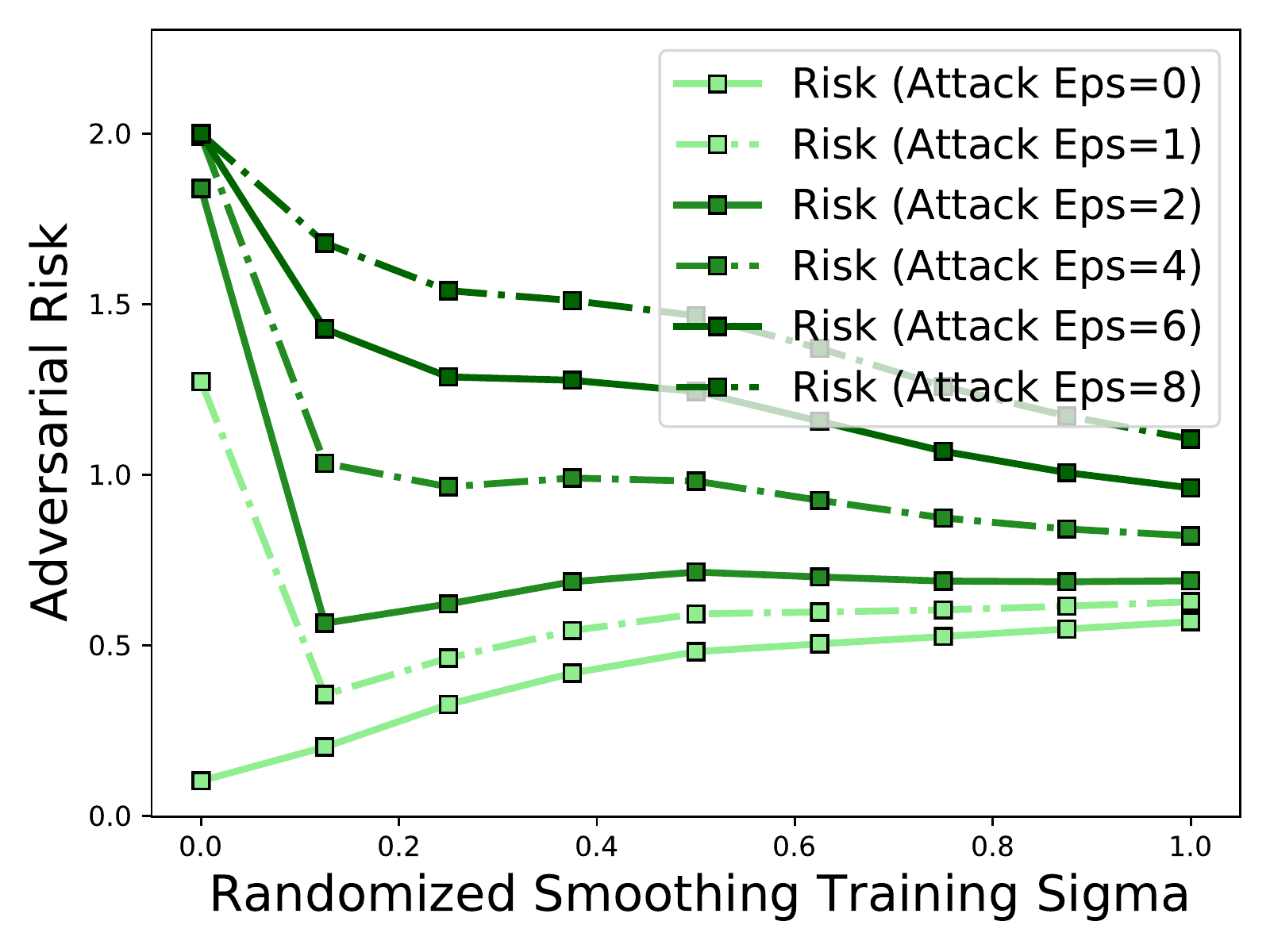}
    }
    \vskip -0.1in
    \caption{Adversarial bias, variance, and risk for \textit{randomized smoothing training} models on the CIFAR10 dataset using WRN-28-10. Each curve corresponds to the $\varepsilon$-adversarial bias-variance decomposition, and the \textsf{EPS} represents the $\ell_{\infty} = \textsf{EPS}/255.0$ PGD attack. (\textbf{Left}) Adversarial bias. (\textbf{Middle}) Adversarial variance. (\textbf{Right}) Adversarial risk. 
    }
  \label{fig:rs-training-adv-bvr}
  \end{center}
  \vskip -0.1in
\end{figure*}

The results for the $\ell_{\infty}$ adversarially trained models are summarized in Figure~\ref{fig:adv-bvr-mainline}. With increasing perturbation size for evaluating the adversarial bias and variance, the adversarial bias curve is changed from monotone increasing to monotone decreasing, and the variance curve is changed from unimodal to monotone decreasing. The adversarial variance is relatively small compared with the adversarial bias under various attack radii; similar to the standard bias-variance results in Figure~\ref{fig:bvr-mainline}. Furthermore, from Figure~\ref{fig:rs-training-adv-bvr}, we observe that models trained with randomized smoothing have similar characteristic adversarial bias-variance curve shapes as adversarially trained models (in Figure~\ref{fig:adv-bvr-mainline}): the adversarial variance under all adversarial perturbations becomes smaller when increasing the training variance $\sigma^{2}$, and the bias shape is changed from monotone increasing to monotone decreasing with increased adversarial perturbation $\varepsilon$.

\section{Algorithm for Estimating Bias-variance}\label{sec:appendix-bv-prelim}
In this section, we present the algorithm for  estimating the (\textit{adversarial}) variance for squared loss. Also, we introduce the \textit{adversarial} bias-variance decomposition for cross-entropy loss.

\subsection{Algorithm for Estimating the Variance}
The algorithm for estimating the variance is described in Algorithm~\ref{alg:l2advV}. The adversarial variance and bias can be estimated as follows:

\begin{equation}\label{eq:adv-bv-eval}
\begin{aligned}
  \widehat{\textsf{AVar}} &= \E_{\bx, \by}\left[\widehat{\textsf{AVar}}(\bx, \by)\right],\\
\widehat{\textsf{ABias}} &= \frac{1}{K}\sum_{k=1}^K\E_{\cT^{(k)}} \E_{\bx, \by} \left[ \max_{\bdelta\in\Delta} \left\| f_{\widehat{\btheta}(\cT^{(k)})}(\bx + \bdelta)- \by \right\|^2 \right] - \widehat{\textsf{AVar}},
\end{aligned}
\end{equation}
where $\widehat{\textsf{AVar}}(\bx, \by)$ is calculated by Algorithm~\ref{alg:l2advV}. We can also apply Eq.~\eqref{eq:adv-bv-eval} for evaluating the standard bias and variance by setting $\varepsilon=0$ for the perturbation set $\Delta = \{\bdelta : \|\bdelta\|_{p} \leq \varepsilon\}$.
\begin{algorithm}[ht]
\caption{Estimating Adversarial Variance for Squared Loss}
\label{alg:l2advV}
\begin{algorithmic}
 \STATE {\bfseries Input:} Test point $(\bx, \ \by)$, Training set $\cT$, Number of repetitions $K$.
 \FOR{$k=1$ {\bfseries to} $K$}
 \STATE{Split $\cT$ into $\cT_1^{(k)}, \dots, \cT_N^{(k)}$.}
  \FOR{$j=1$ {\bfseries to} $N$}
  \STATE{Perform adversarial training on training dataset $\cT_j^{(k)}$;}
  \STATE{Find $\hat{\btheta}(\cT_j^{(k)})$ that approximately solves  $$\min_{\btheta}  \frac{1}{|\cT_j^{(k)}|} \sum_{i\in \cT_j^{(k)}} \max_{\bdelta_i\in\Delta} \, \ell(f_{\btheta}(\bx_i +\bdelta_i), \by_i);$$}
  \STATE{Find adversarial perturbation $\bdelta(\bx, \by, \cT_j^{(k)})$ that approximately solves   $$\max_{\bdelta \in \Delta} \ell\Big( f_{\hat{\btheta}(\cT_j^{(k)})}(\bx+\bdelta), \by \Big);$$}
  \ENDFOR
 \ENDFOR
 \FOR{$k=1$ {\bfseries to} $K$}
 \STATE{Compute  
 \begin{align*}
 &\widehat{\textsf{AVar}}(\bx, \by, \cT^{(k)}) \\
 = &\frac{1}{N-1}\sum_{j=1}^N  \Big\| f_{\hat{\btheta}(\cT_j^{(k)})}(\bx + \bdelta(\bx, \by, \cT_j^{(k)})) -  {
 \frac{1}{N}\sum_{j=1}^N f_{\hat{\btheta}(\cT_j^{(k)})}(\bx + \bdelta(\bx, \by, \cT_j^{(k)})) }
    \Big\|_2^2;
 \end{align*}
    }
 \ENDFOR
 \STATE{Compute $\widehat{\textsf{AVar}}(\bx, \by) = \frac{1}{K}\sum_{k=1}^K \widehat{\textsf{AVar}}(\bx, \by, \cT^{(k)})$.}
\end{algorithmic}
\end{algorithm}
\vskip -0.25in

\subsection{Bias-variance Decomposition for Cross-entropy Loss}
Inspired by \cite{pfau2013bergman}, \cite{YYBV2020} provides a bias-variance decomposition for cross-entropy loss.  We also extend their decomposition to the adversarial setting. We use boldface $\mbf$ instead of $f$ to emphasize that $\mbf_{\btheta}\in\R^c$ is now a vector that represents a probability distribution over the class labels, i.e., $\sum_{i=1}^c \mbf_\btheta(\bx)_i = 1$. We are given a trained model $\hat{\btheta}(\cT)$, a test point $(\bx, \by)\in\R^d\times\R^c$, where $\by$ is the one-hot encoding of the class membership of input $\bx$. We first compute the worst-case perturbation as
\[
  \bdelta(\bx, \by, \cT) \in \arg\max_{\bdelta\in\Delta}\left\{ H(\mbf_{\hat{\btheta}}(\bx+\bdelta), \by)\right\},
\]
where $H(\ba, \bb) = - \sum_{i=1}^c \ba_i \log(\bb_i)$ is the standard cross-entropy loss. At the adversarially perturbed test point $\bx' = \bx'(\bx, \by, \cT) = \bx + \bdelta (\bx, \by, \cT)$, we define the average prediction $\overline{\mbf}$ at $\bx'$ as 
\[
\begin{aligned}
\overline{\mbf}_i \propto \exp\left[ \E_{\cT} \log \mbf_{\hat{\btheta}}(\bx')_i \right],~~ i \in [c], ~~\text{and}~~ \sum_{i=1}^c \overline{\mbf}_i = 1.
\end{aligned}
\]
Now we are ready to state the formula for the adversarial ``bias-variance'' decomposition of the cross-entropy loss:
\begin{equation}\label{eq:adv-ce-bvr}
\begin{aligned}
  &\underbrace{\E_{\cT} \E_{\bx, \by} \left[ \max_{\bdelta\in\Delta} H\left( f_{\hat{\btheta}(\cT)}(\bx + \bdelta), \by \right) \right]}_{\textsf{Adversarial Risk (CE)}} \\
  =\,\,&\E_{\bx, \by} \E_{\cT} H\left( f_{\hat{\btheta}(\cT)}(\bx + \bdelta(\bx, \by, \cT)), \by \right)
  + \underbrace{ \E_{\bx, \by} D_{\textsf{KL}} \left( \by\| \overline{\mbf} \right) }_{\textsf{Adversarial Bias (CE)}}.
\end{aligned}
\end{equation}
The standard bias and variance can be computed analogously.

\section{Additional Experiments for Deep Neural Networks}\label{sec:appendix-additional-exp-results}
In this section, we provide additional experimental results details related to Section~\ref{sec:bv-analysis-dl}. Specifically, we provide the detailed experimental setup for Gaussian noise training in Section~\ref{sec:gauss-noise-setup}. In Section~\ref{sec:appendix-additional-exp-results-c2}, we present additional experimental results for the $\ell_{2}$ and $\ell_{\infty}$ adversarial training on CIFAR10, CIFAR100, and ImageNet10 datasets. 
We also study the effect of the width factor in adversarial training in Section~\ref{sec:appendix-additional-exp-results-c2}. In Section~\ref{sec:appendix-additional-exp-results-c3}, we study the bias and variance of deep models on out-of-distribution (OOD) datasets. In Section~\ref{sec:appendix-additional-exp-results-c4}, we evaluate the \textit{cross-entropy loss} bias and variance for $\ell_{\infty}$ and $\ell_{2}$ adversarially trained models on the CIFAR10 dataset.

\subsection{Experimental setup for training with Gaussian Noise}\label{sec:gauss-noise-setup}
\paragraph{Experimental setup.} For randomized smoothing training, following previous works~\cite{lecuyer2019certified, cohen2019certified}, we train the models with additive Gaussian data augmentation with variance $\sigma^{2}$ on the CIFAR10 dataset. 
The variance $\sigma^{2}$ is chosen as $\sigma \in \{k\cdot 0.125\,:\, k=0, \dots, 8\}$.  
After adding noise to the input, we clip the pixels to $[0.0, 1.0]$. 
For training on Gaussian-perturbed data, we perturb the training images from CIFAR10 with random additive Gaussian noise with variance $\sigma^{2}$, where $\sigma \in \{k\cdot 0.125\,:\, k=0,\dots, 8\}$, then clip pixels to $[0.0, 1.0]$ and save the images using the standard JPEG compression. 
The same architecture (WRN-28-10) and optimization procedure are applied for randomized smoothing training and training on Gaussian-perturbed data as for $\ell_{\infty}$ adversarial training, described earlier. The \textbf{robust error} (in Figure~\ref{fig:rs-bvr}) of randomized smoothing trained models is evaluated by adding random Gaussian noise (with the same variance $\sigma^{2}$ as in training) to the training dataset.

\subsection{Additional Experiments on Measure Bias-variance of Adversarial Training models on CIFAR10/100 and ImageNet10}\label{sec:appendix-additional-exp-results-c2}
For $\ell_{2}$ adversarial training, we apply a 10-step $(\ell_{2})$ PGD attack with perturbation step size $\eta = 0.25\cdot(\varepsilon/255)$. 

\begin{figure*}[ht]
  \begin{center}
    \includegraphics[width=.31\textwidth]{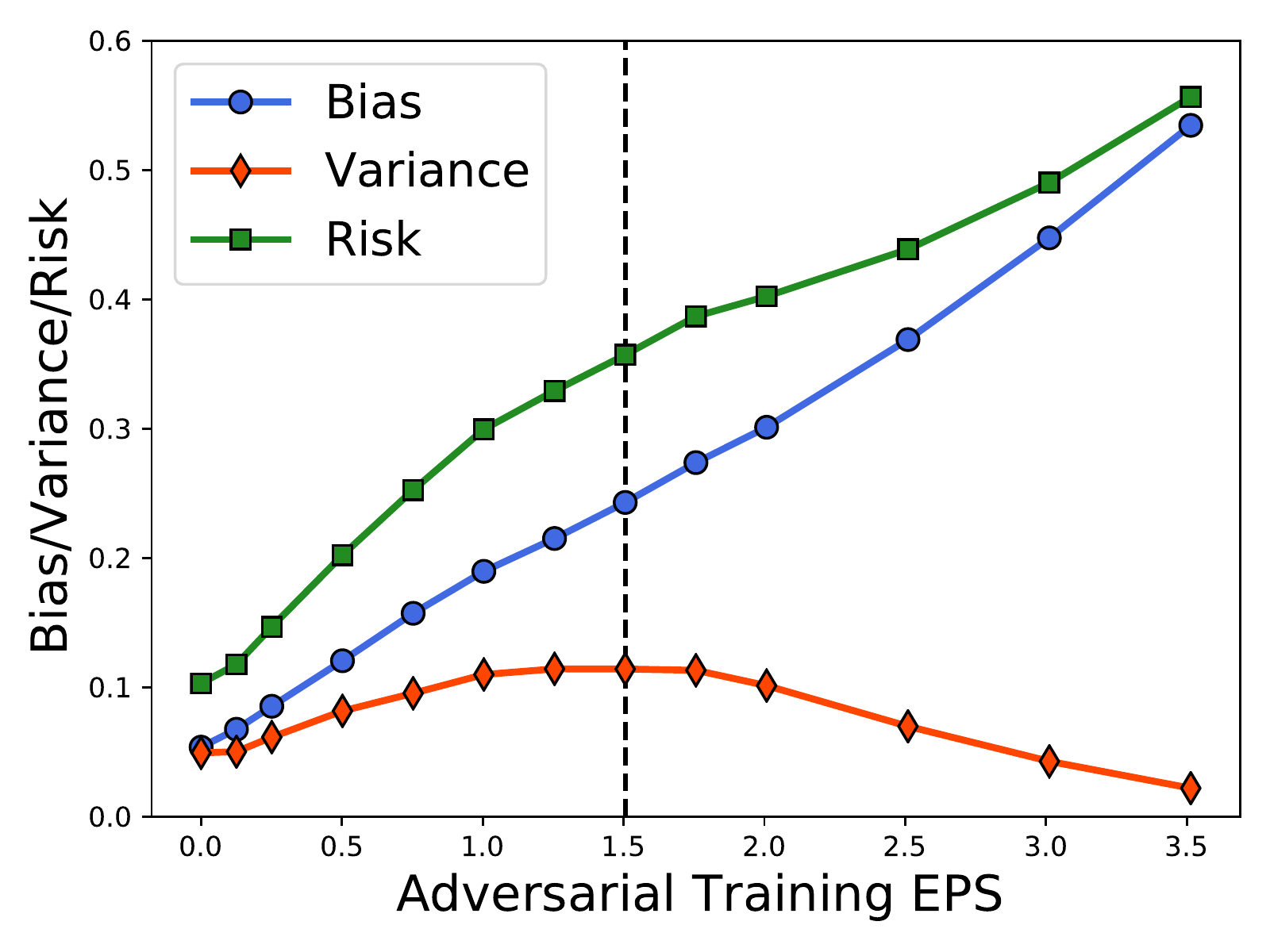}
    \includegraphics[width=.31\textwidth]{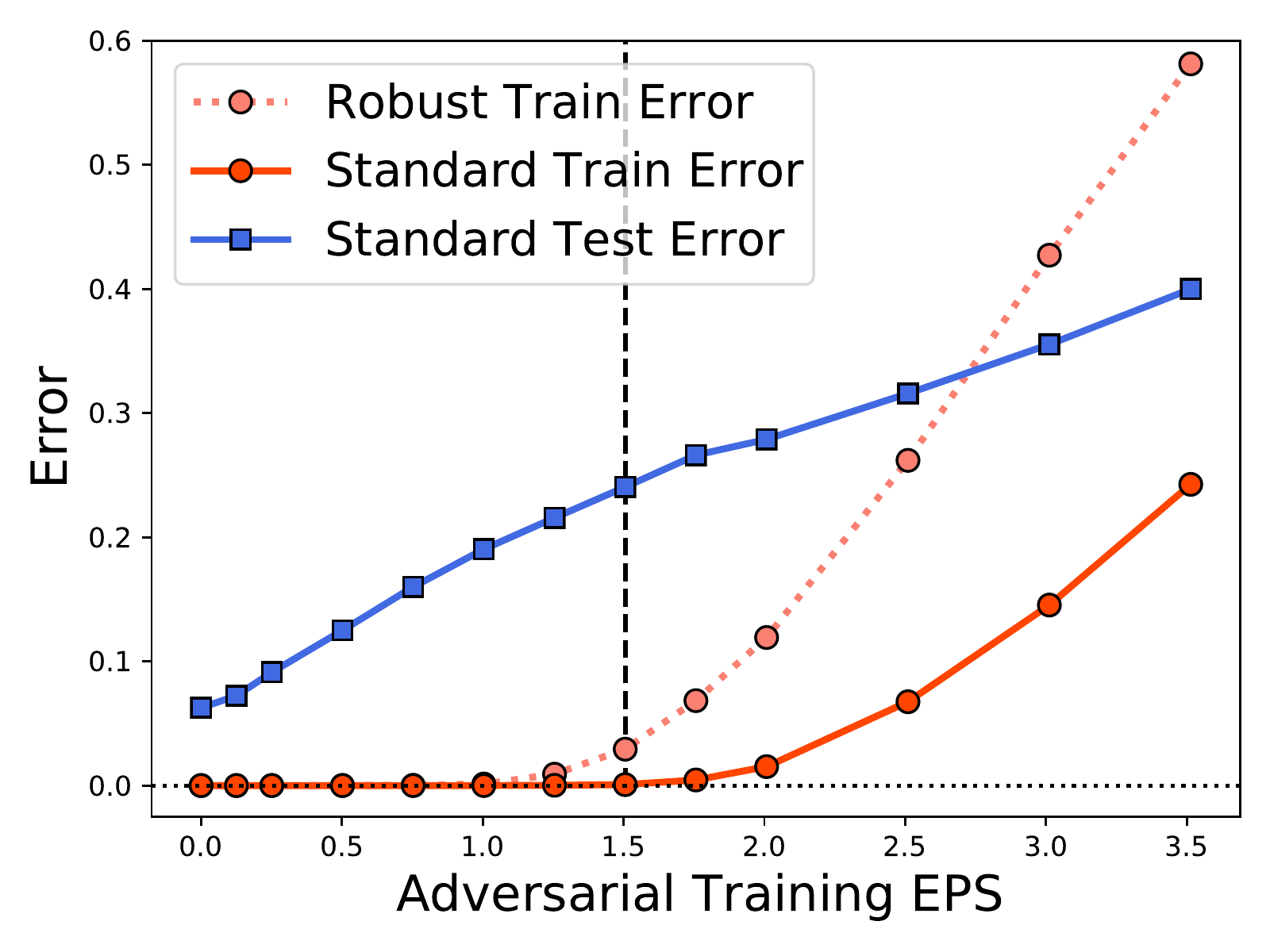}
    \caption{Measuring performance for $(\ell_{2}=\varepsilon)$-adversarial training (with increasing perturbation size) on \textit{CIFAR10} dataset. 
    (\textbf{left}) Evaluating bias, variance, and risk  for the $\ell_{2}$-adversarially trained models (WideResNet-28-10) on the \textit{CIFAR10} dataset.
    (\textbf{right}) Evaluating robust training error, standard training/test error for the $\ell_{2}$-adversarially trained models (WideResNet-28-10) on the \textit{CIFAR10} dataset. 
    }
  \label{fig:adv-l2-bvr-last-epoch-appendix}
  \end{center}
\end{figure*}

\vspace*{-0.75em}
\paragraph{$\ell_{2}$ adversarial training on CIFAR10.} We first study the bias-variance decomposition of $\ell_{2}$ adversarially trained models on the CIFAR10 dataset. We summarize the results in Figure~\ref{fig:adv-l2-bvr-last-epoch-appendix}. The bias and variance of $\ell_{2}$ adversarially trained models are similar to the ones in $\ell_{\infty}$ adversarial training~(in Figure~\ref{fig:bvr-mainline}). We find that the variance is unimodal and the bias is monotonically increasing with the perturbation size. Also, the variance peak  is near the robust interpolation threshold. In summary, we observe properties P1-P4 for $\ell_{2}$ adversarial training on CIFAR10.

\begin{figure*}[ht]
  \begin{center}
    \includegraphics[width=.31\textwidth]{tex_files/Figures/cifar100_at_linf_training_bvr_epoch200.pdf}
    \includegraphics[width=.31\textwidth]{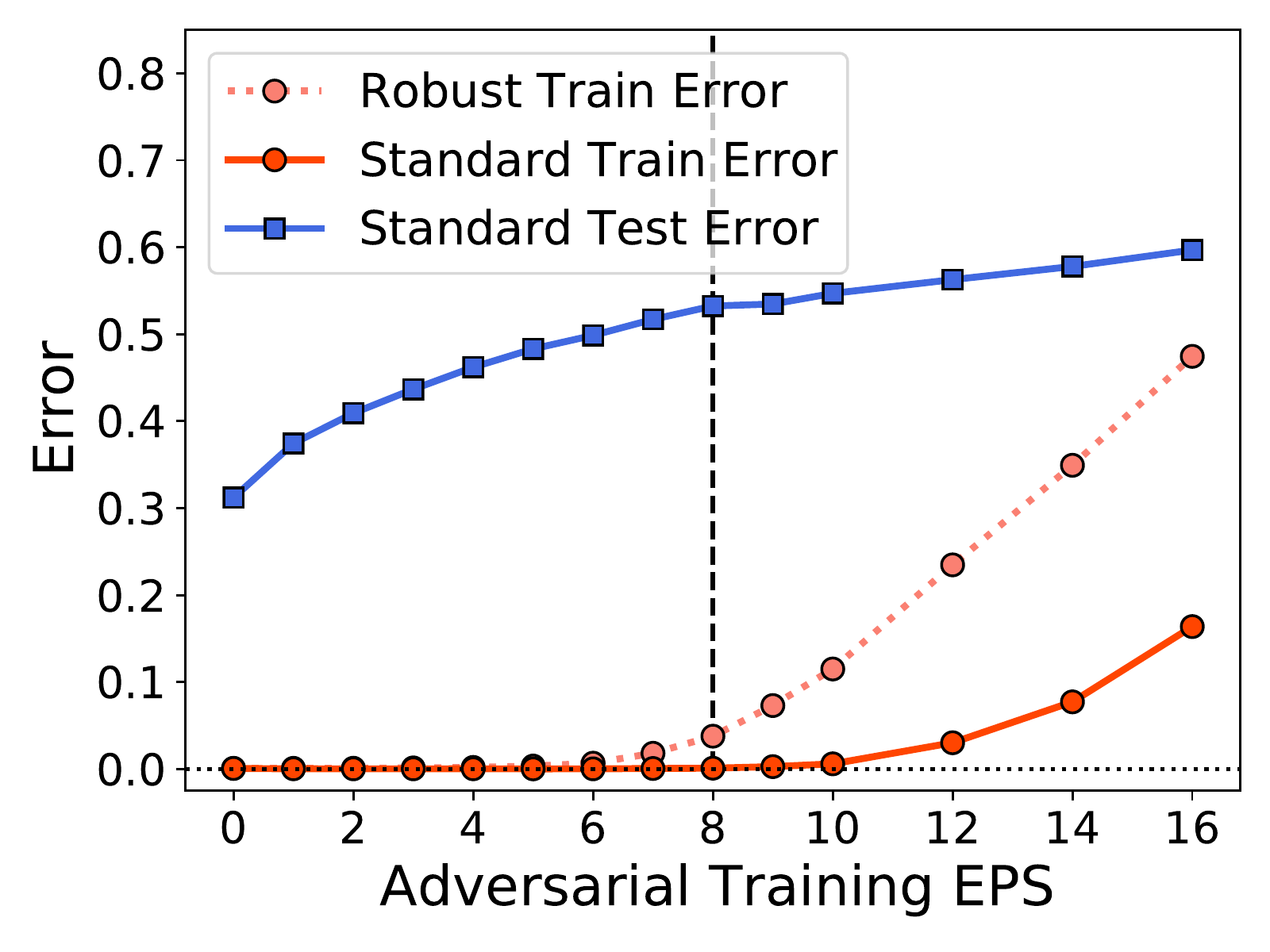}
    \caption{Measuring performance for $(\ell_{\infty}=\varepsilon/255.0)$-adversarial training (with increasing perturbation size) on \textit{CIFAR100} dataset.
    (\textbf{left}) Evaluating bias, variance, and risk  for the $\ell_{\infty}$-adversarially trained models (PreResNet-18) on the \textit{CIFAR100} dataset.
    (\textbf{right}) Evaluating robust training error, standard training/test error for the $\ell_{\infty}$-adversarially trained models (PreResNet-18) on the \textit{CIFAR100} dataset. 
    }
  \label{fig:adv-linf-bvr-last-epoch-cifar100-appendix}
  \end{center}
  \vskip -0.1in
\end{figure*}

\begin{figure*}[ht!]
  \begin{center}
    \includegraphics[width=.31\textwidth]{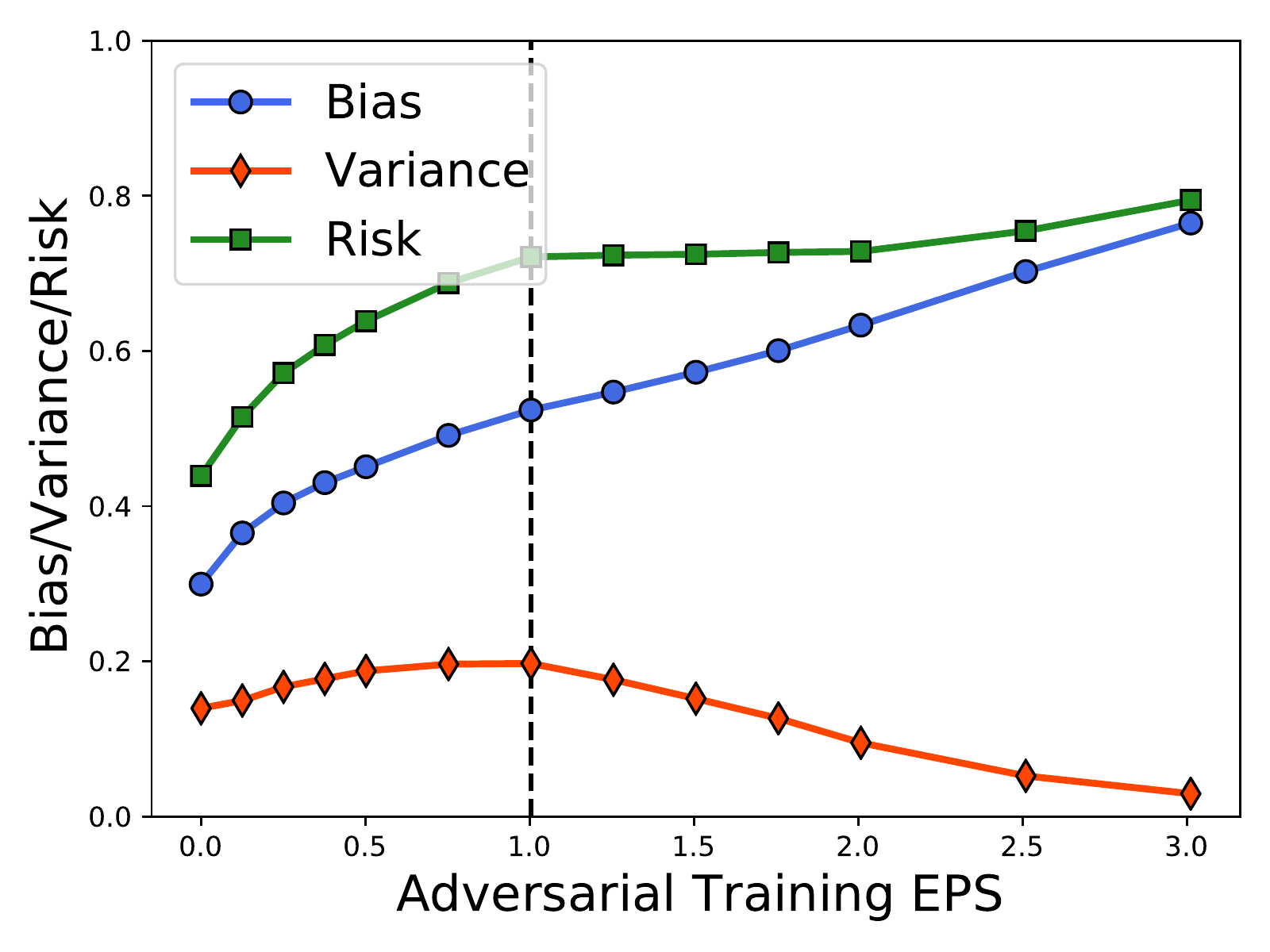}
    \includegraphics[width=.31\textwidth]{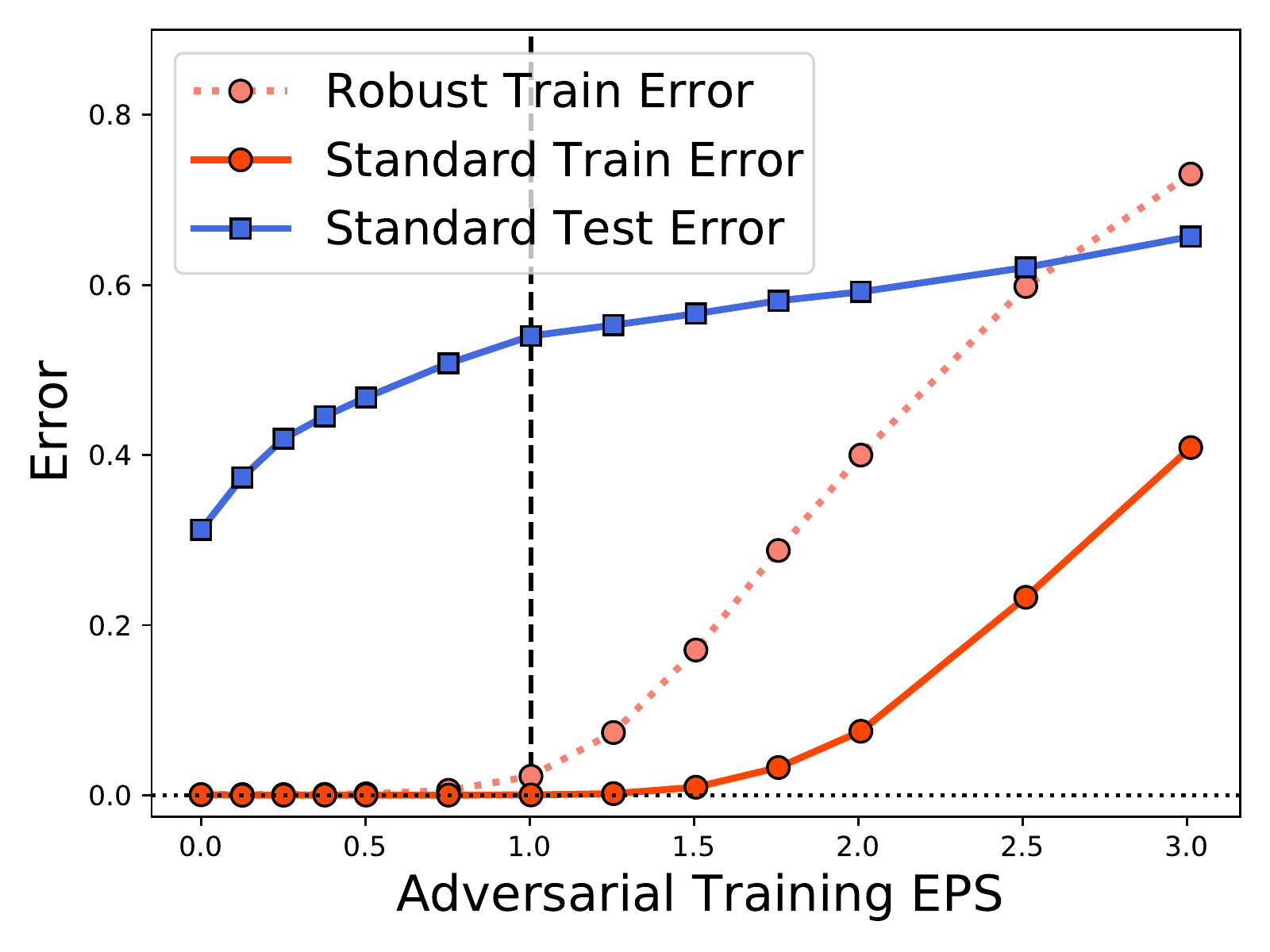}
    \caption{Measuring performance for $(\ell_{2}=\varepsilon)$-adversarial training (with increasing perturbation size) on \textit{CIFAR100} dataset. 
    (\textbf{left}) Evaluating bias, variance, and risk  for the $\ell_{2}$-adversarially trained models (PreResNet-18) on the \textit{CIFAR100} dataset.
    (\textbf{right}) Evaluating robust training error, standard training/test error for the $\ell_{2}$-adversarially trained models (PreResNet-18) on the \textit{CIFAR100} dataset. 
    }
  \label{fig:adv-l2-bvr-last-epoch-cifar100-appendix}
  \end{center}
\end{figure*}

\vspace*{-0.75em}
\paragraph{Results on CIFAR100.} In Figure~\ref{fig:adv-linf-bvr-last-epoch-cifar100-appendix} and Figure~\ref{fig:adv-l2-bvr-last-epoch-cifar100-appendix}, we study the bias-variance behavior on the CIFAR100 dataset. Following \cite{rice2020overfitting}, we use pre-activation ResNet18 (PreResNet-18) architecture for the CIFAR100 dataset and apply the same training parameters as used for CIFAR10. We find that, on the CIFAR100 dataset, the variance is indeed unimodal, the bias  dominates the risk and monotonically increasing, and the peak of the variance is close to the robust interpolation threshold. 
Similar to the results on CIFAR10, we observe properties P1-P4 for $\ell_{2}$ and $\ell_{\infty}$ adversarial training on CIFAR100.

\begin{figure*}[ht!]
  \begin{center}
    \includegraphics[width=.31\textwidth]{tex_files/Figures/imagenet10_at_linf_training_bvr_epoch200.pdf}
    \includegraphics[width=.31\textwidth]{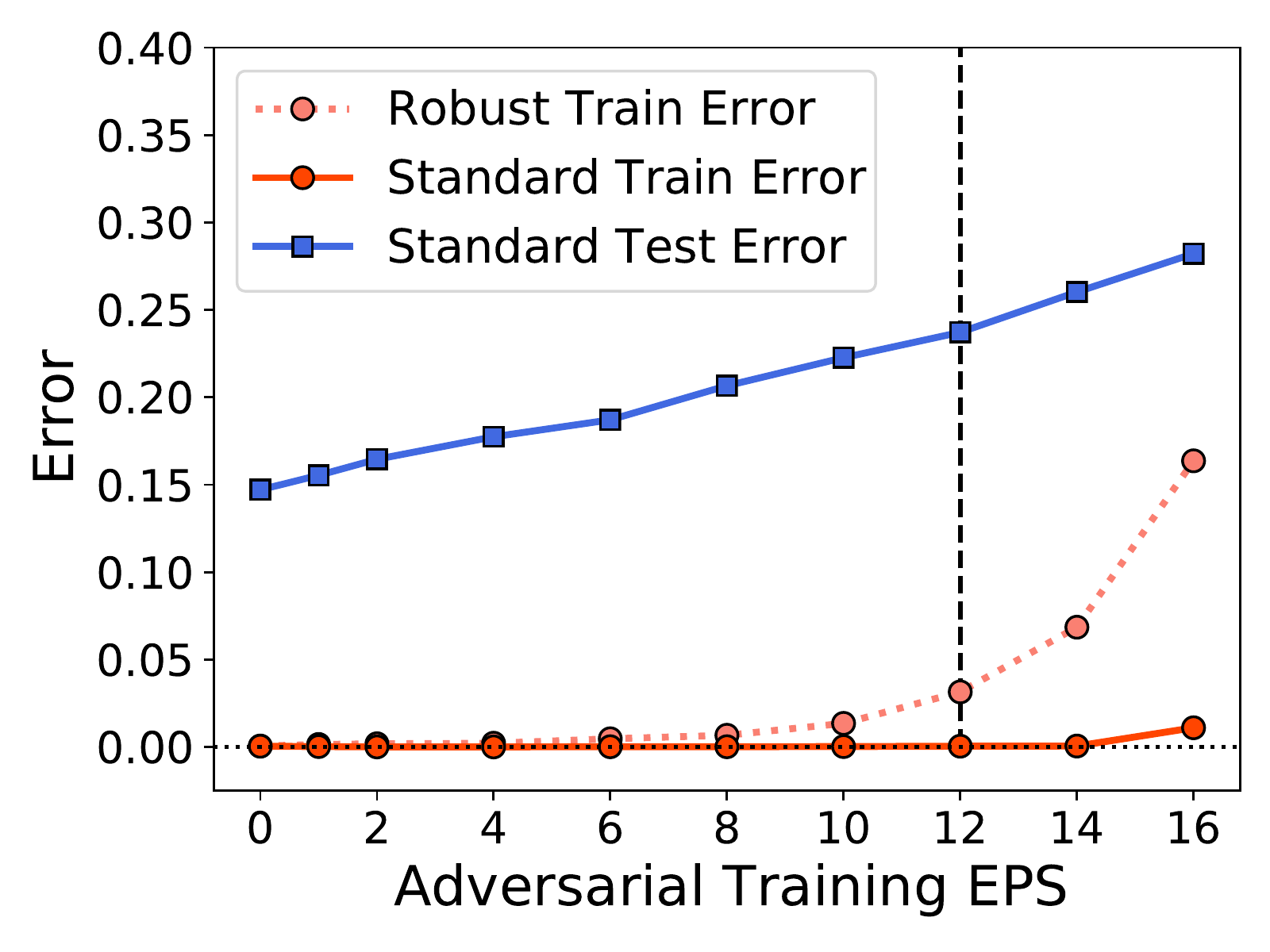}
    \caption{Measuring performance for $(\ell_{\infty}=\varepsilon/255.0)$-adversarial training (with increasing perturbation size) on \textit{ImageNet10} dataset. 
    (\textbf{left}) Evaluating bias, variance, and risk  for the $\ell_{\infty}$-adversarially trained models (ResNet-18) on the \textit{ImageNet10} dataset.
    (\textbf{right}) Evaluating robust training error, standard training/test error for the $\ell_{\infty}$-adversarially trained models (ResNet-18) on the \textit{ImageNet10} dataset. 
    }
  \label{fig:adv-l2-bvr-last-epoch-imagenet10-appendix}
  \end{center}
\end{figure*}

\vspace*{-0.75em}
\paragraph{Results on ImageNet10.} In Figure~\ref{fig:adv-l2-bvr-last-epoch-imagenet10-appendix}, we study the bias-variance behavior on the ImageNet10 dataset. We use the standard ResNet18 architecture  for the ImageNet10 dataset. We apply the hyperparameters for adversarial training as in the CIFAR10/100 dataset. As shown in Figure~\ref{fig:adv-l2-bvr-last-epoch-imagenet10-appendix}, We find that the variance is unimodal and the bias is monotonically increasing dominates the risk and monotonically increasing, and the peak of the variance is close to the robust interpolation threshold.  In summary, we observe properties P1-P4 for $\ell_{\infty}$ adversarial training on ImageNet10.

\begin{figure*}[ht!]
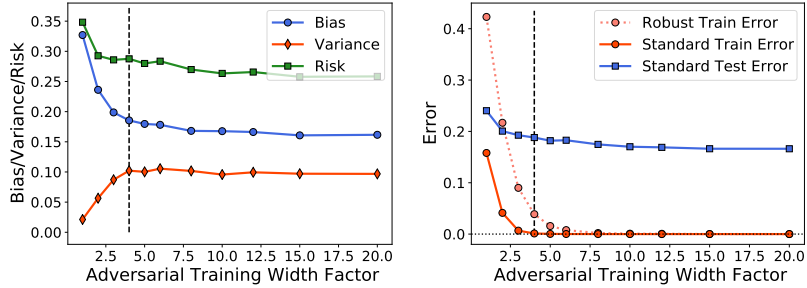

  \begin{center}
    \includegraphics[width=.31\textwidth]{tex_files/Figures/width_at_linf_training_bvr_epoch200.pdf}
    \includegraphics[width=.31\textwidth]{tex_files/Figures/width_at_linf_training_error_epoch200.pdf}
    \caption{Measuring performance for $(\ell_{\infty} = 8.0 / 255.0)$-adversarial training (with increasing width factor for WideResNet-28-[\texttt{width}]) on \textit{CIFAR10} dataset.
    (\textbf{left}) Evaluating bias, variance, and risk  for the $(\ell_{\infty} = 8.0 / 255.0)$-adversarially trained models (WideResNet-28-[\texttt{width}]) on the \textit{CIFAR10} dataset.
    (\textbf{right}) Evaluating robust training error, standard training/test error for the $\ell_{\infty}$-adversarially trained models (WideResNet-28-[\texttt{width}]) on the \textit{CIFAR10} dataset. 
    }
  \label{fig:width-adv-linf-bvr-last-epoch-cifar10-appendix}
  \end{center}
\end{figure*}

\vspace*{-0.75em}
\paragraph{Effect of width factor.} We study the effect of the width factor of the WideResNet-28 in $(\ell=8.0/255.0)$ adversarial training on the CIFAR10 dataset. We apply the same training parameters as mentioned in Section~\ref{sec:adv-training-on-2d-cifar10} and only change the wide factor of WideResNet-28 from 1 to 20. As shown in Figure~\ref{fig:width-adv-linf-bvr-last-epoch-cifar10-appendix}, the variance changes monotonicity at the interpolation threshold (\texttt{width}=4). We observe that the bias is monotonically decreasing with the width factor. The variance is increasing before the robust interpolation threshold. After the threshold, the variance value does not change too much.

\begin{figure*}[ht]
  \begin{center}
    \subfigure{
    \includegraphics[width=.31\textwidth]{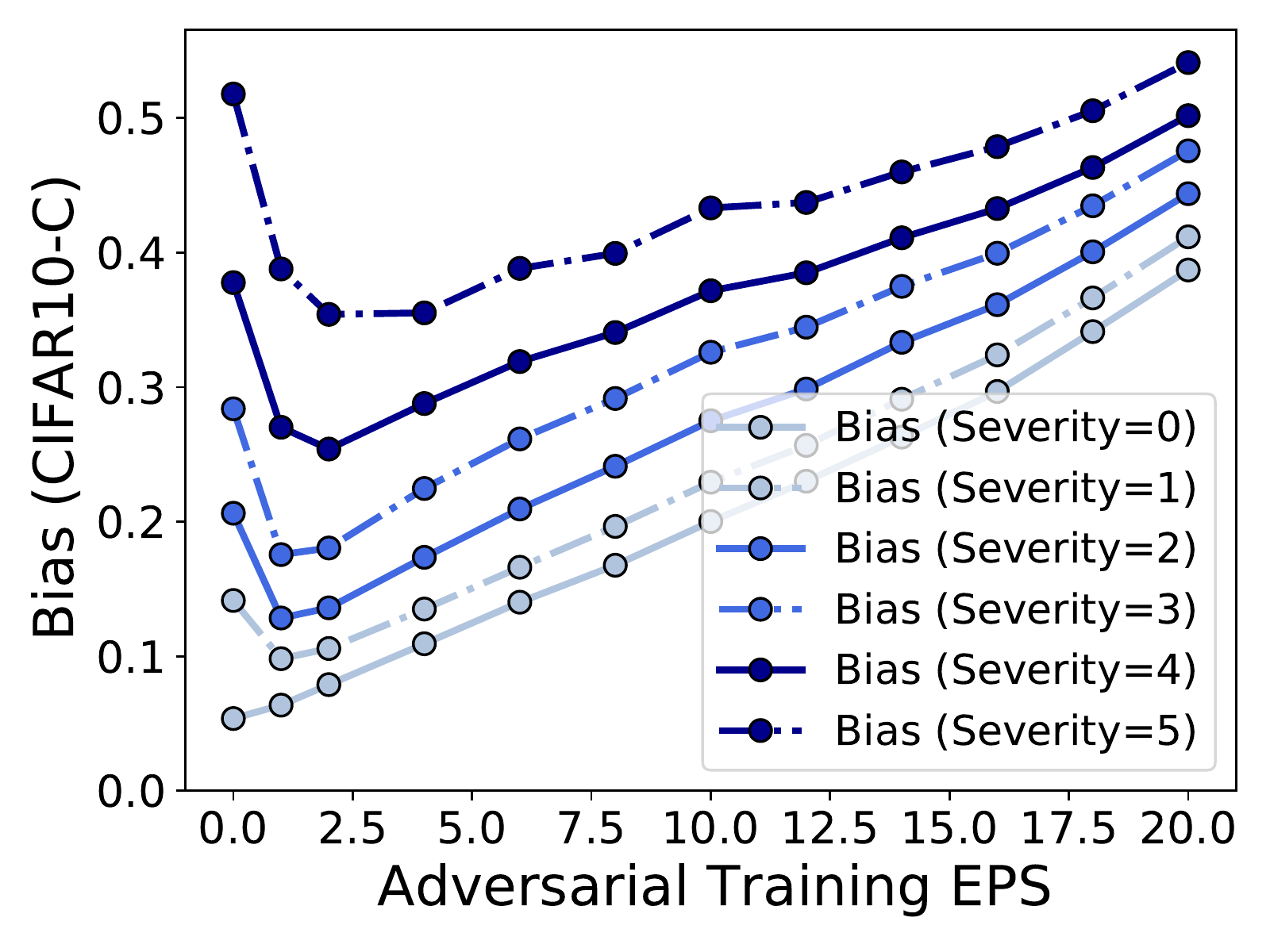}
    }
    \subfigure{
    \includegraphics[width=.31\textwidth]{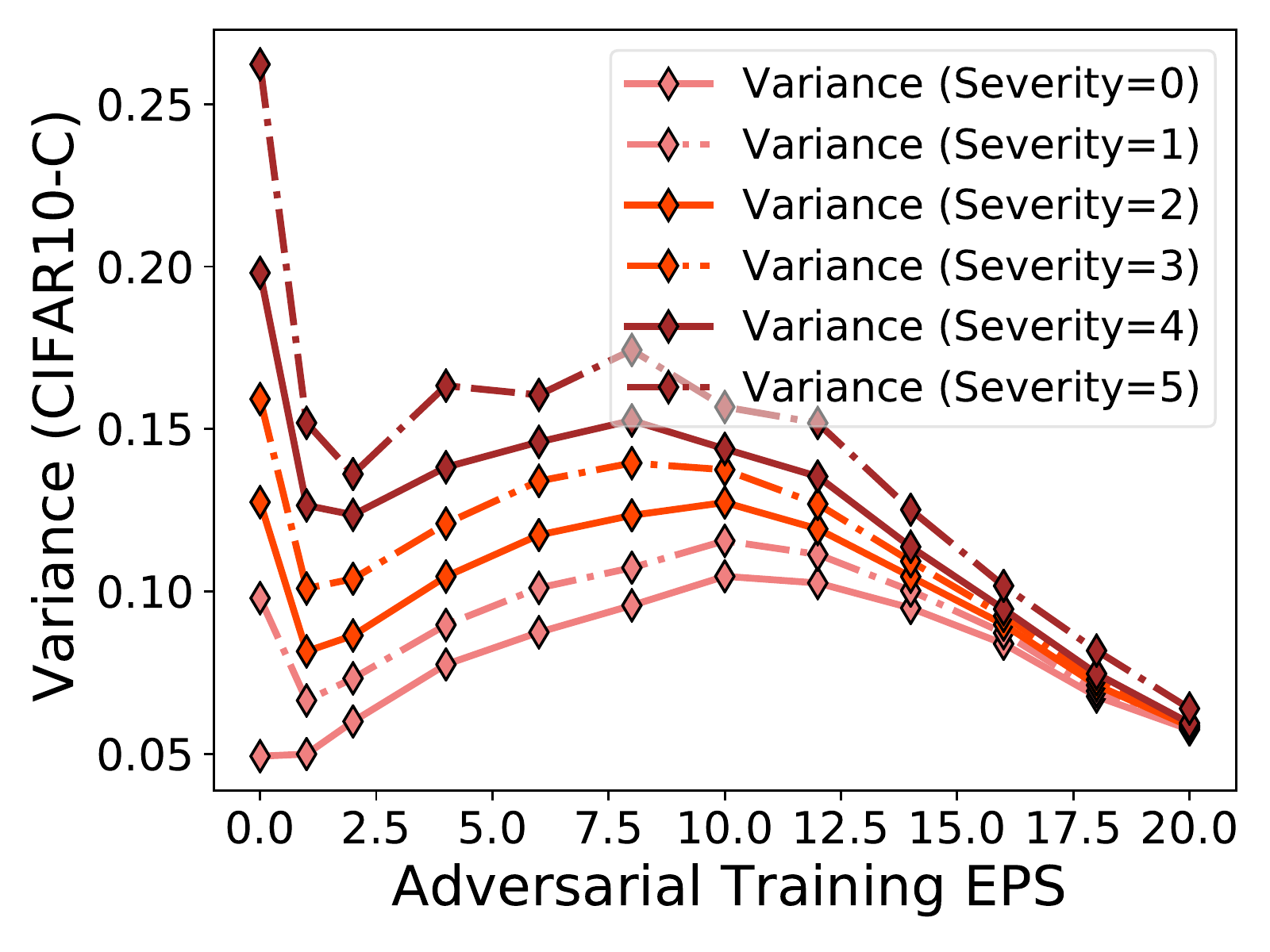}
    }
    \subfigure{
    \includegraphics[width=.31\textwidth]{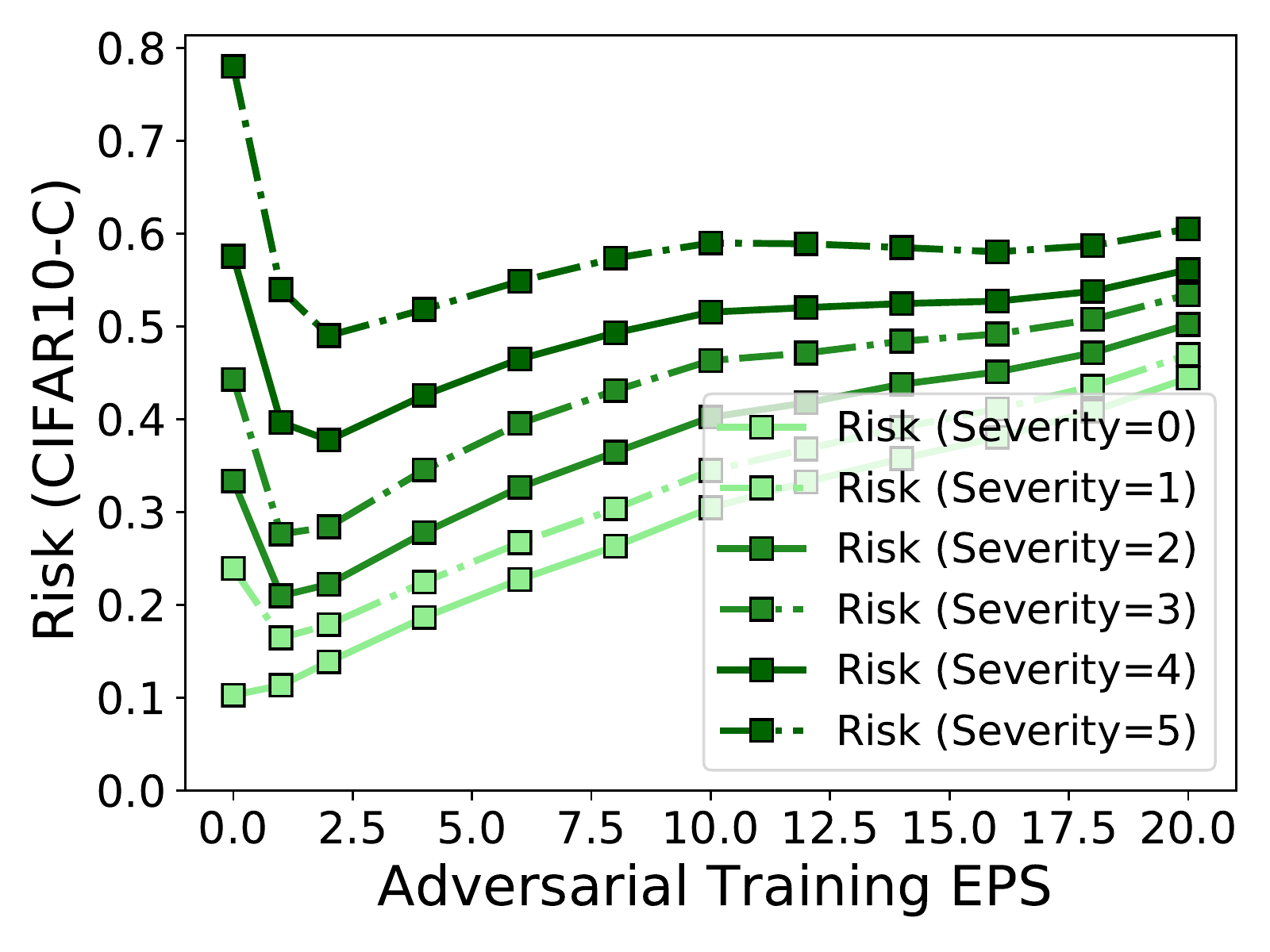}
    }
    \vskip -0.1in
    \caption{Bias, variance, and risk for $\ell_{\infty}$-adversarial training models evaluated on the CIFAR10-C dataset with different severity. Each curve corresponds to one level of severity, and \textsf{severity}=0 corresponds to the standard test CIFAR10 testset. (\textbf{Left}) Bias. (\textbf{Middle}) Variance. (\textbf{Right}) Risk. 
    }
  \label{fig:bvr-AT-cifar10c-epoch200}
  \end{center}
  \vskip -0.05in
\end{figure*}

\newpage
\subsection{Measuring Bias-variance of Adversarial Training Models on OOD datasets.}\label{sec:appendix-additional-exp-results-c3}

\paragraph{Adversarial training evaluated on OOD data.} We evaluate the bias, variance, and risk of the adversarially trained models with various $\ell_{\infty}$ perturbation radii on the CIFAR10-C dataset. CIFAR10-C contains 15 different common corruptions with five levels of severity, where samples with larger severity level are generally harder to classify. We evaluate the bias and variance on test samples with various severity levels. The results are summarized in Figure~\ref{fig:bvr-AT-cifar10c-epoch200}.

We observe that the bias curve becomes U-shaped on CIFAR10-C and the variance curve changes to a ``double-descent-shaped'' curve (with two dips) when evaluated on OOD data. Another intriguing observation is that when the adversarial perturbation radius $\varepsilon$ is small, both the bias and the variance (evaluated on OOD data) of the adversarially trained models are \textit{smaller} than for models with standard training ($\varepsilon = 0$). This suggests that $\ell_{\infty}$ adversarial training with small perturbation radii improves model performance (in terms of both bias and variance) on common corruptions, which aligns well with findings in \citet{kireev2021effectiveness}. It would be a valuable direction for future work to explore how to better trade off bias and variance of adversarially trained models for better OOD generalization.

We also study the bias-variance decomposition for $\ell_{\infty}$-adversarially trained models and AugMix~\cite{hendrycks2019augmix}-trained models (on the CIFAR10 dataset) evaluated on datasets under distributional shift. This includes natural distributional shift, in the CIFAR10-v6 dataset~\cite{recht2018cifar}, and a common corruption benchmark, the CIFAR10-C dataset~\cite{hendrycks2019benchmarking}.

\begin{figure*}[ht]
  \begin{center}
    \subfigure{
    \includegraphics[width=.31\textwidth]{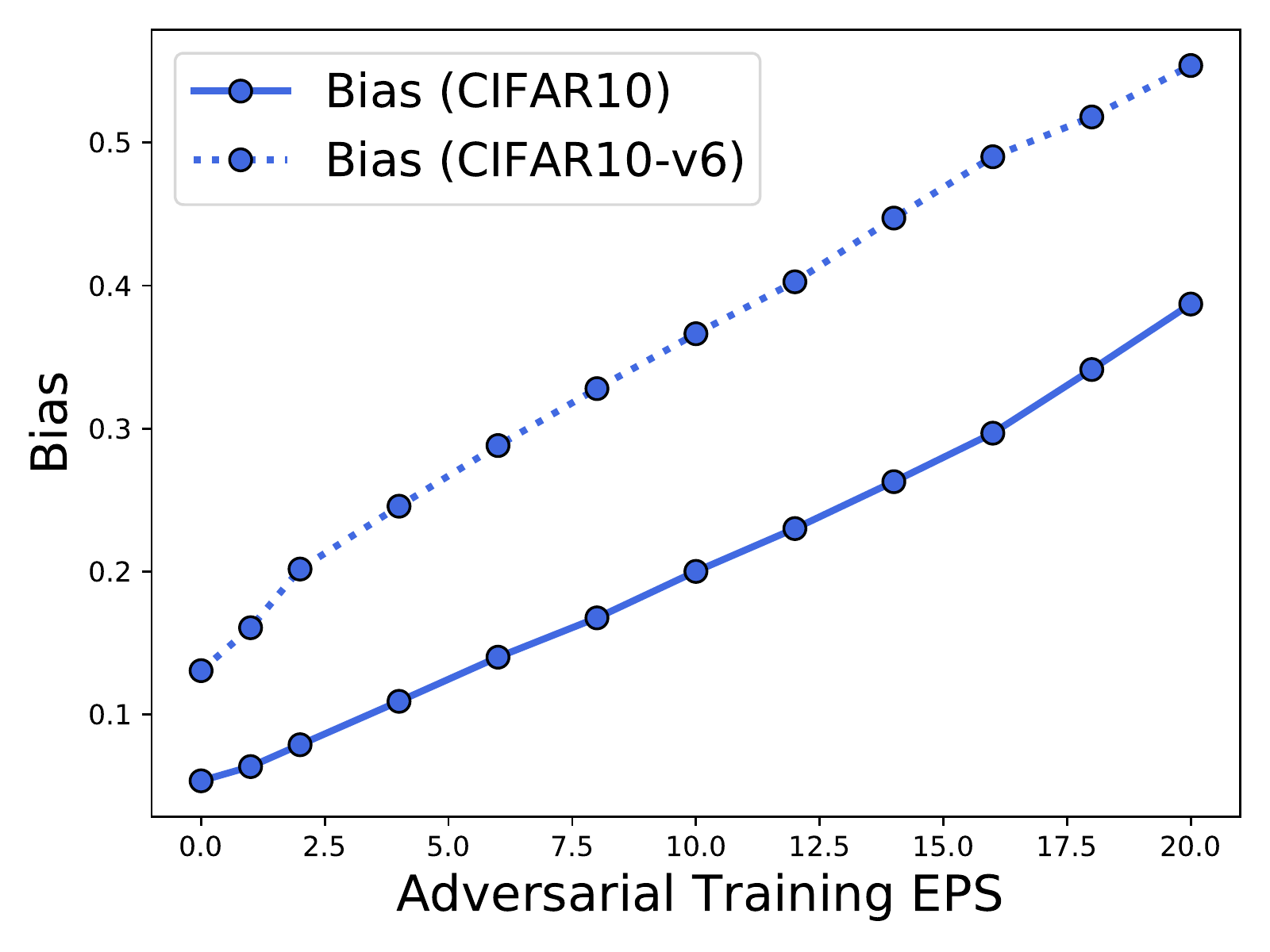}
    }
    \subfigure{
    \includegraphics[width=.31\textwidth]{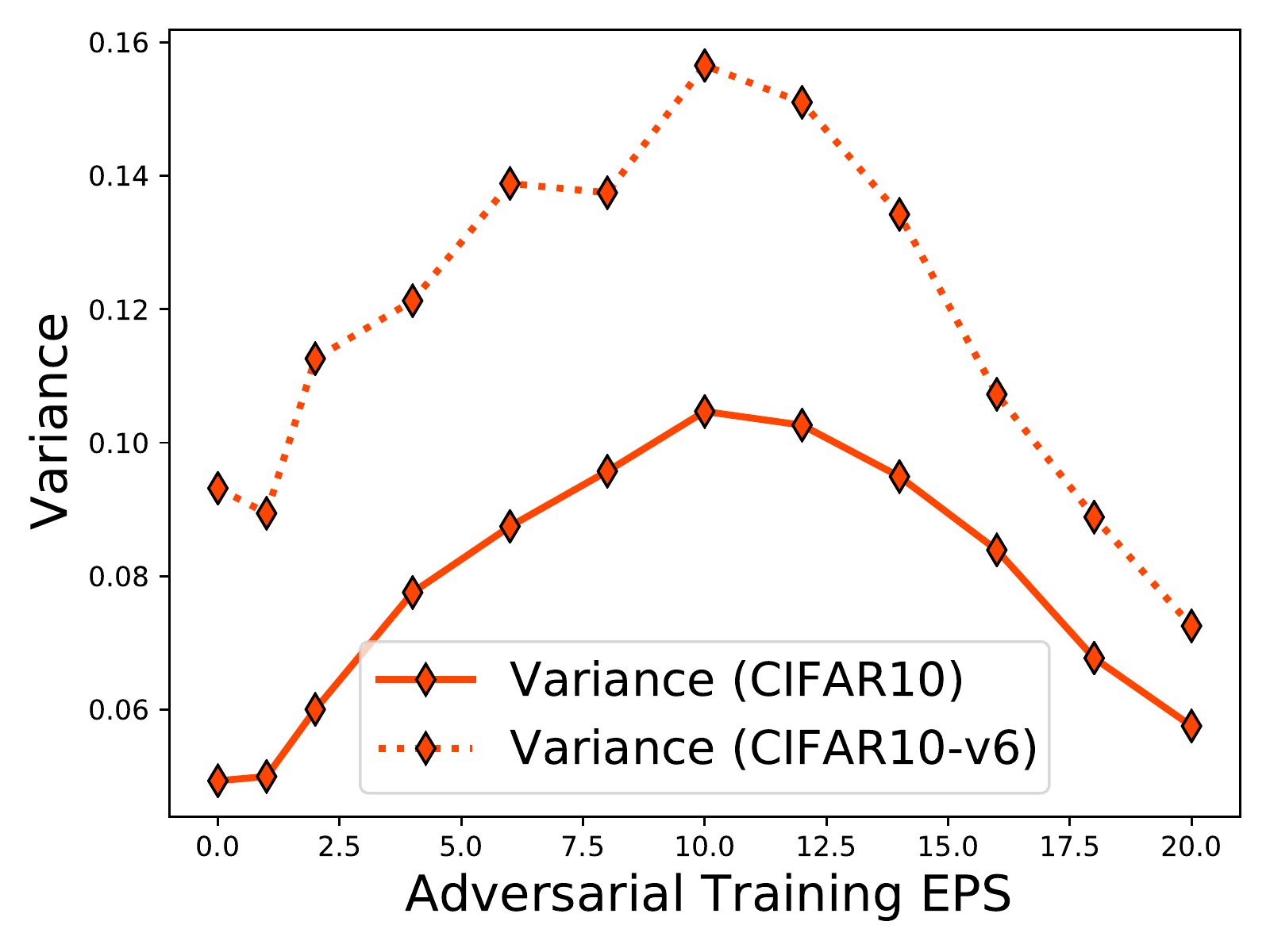}
    }
    \subfigure{
    \includegraphics[width=.31\textwidth]{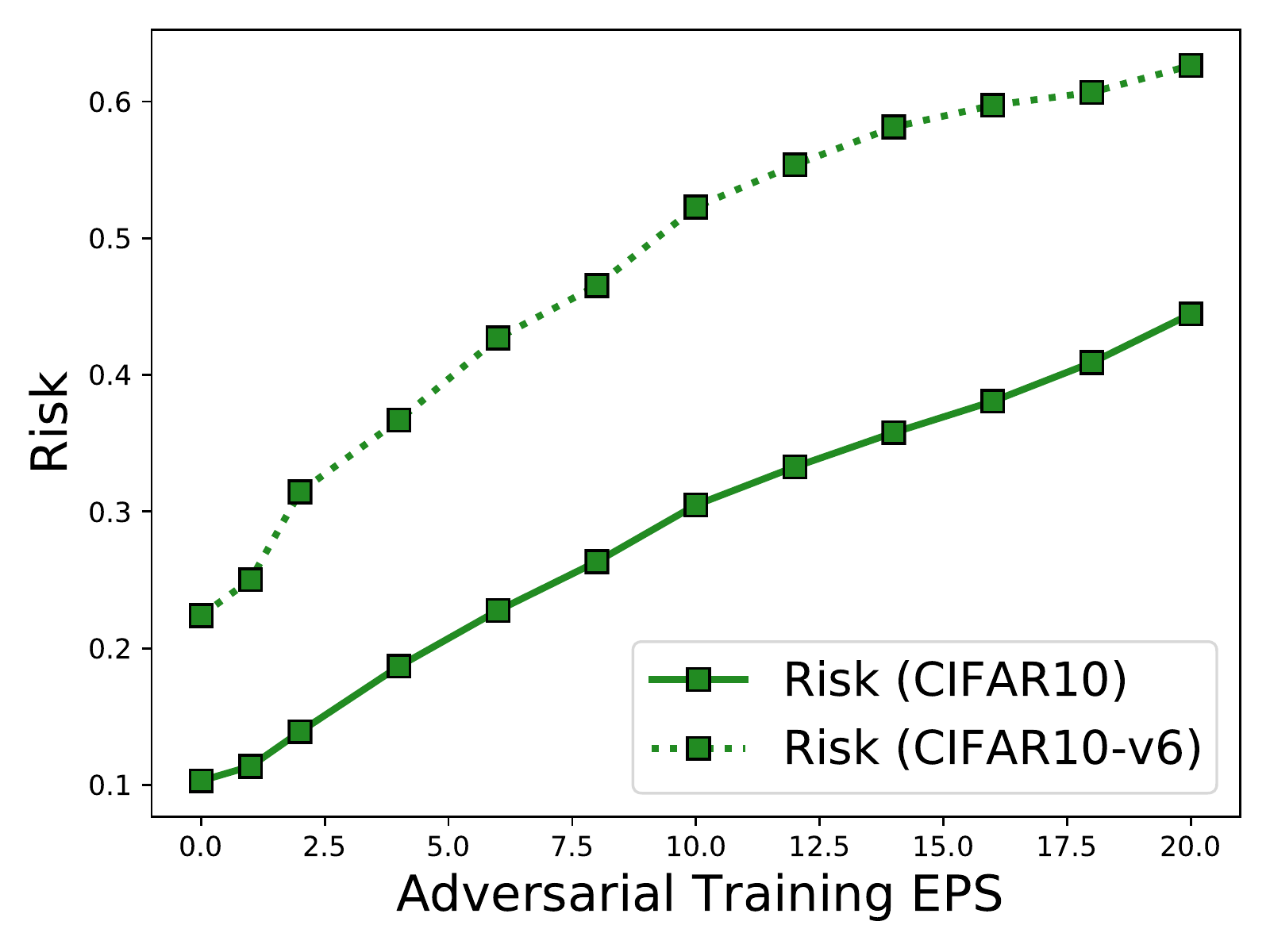}
    }
    \vskip -0.1in
    \caption{Compare bias, variance, and risk for $\ell_{\infty}$-adversarial training models evaluated on the CIFAR10-v6 and standard CIFAR10 testset. (\textbf{Left}) Bias. (\textbf{Middle}) Variance. (\textbf{Right}) Risk. 
    }
  \label{fig:bvr-AT-cifar10v6-epoch200}
  \end{center}
  \vskip -0.2in
\end{figure*}

\vspace*{-0.75em}
\paragraph{CIFAR10-v6 results.} CIFAR10-v6 contains new collected test images for the  CIFAR10 dataset. The results of  $\ell_{\infty}$-adversarial training on CIFAR10-v6 are summarized in Figure~\ref{fig:bvr-AT-cifar10v6-epoch200}. We find that the bias and variance on CIFAR10-v6 are similar to the standard test dataset case~(in Figure~\ref{fig:bvr-mainline}), whereas the variance decreases after $\varepsilon=10$. 

\begin{figure*}[ht]
  \begin{center}
    \subfigure{
    \includegraphics[width=.31\textwidth]{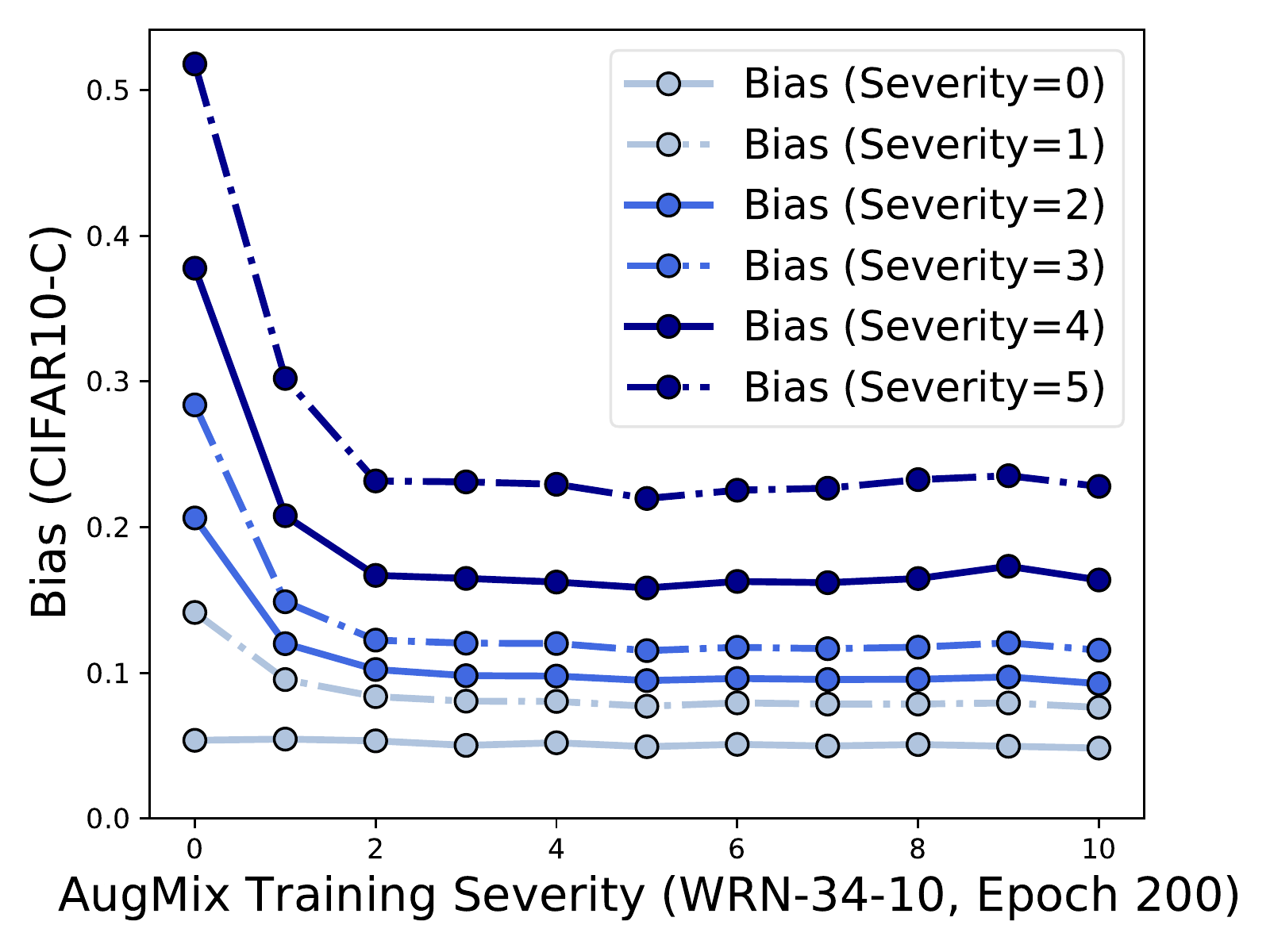}
    }
    \subfigure{
    \includegraphics[width=.31\textwidth]{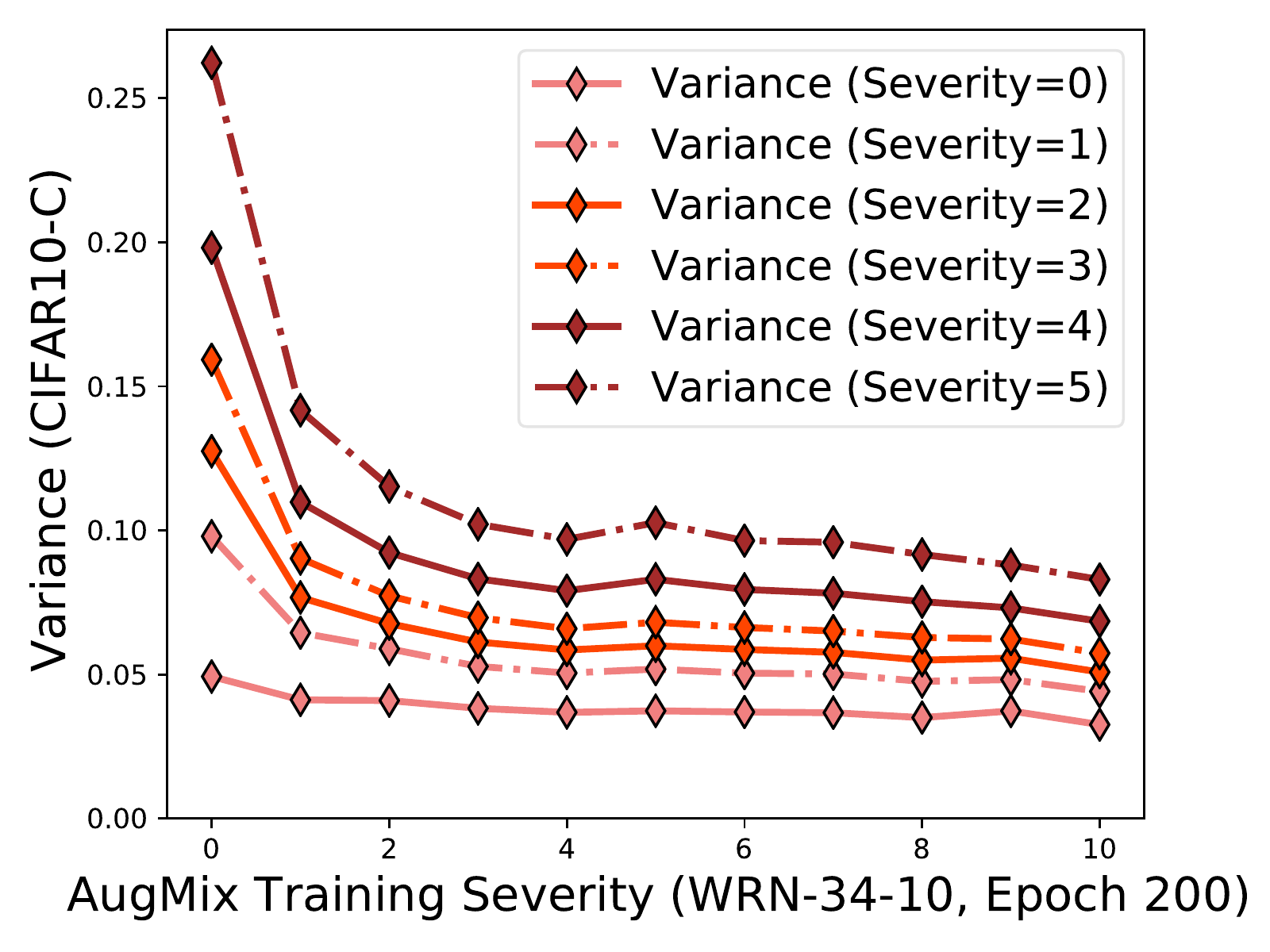}
    }
    \subfigure{
    \includegraphics[width=.31\textwidth]{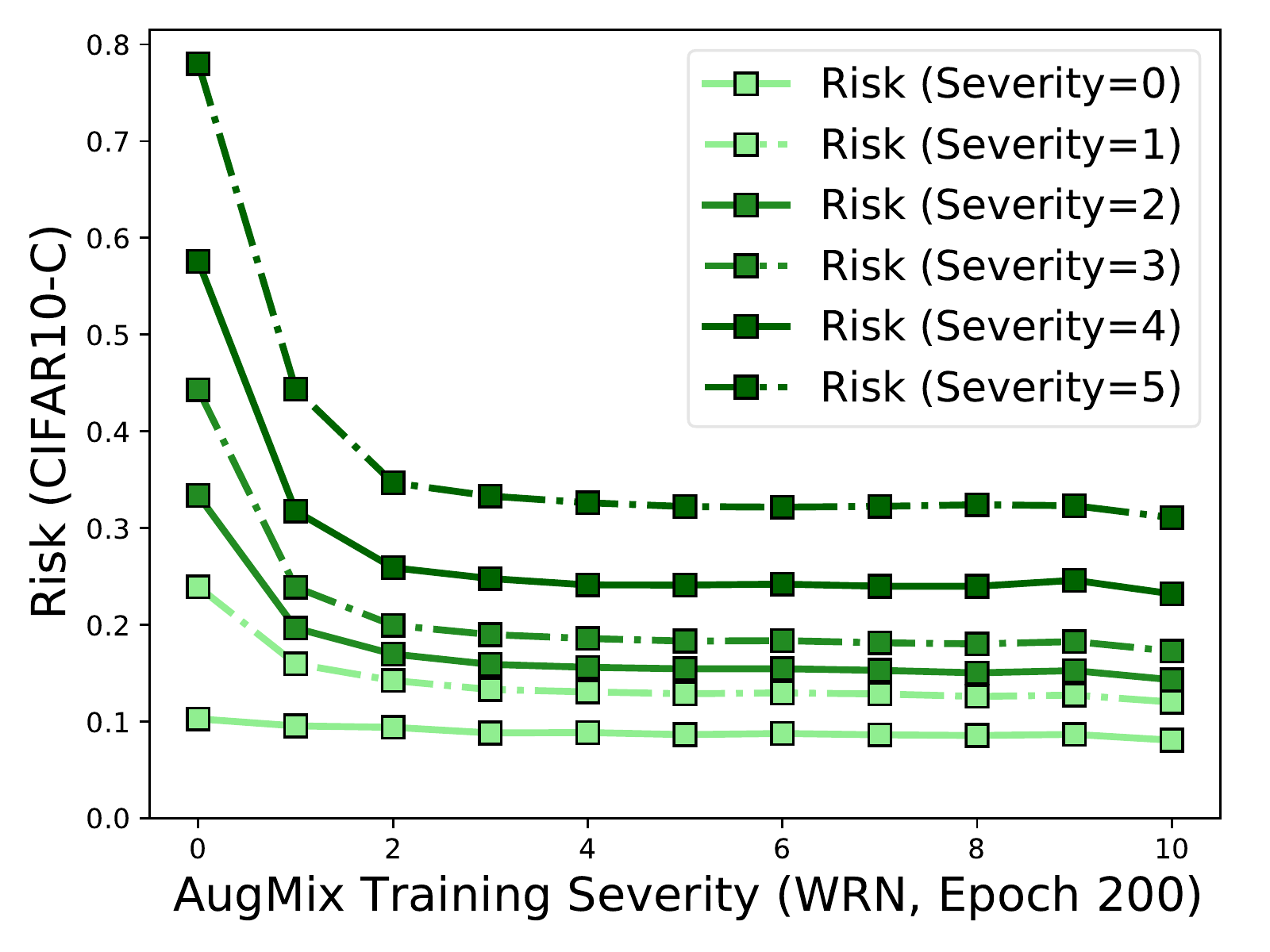}
    }
    \vskip -0.1in
    \caption{Bias, variance, and risk for AugMix training models (without applying JSD loss) evaluated on the CIFAR10-C dataset with different severity. Each curve corresponds to one level of severity, and \textsf{severity}=0 corresponds to the standard test CIFAR10 testset. (\textbf{Left}) Bias. (\textbf{Middle}) Variance. (\textbf{Right}) Risk. 
    }
  \label{fig:bvr-augmix-cifar10c-epoch200}
  \end{center}
  \vskip -0.2in
\end{figure*}

\vspace*{-0.75em}
\paragraph{AugMix training.} In addition, we study the bias-variance decomposition for AugMix training on the CIFAR10 dataset. We change the severity (augmentation severity parameter in AugMix) from 0 to 10, where 0 corresponds to the standard training and 10 is the maximum severity defined in AugMix. We summarize the bias-variance decomposition results in Figure~\ref{fig:bvr-augmix-cifar10c-epoch200}. For AugMix training, we find that bias, variance, and risk are monotonically decreasing with increased augmentation severity. The bias-variance behavior of AugMix training is different from $\ell_{\infty}$-adversarial training (in Figure~\ref{fig:bvr-AT-cifar10c-epoch200}).

\subsection{Measuring Bias-variance Decomposition for Cross-Entropy Loss.}\label{sec:appendix-additional-exp-results-c4}
\begin{figure*}[ht]
  \begin{center}
    \includegraphics[width=.31\textwidth]{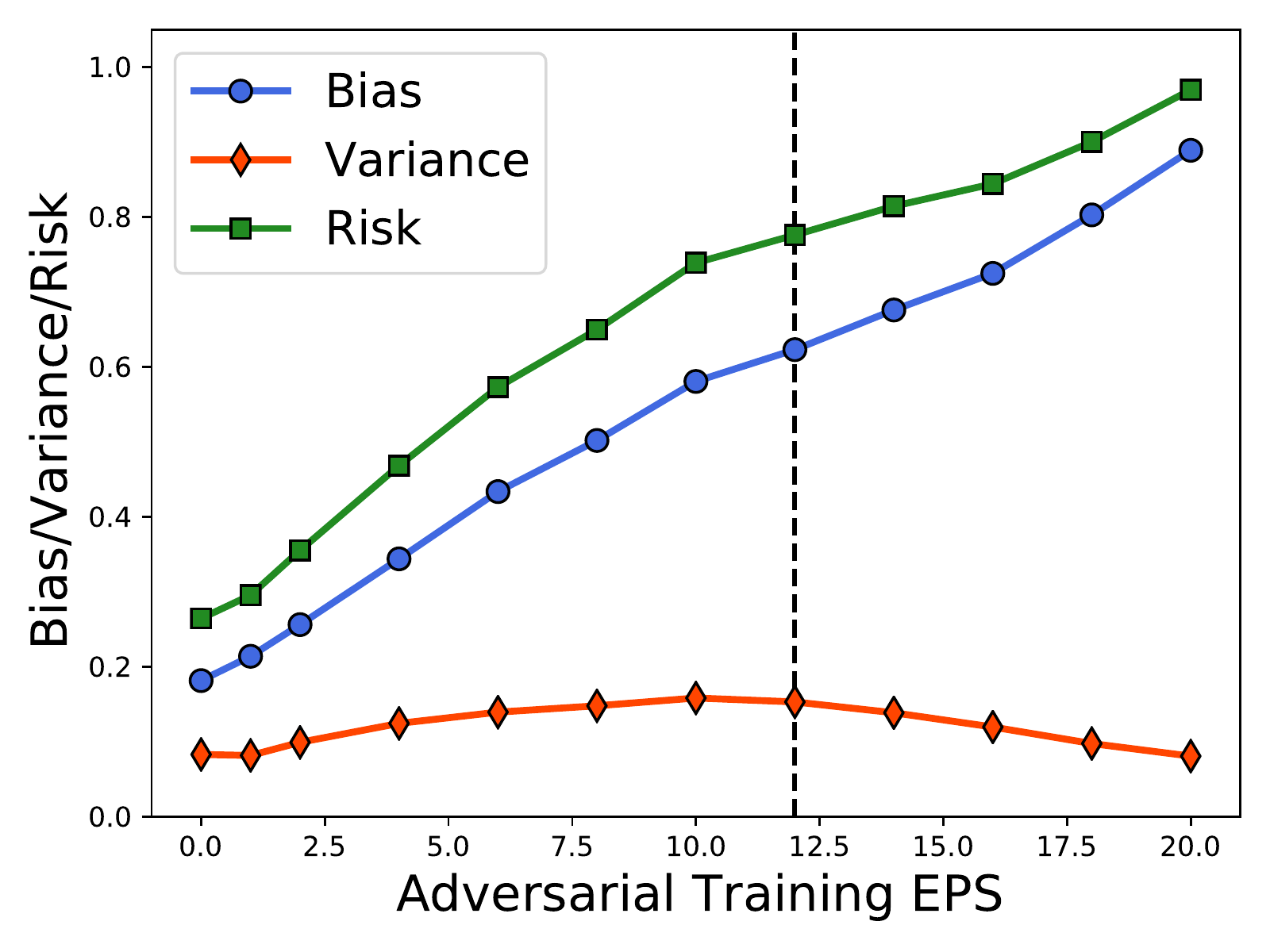}
    \includegraphics[width=.31\textwidth]{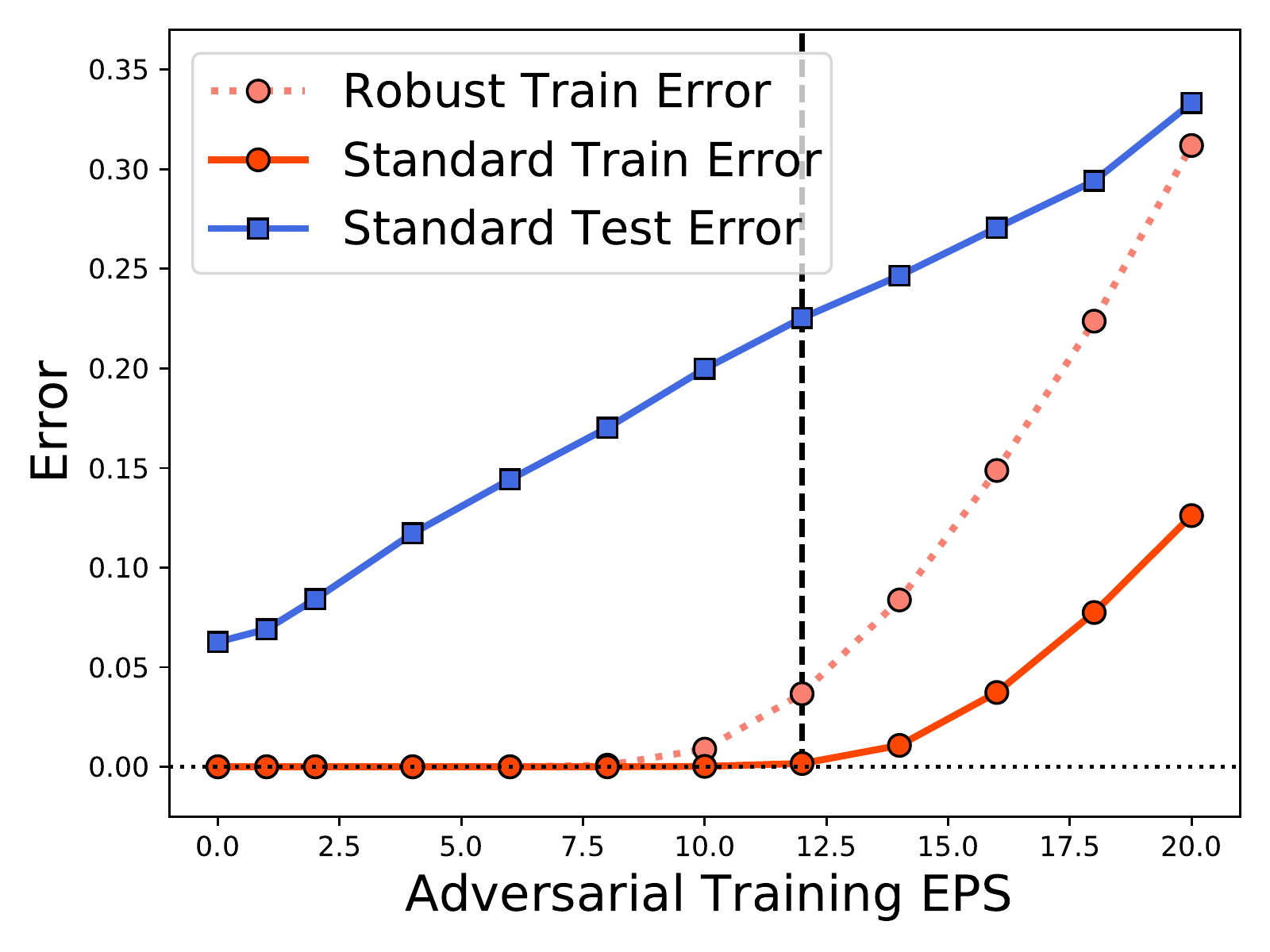}
    \caption{Measuring performance for $(\ell_{\infty}=\varepsilon/255.0)$-adversarial training (with increasing perturbation size) on \textit{CIFAR10} dataset.
    (\textbf{left}) Evaluating (\textit{cross-entropy loss}) bias, variance, and risk for the $\ell_{\infty}$-adversarially trained models (WideResNet-28-10) on the \textit{CIFAR10} dataset.
    (\textbf{right}) Evaluating robust training error, standard training/test error for the $\ell_{\infty}$-adversarially trained models (WideResNet-28-10) on the \textit{CIFAR10} dataset. 
    }
  \label{fig:adv-linf-bvr-kl-last-epoch-cifar100-appendix}
  \end{center}
  \vskip -0.1in
\end{figure*}

\begin{figure*}[ht!]
  \begin{center}
    \includegraphics[width=.31\textwidth]{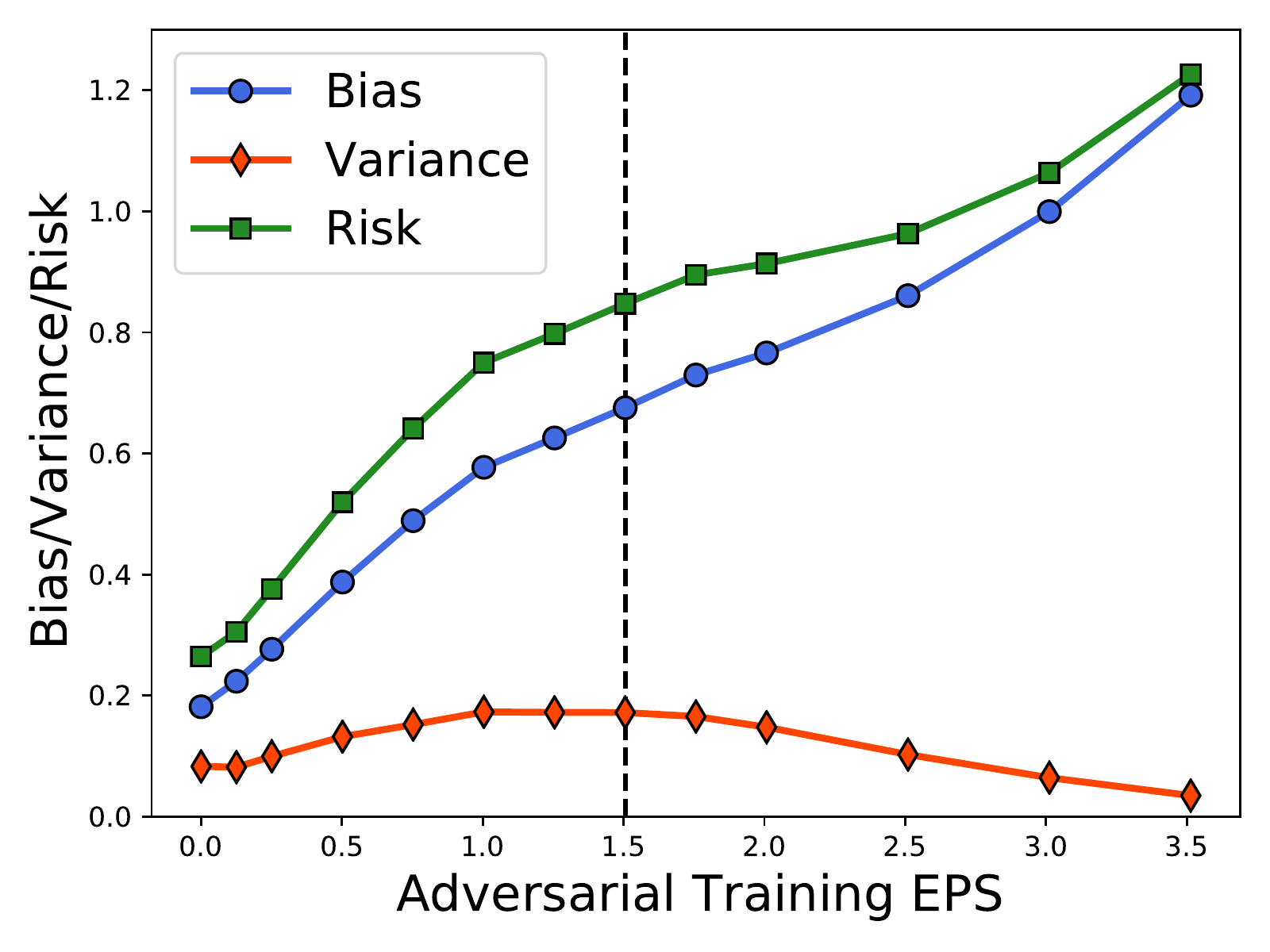}
    \includegraphics[width=.31\textwidth]{tex_files/Figures/at_l2_training_error_epoch200.pdf}
    \caption{Measuring performance for $(\ell_{2}=\varepsilon)$-adversarial training (with increasing perturbation size) on \textit{CIFAR10} dataset.
    (\textbf{left}) Evaluating (\textit{cross-entropy loss}) bias, variance, and risk  for the $\ell_{2}$-adversarially trained models (WideResNet-28-10) on the \textit{CIFAR10} dataset.
    (\textbf{right}) Evaluating robust training error, standard training/test error for the $\ell_{2}$-adversarially trained models (WideResNet-28-10) on the \textit{CIFAR10} dataset.
    }
  \label{fig:adv-l2-bvr-kl-last-epoch-cifar100-appendix}
  \end{center}
  \vskip -0.05in
\end{figure*}

We study the bias-variance decomposition for cross-entropy (CE) loss defined above. We follow Algorithm~1 in \cite{YYBV2020} to evaluate the bias and variance for the cross-entropy loss. Specifically, we evaluate the (\textit{cross-entropy loss}) bias-variance decomposition for $\ell_{\infty}$-adversarially trained models (as shown in Figure~\ref{fig:adv-linf-bvr-kl-last-epoch-cifar100-appendix}) and $\ell_{2}$-adversarially trained models (shown in Figure~\ref{fig:adv-l2-bvr-kl-last-epoch-cifar100-appendix}) on the CIFAR10 dataset. From Figure~\ref{fig:adv-linf-bvr-kl-last-epoch-cifar100-appendix} and Figure~\ref{fig:adv-l2-bvr-kl-last-epoch-cifar100-appendix}, we observe the unimodal variance curve and monotonically increasing bias curve for the (\textit{cross-entropy loss}) bias-variance decomposition. The peak of the (\textit{cross-entropy loss}) variance is also near the robust interpolation threshold.

\newpage
\section{Additional Experiments on Simplified Models}\label{sec:appendix-toy-model}
In this section, we present additional experiments on 2D box example and adversarial logistic regression as described in Section~\ref{sec:testing_toy}.

\begin{figure*}[ht]
\vskip -0.1in
\begin{center}
\subfigure[$d=2$]{
\includegraphics[width=.26\textwidth]{tex_files/Figures/bv_dim2_trial30.pdf}
}
\subfigure[$d=3$]{
\includegraphics[width=.26\textwidth]{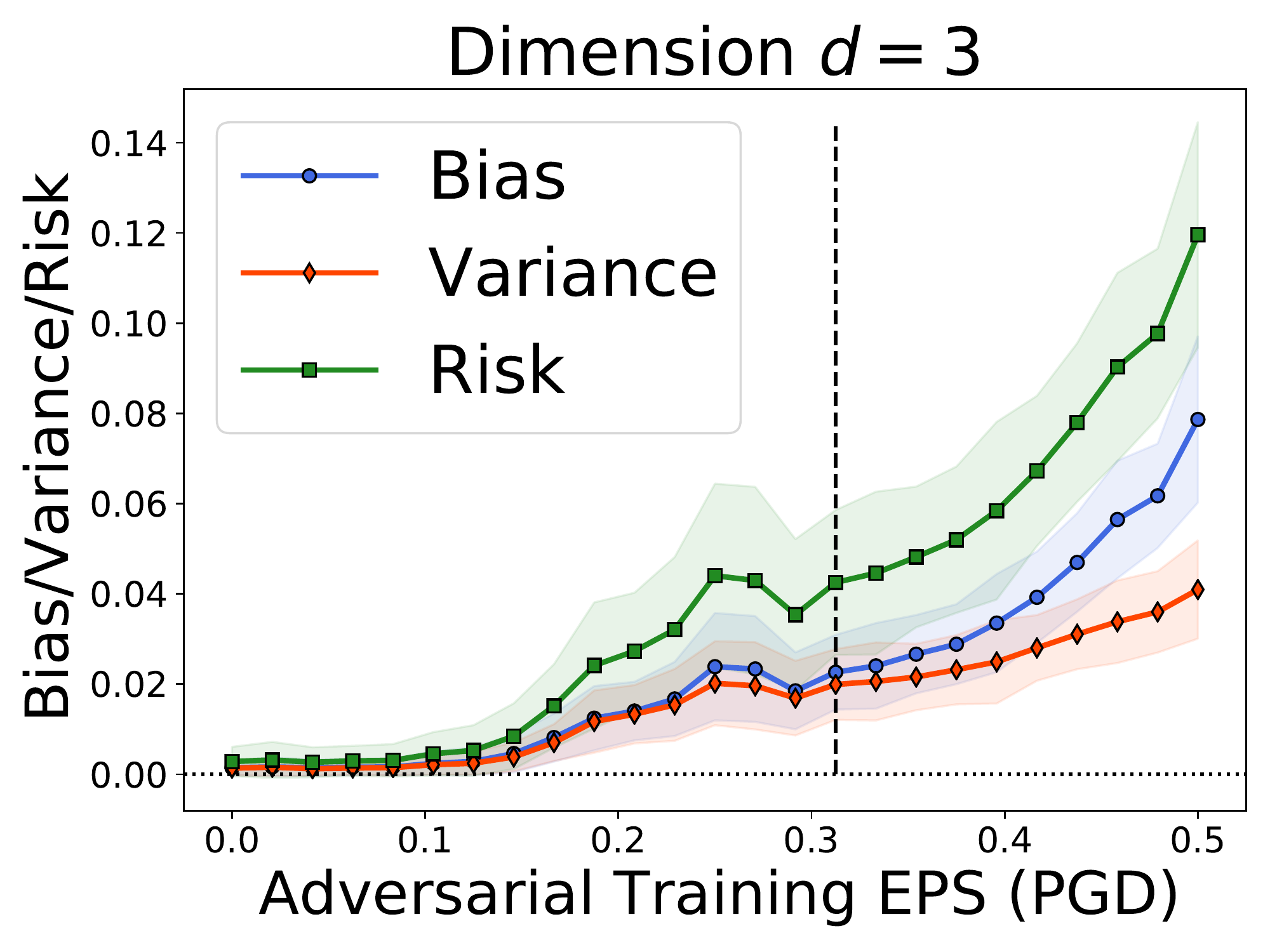}
}
\subfigure[$d=4$]{
\includegraphics[width=.26\textwidth]{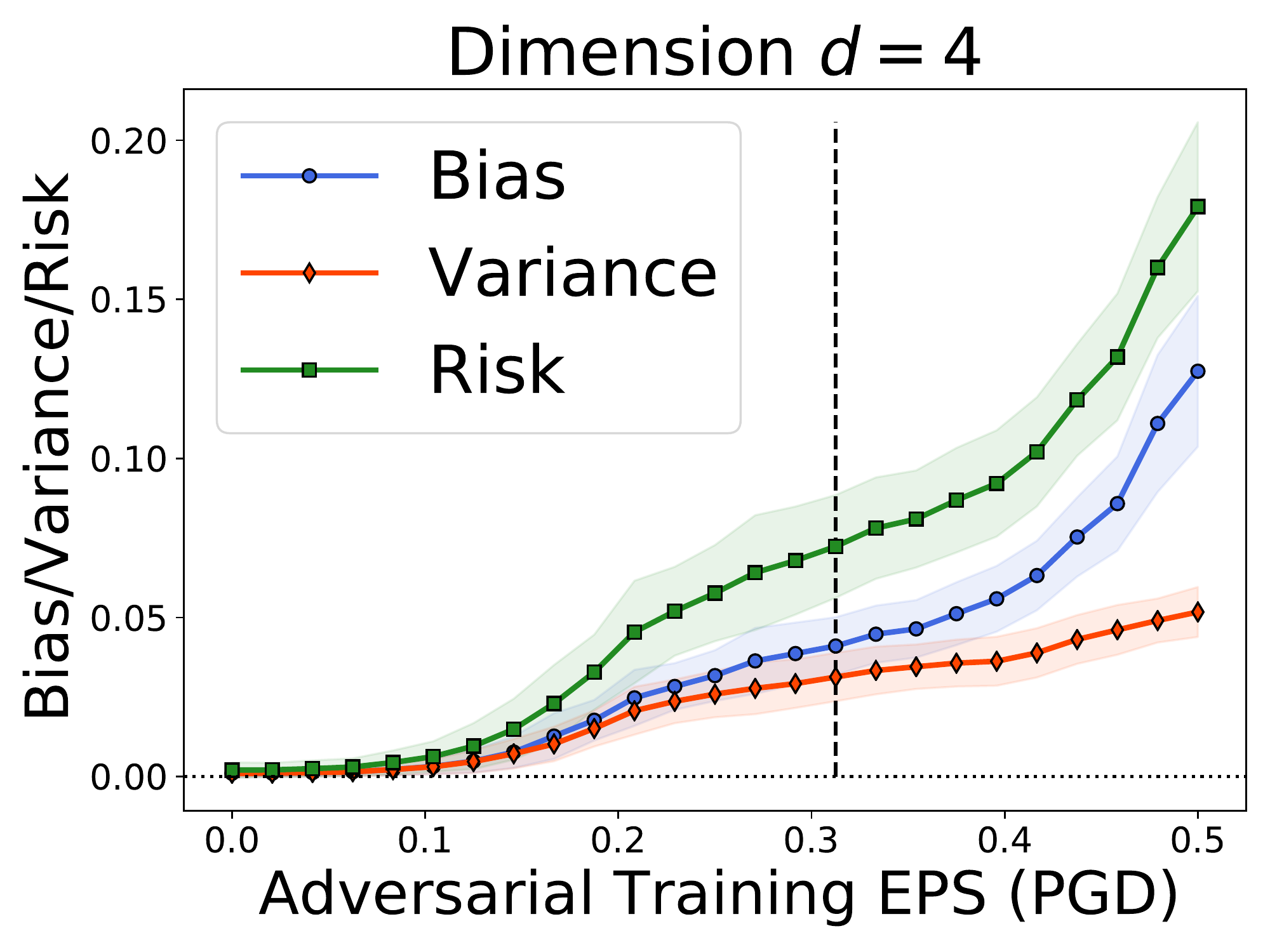}
}
\subfigure[$d=5$]{
\includegraphics[width=.26\textwidth]{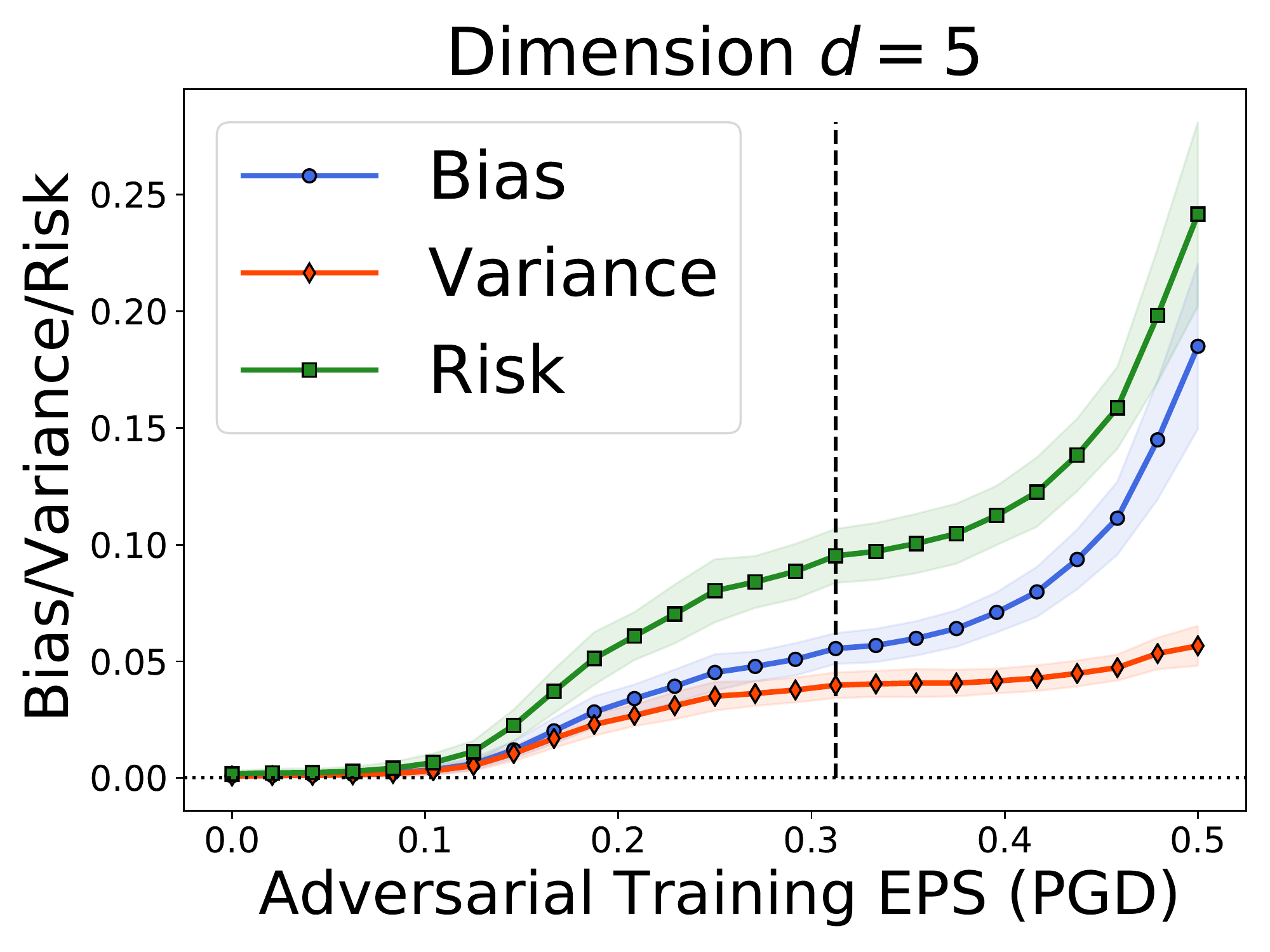}
}
\subfigure[$d=6$]{
\includegraphics[width=.26\textwidth]{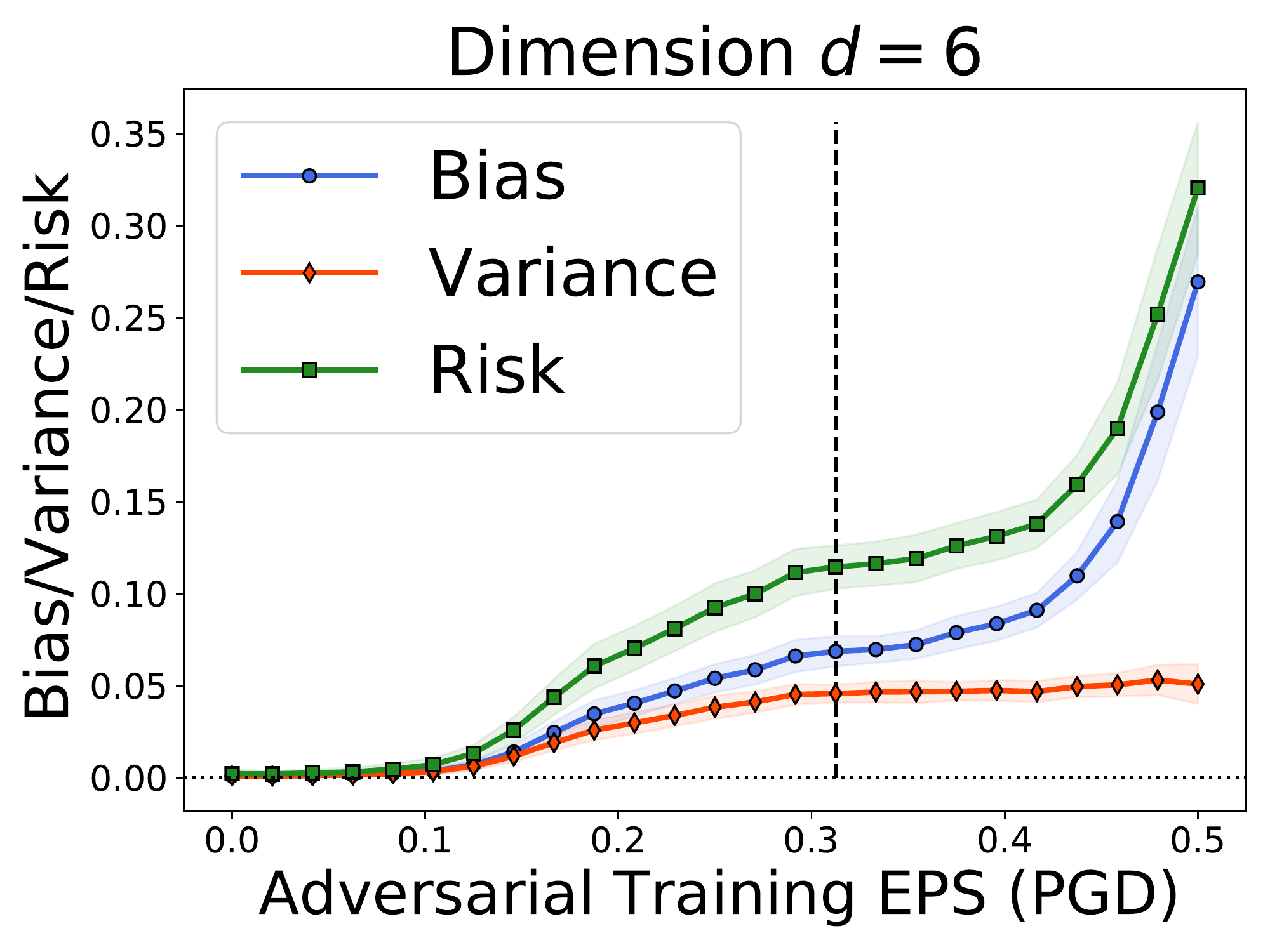}
}
\subfigure[$d=7$]{
\includegraphics[width=.26\textwidth]{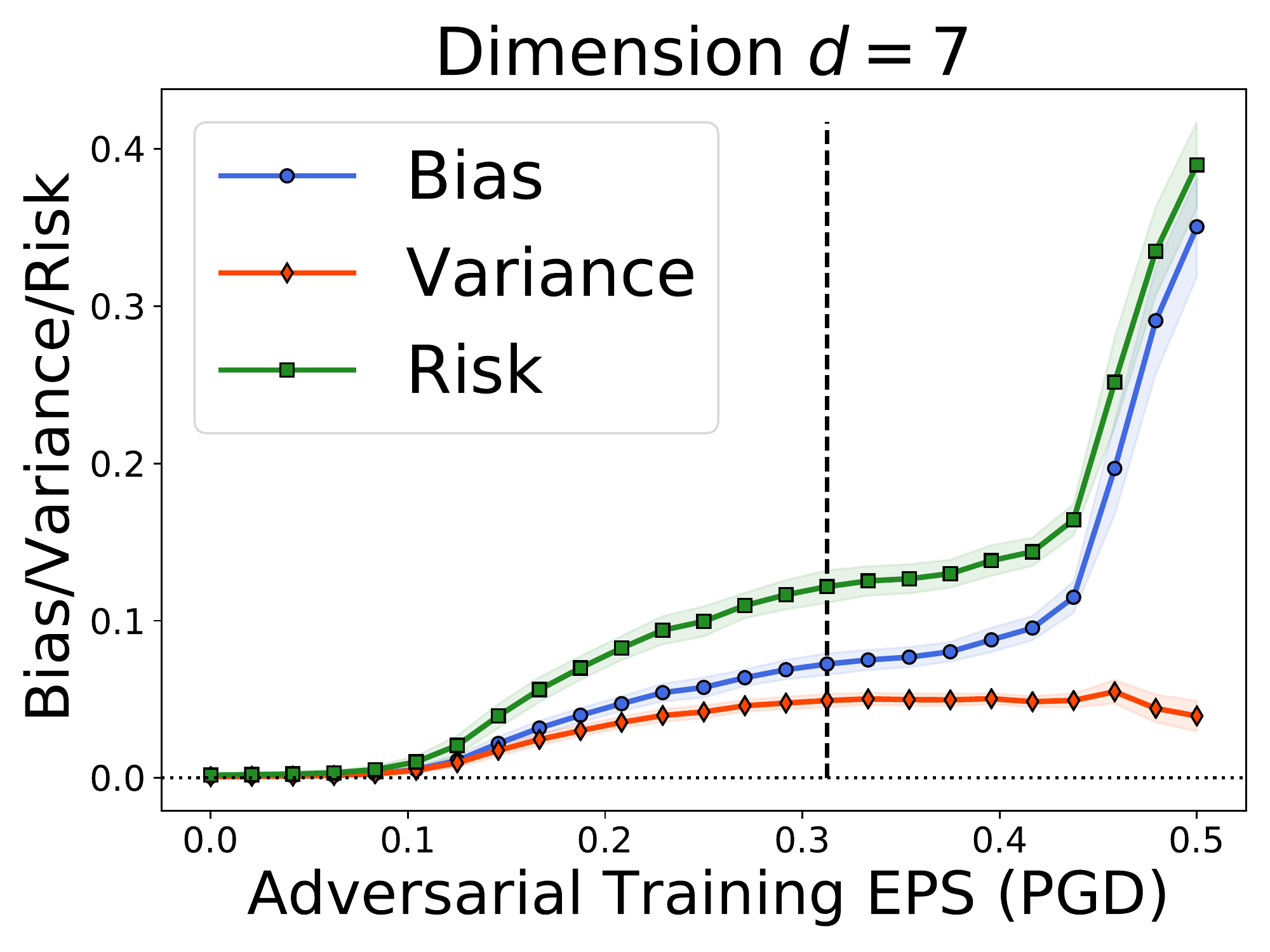}
}
\subfigure[$d=8$]{
\includegraphics[width=.26\textwidth]{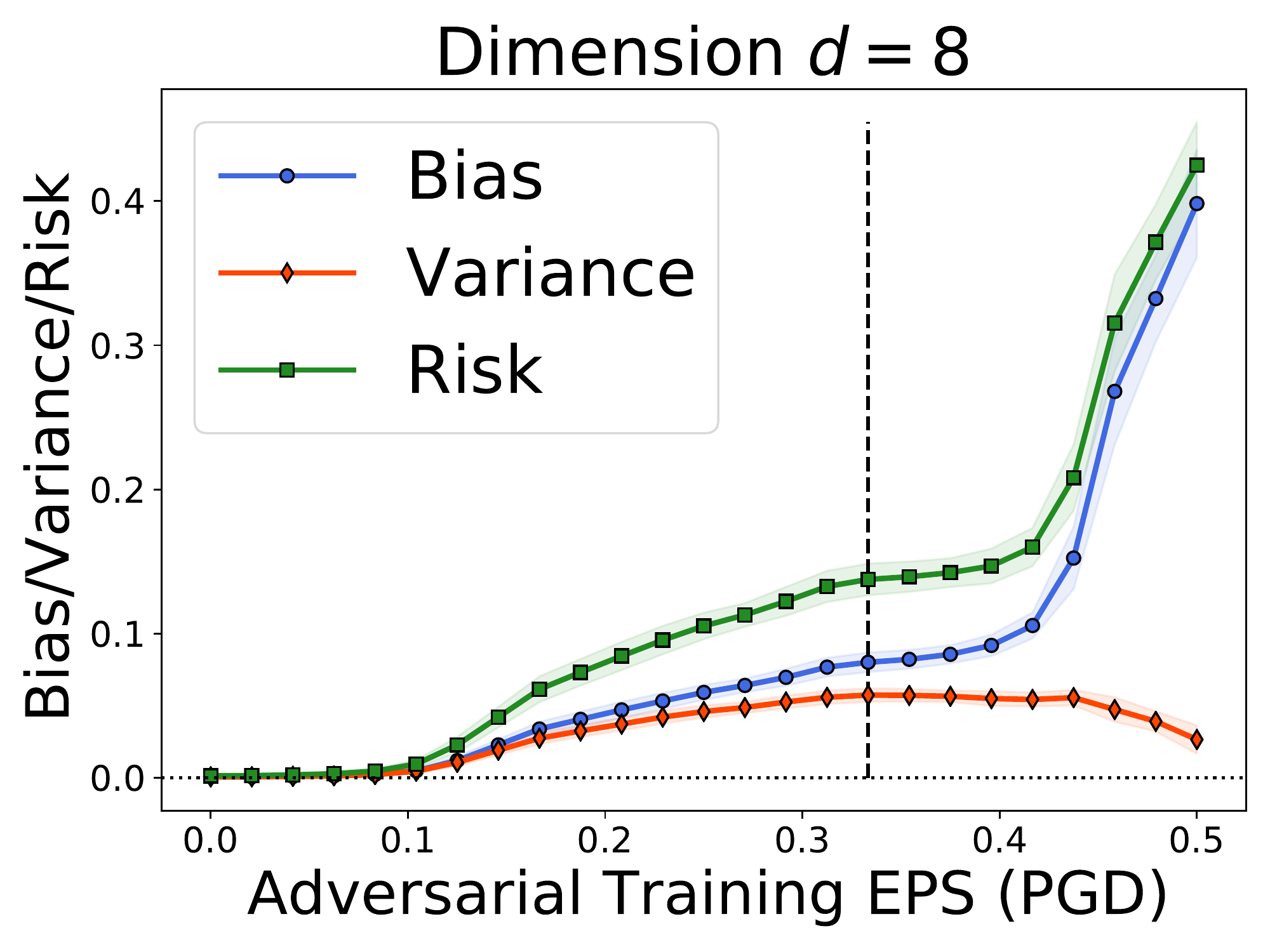}
}
\subfigure[$d=9$]{
\includegraphics[width=.26\textwidth]{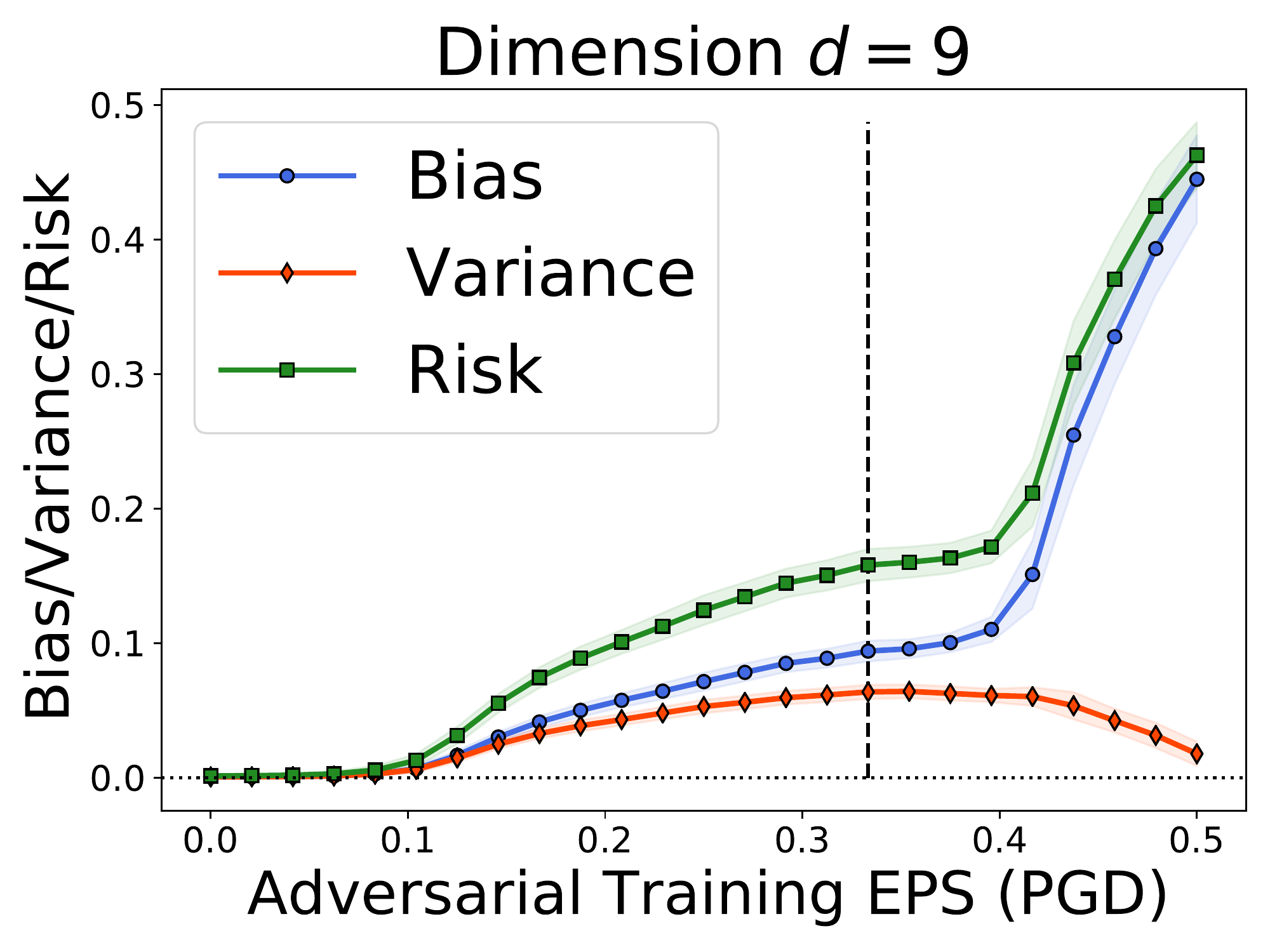}
}
\subfigure[$d=10$]{
\includegraphics[width=.26\textwidth]{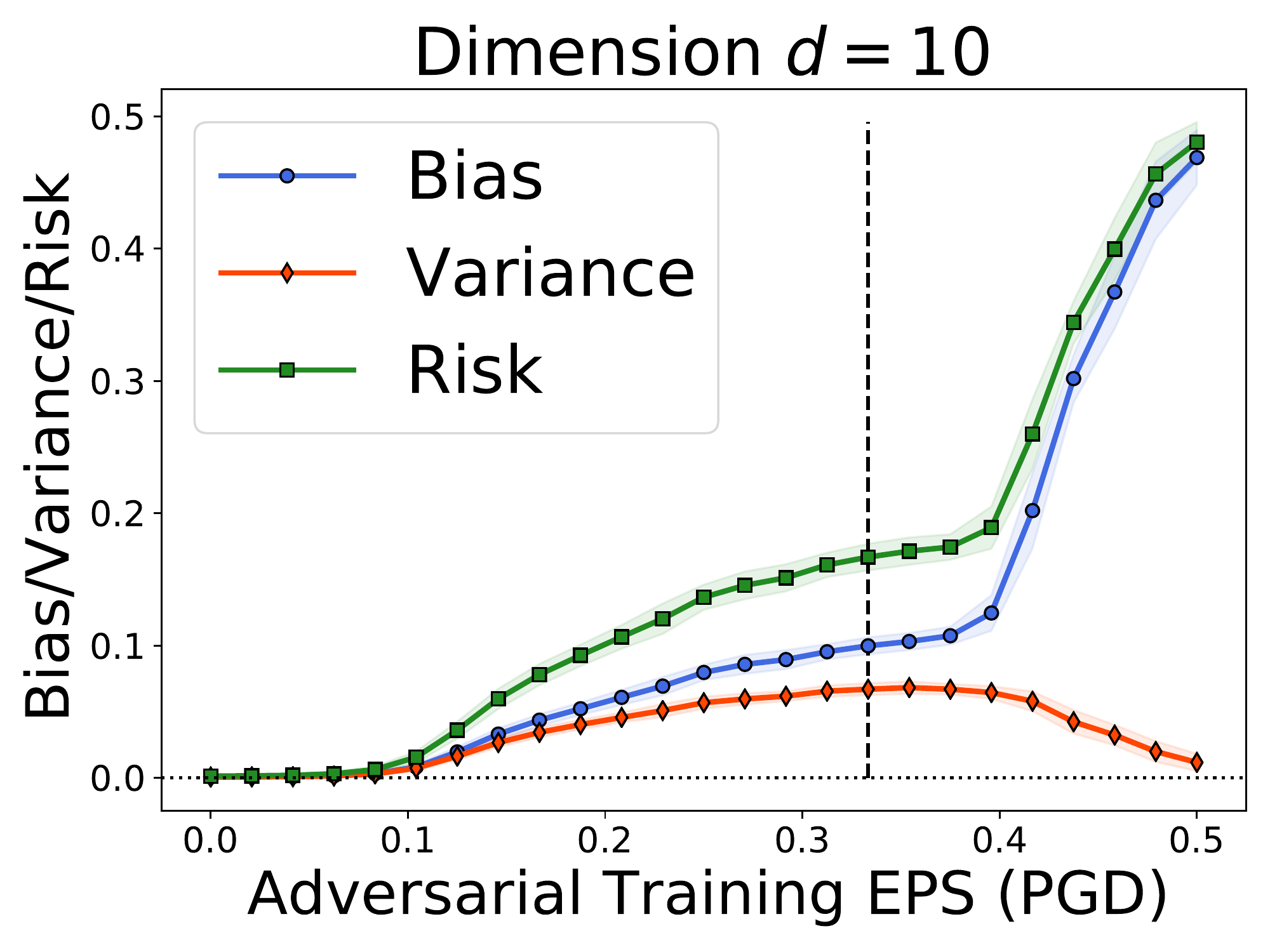}
}
\subfigure[$d=15$]{
\includegraphics[width=.26\textwidth]{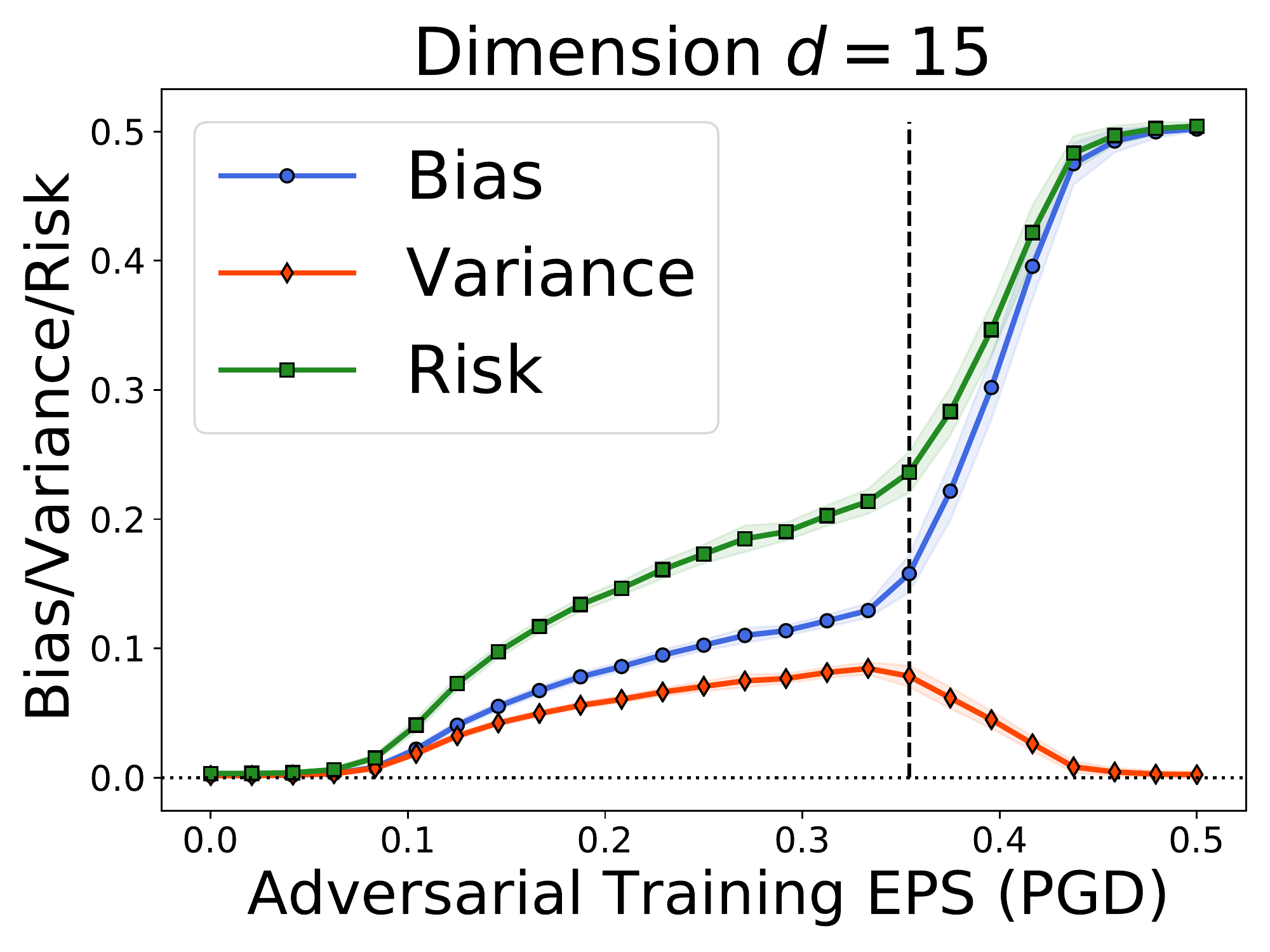}
}
\subfigure[$d=20$]{
\includegraphics[width=.26\textwidth]{tex_files/Figures/bv_dim20_trial30.pdf}
}
\subfigure[$d=25$]{
\includegraphics[width=.26\textwidth]{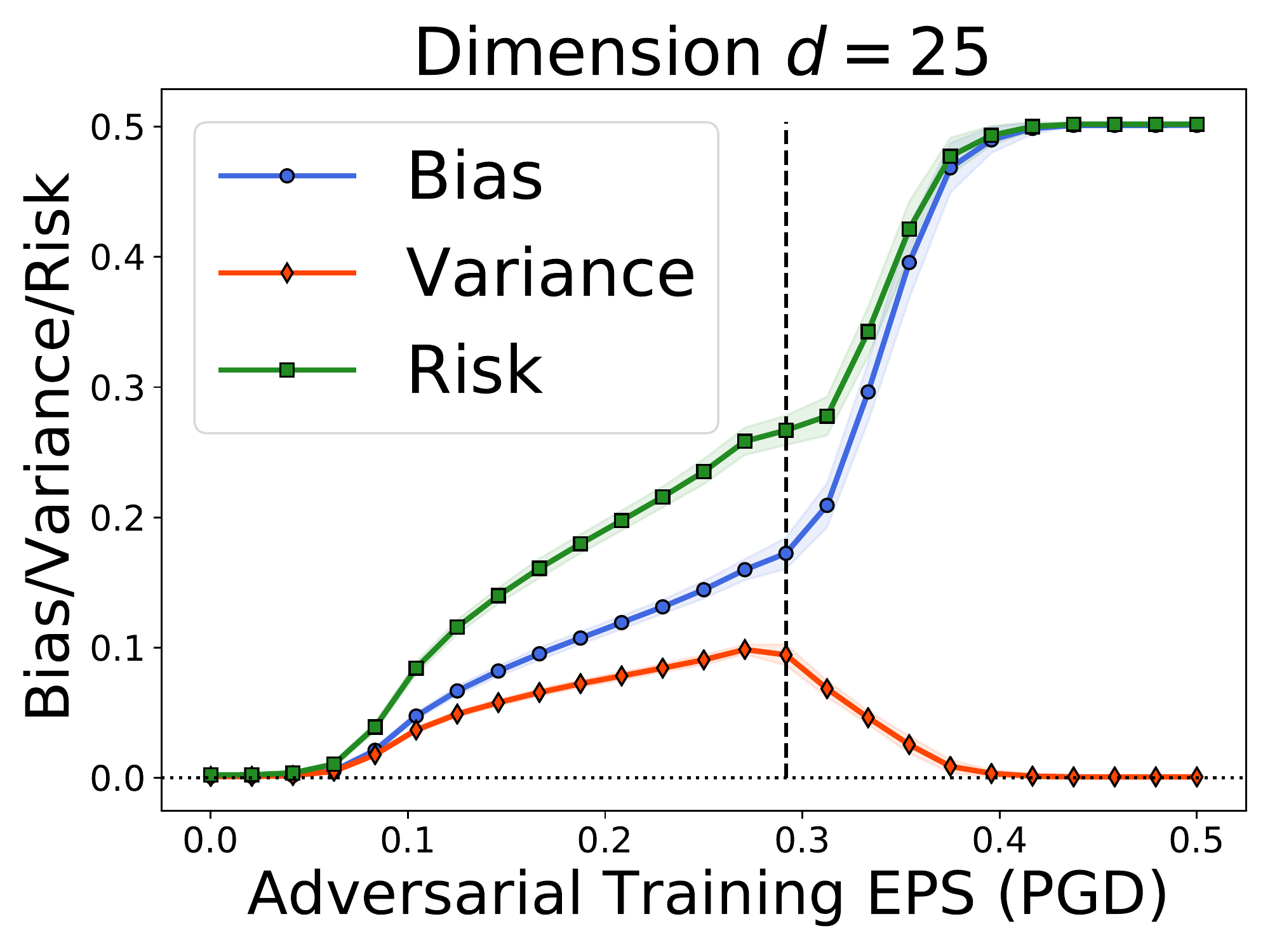}
}
\subfigure[$d=30$]{
\includegraphics[width=.26\textwidth]{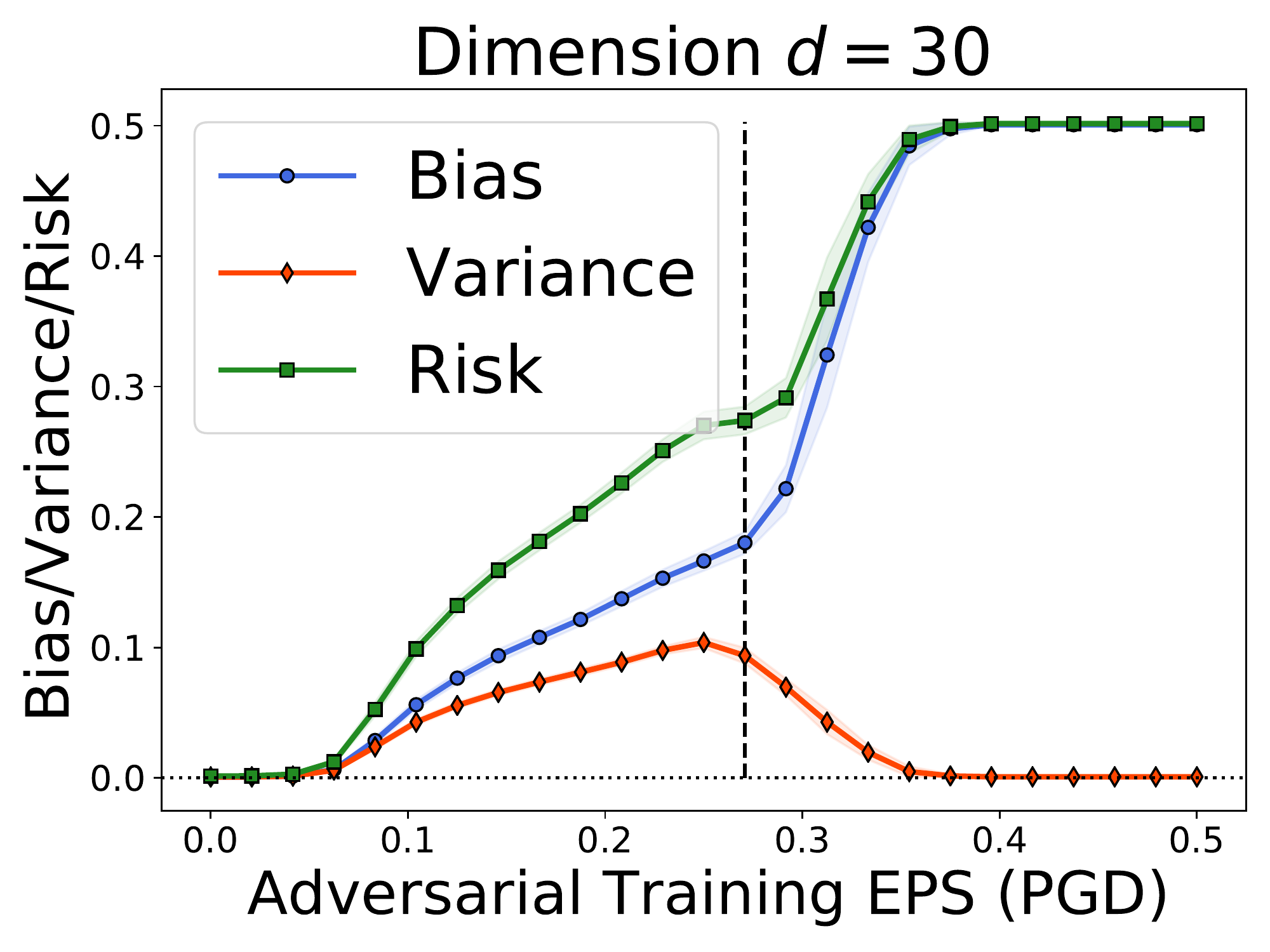}
}
\subfigure[$d=40$]{
\includegraphics[width=.26\textwidth]{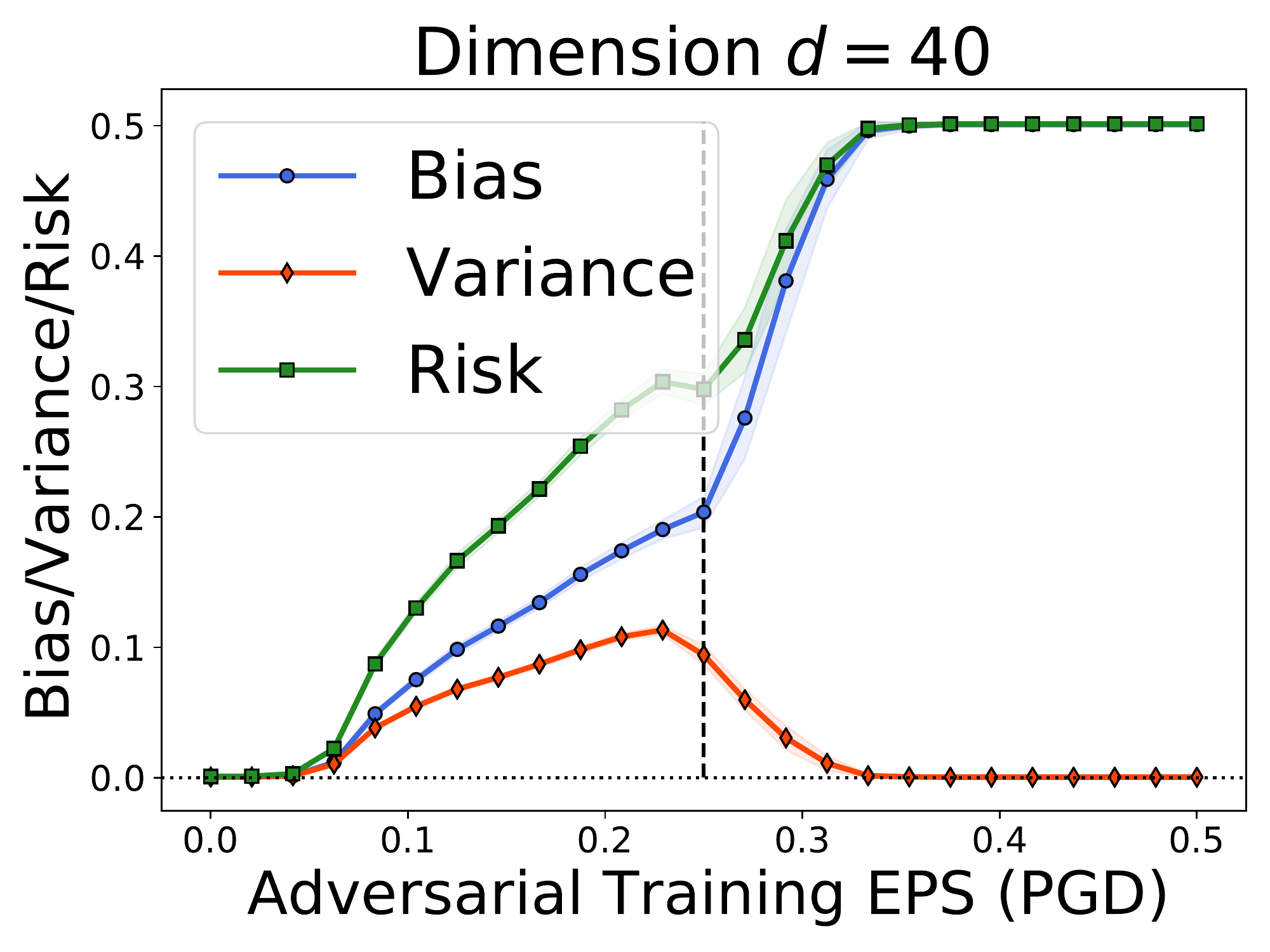}
}
\subfigure[$d=50$]{
\includegraphics[width=.26\textwidth]{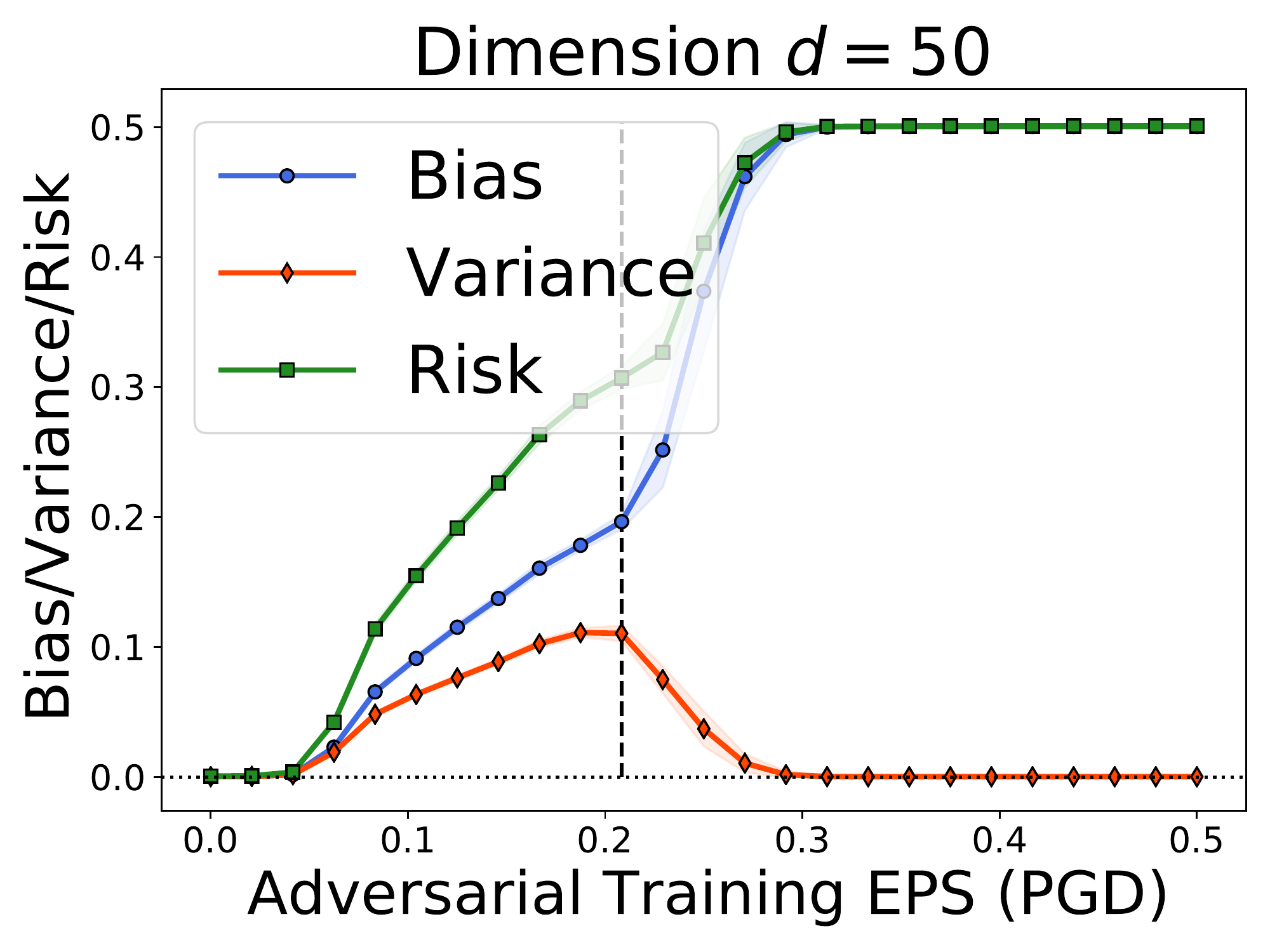}
}
\vskip -0.1in
\caption{Bias-variance behavior of adversarial training on the box dataset with different dimensions ($d \in \{2, \cdots, 50\}$).}
\label{fig:appendix-box-varying-dim}
\end{center}
\end{figure*}

First of all, we present the bias-variance experimental results on 2D box example with different dimension $(2 \leq d \leq 50)$ in Figure~\ref{fig:appendix-box-varying-dim}. Regarding the experimental setup, we only change the dimension $d$, and the number of training samples $n=10\cdot d$ (i.e., same setup as described in Section~\ref{sec:toy_2d}). 

\begin{figure*}[ht]
\vskip -0.1in
\begin{center}
\subfigure[$n=10, d=10, \sigma=0.7$]{
\includegraphics[width=.21\textwidth]{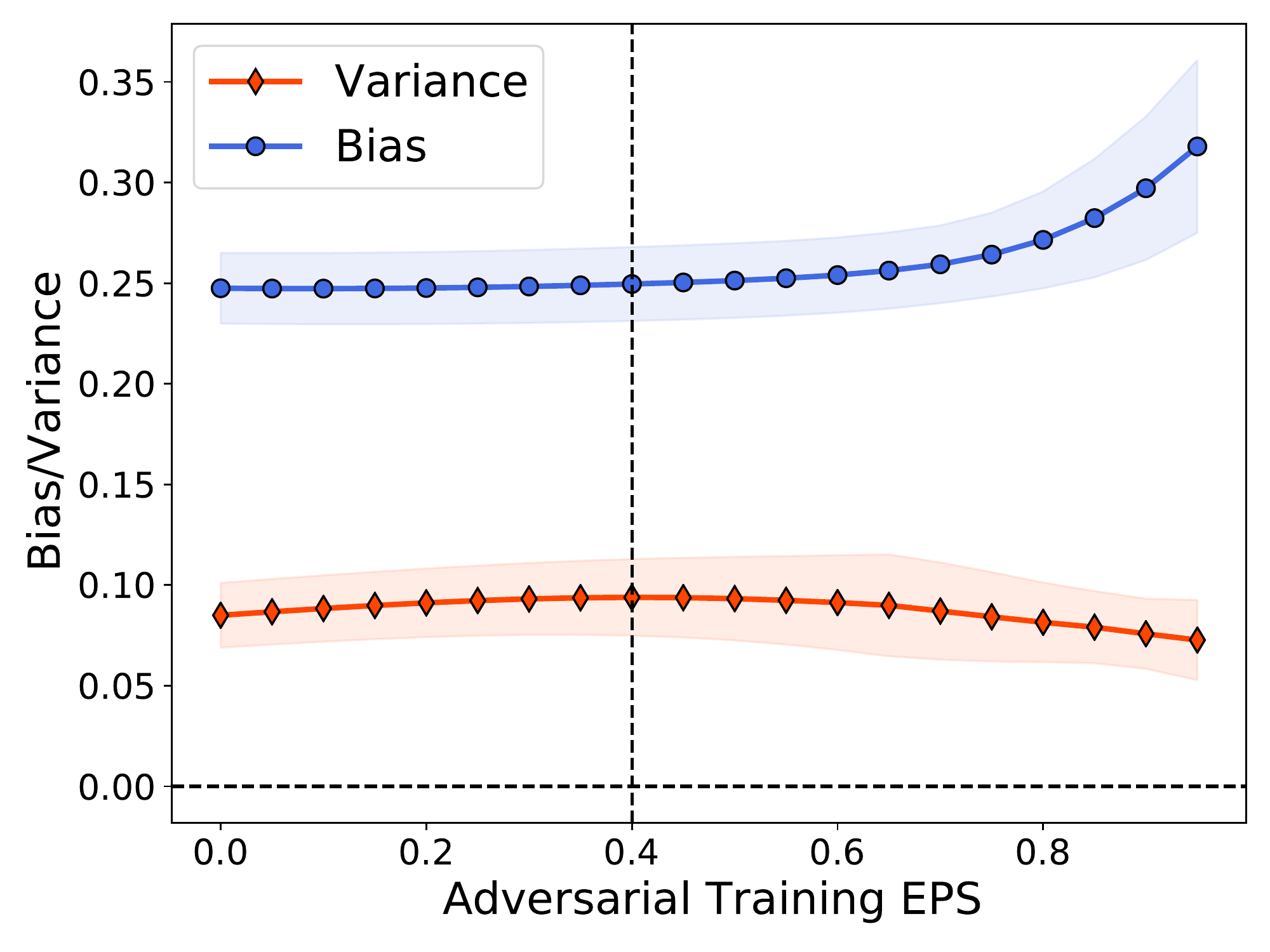}
\includegraphics[width=.235\textwidth]{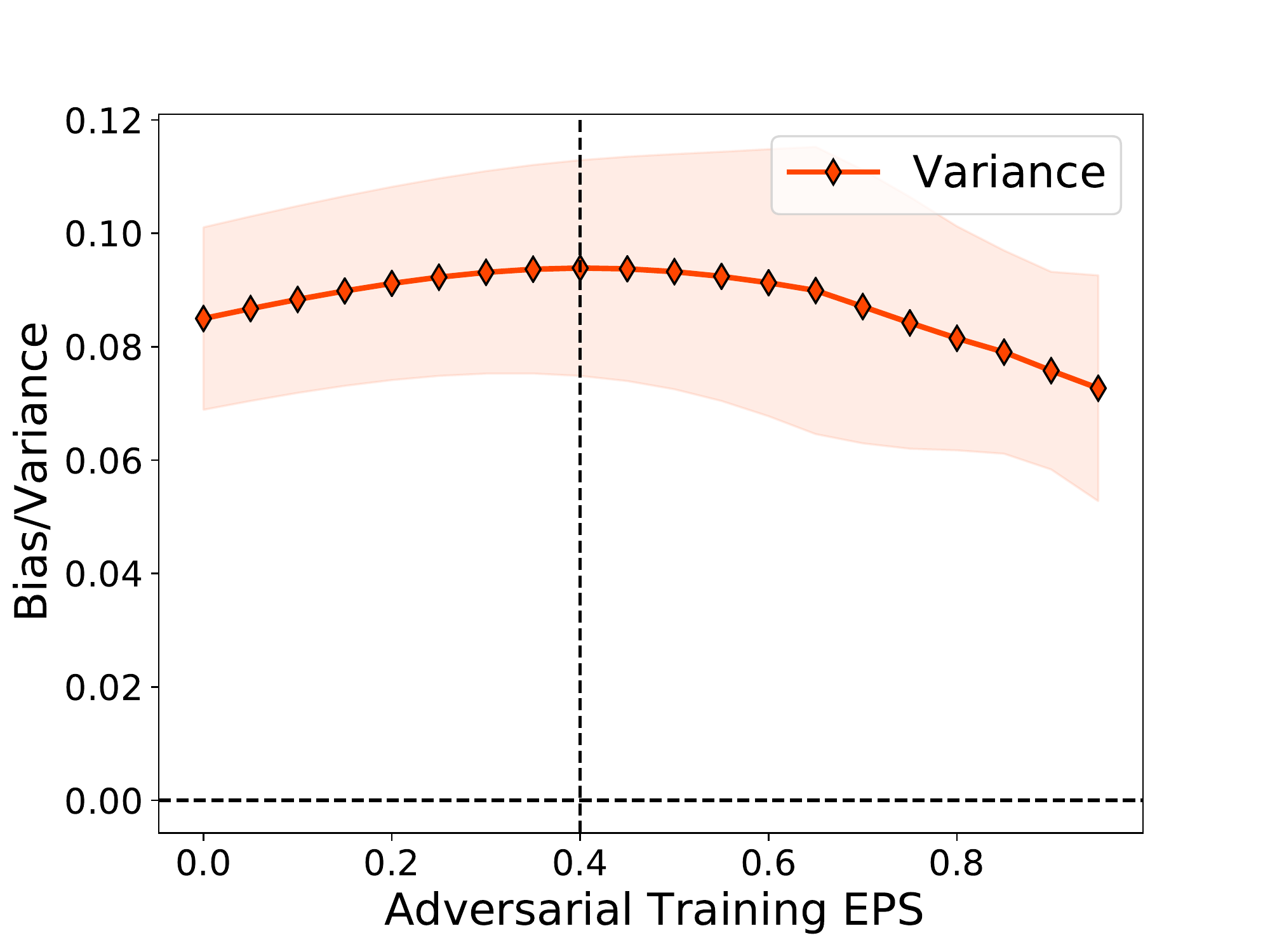}
}
\subfigure[$n=20, d=20, \sigma=0.7$]{
\includegraphics[width=.21\textwidth]{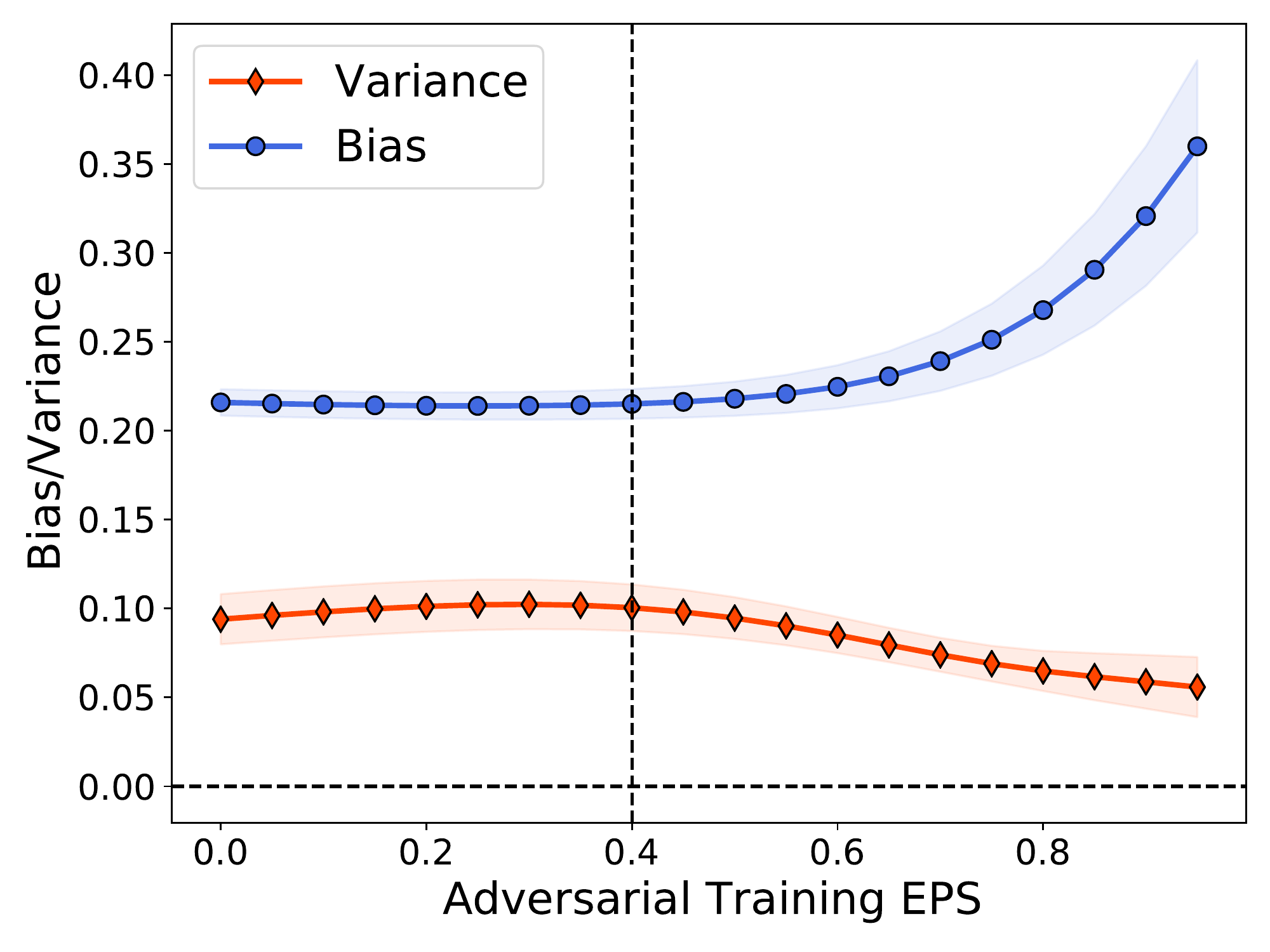}
\includegraphics[width=.235\textwidth]{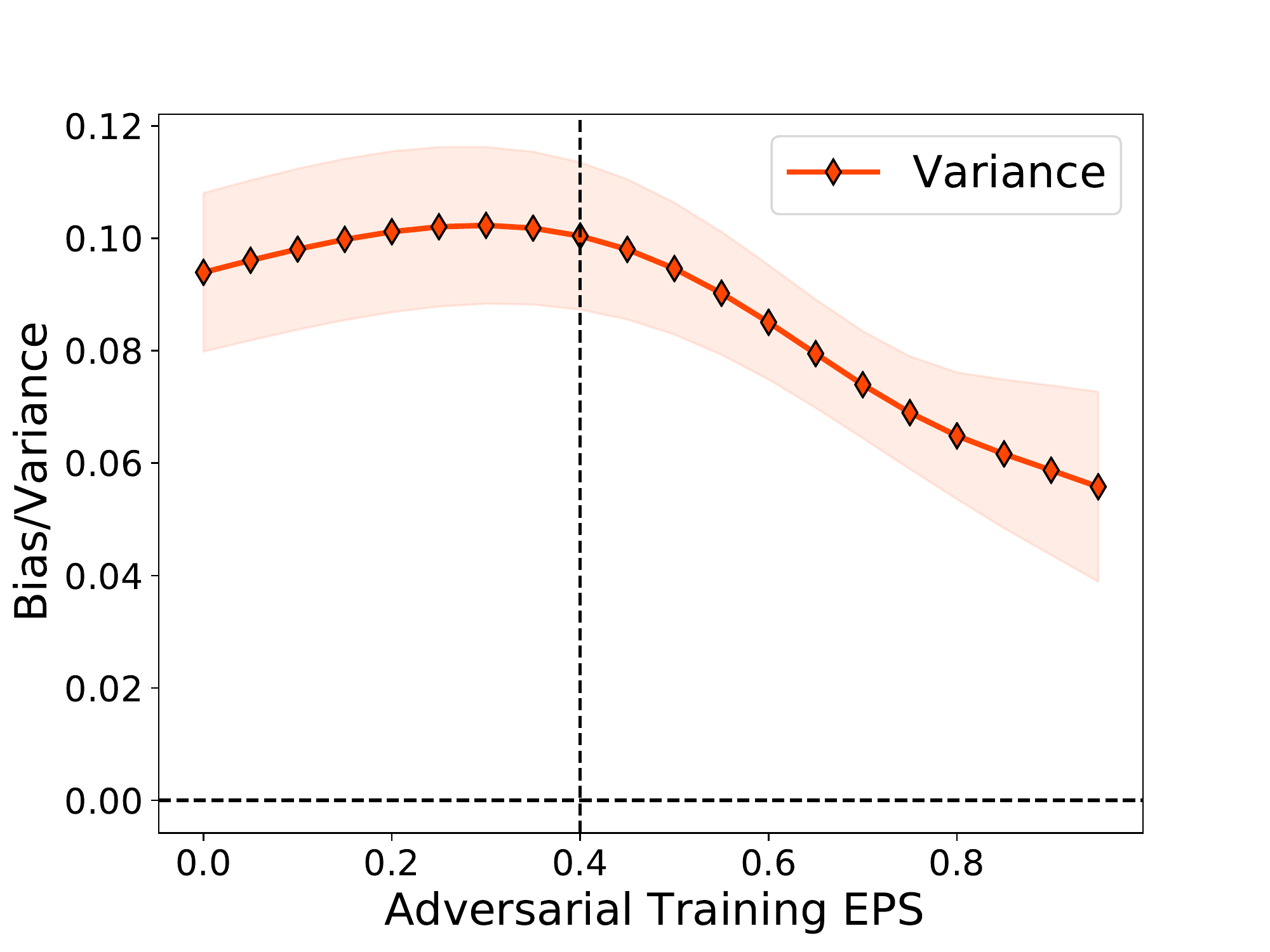}
}
\subfigure[$n=50, d=50, \sigma=0.7$]{
\includegraphics[width=.21\textwidth]{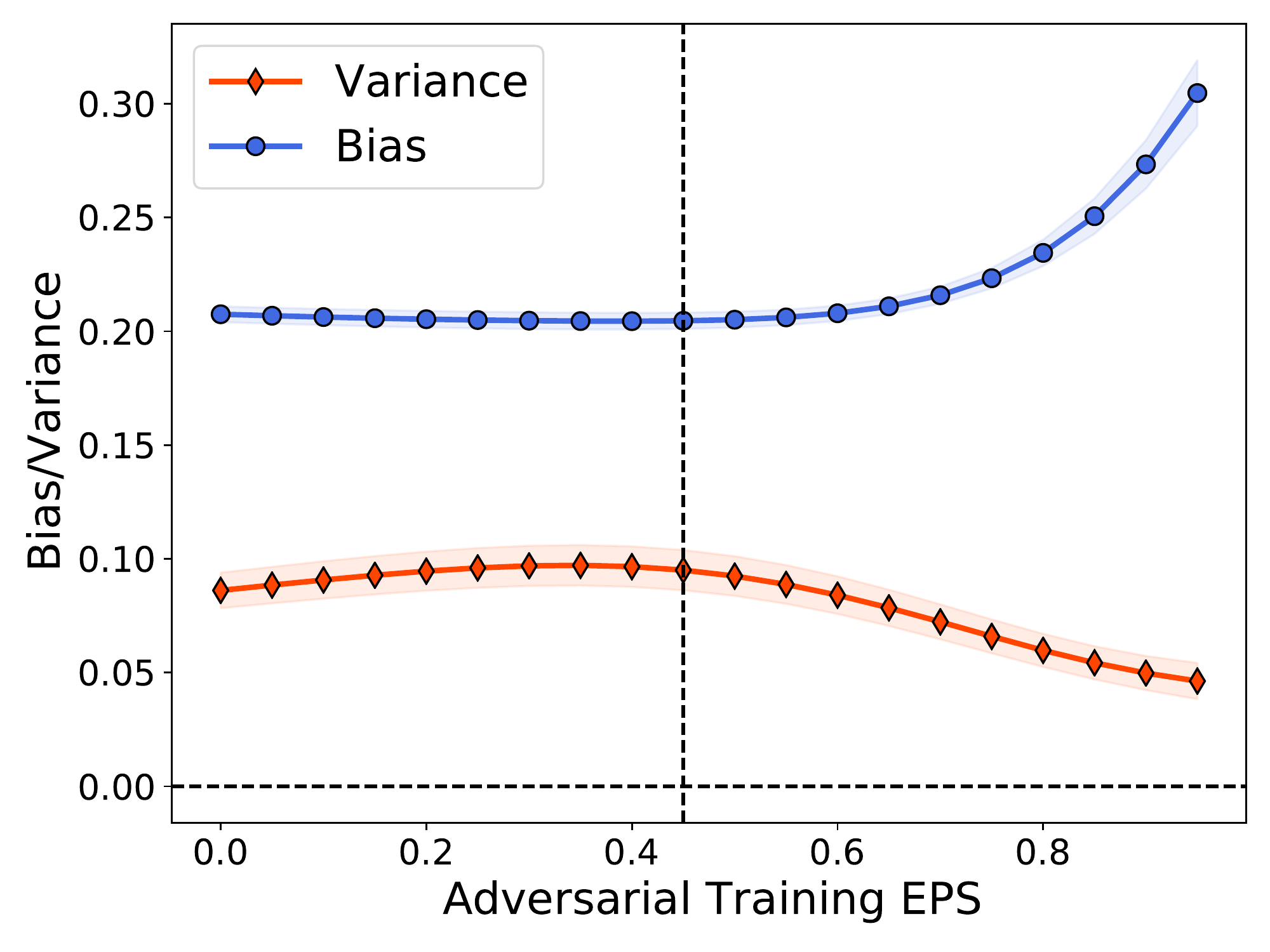}
\includegraphics[width=.235\textwidth]{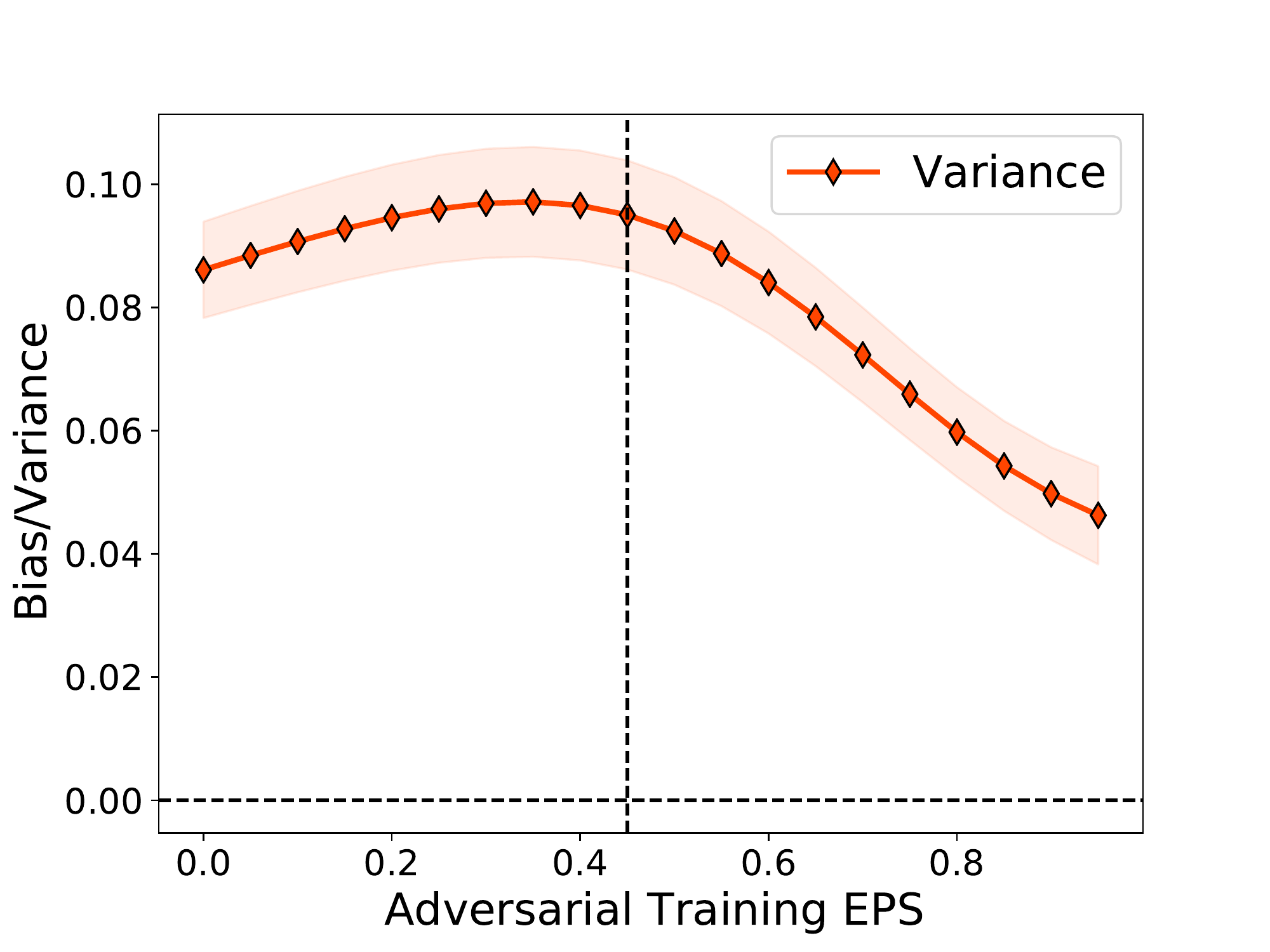}
}
\subfigure[$n=100, d=100, \sigma=0.7$]{
\includegraphics[width=.21\textwidth]{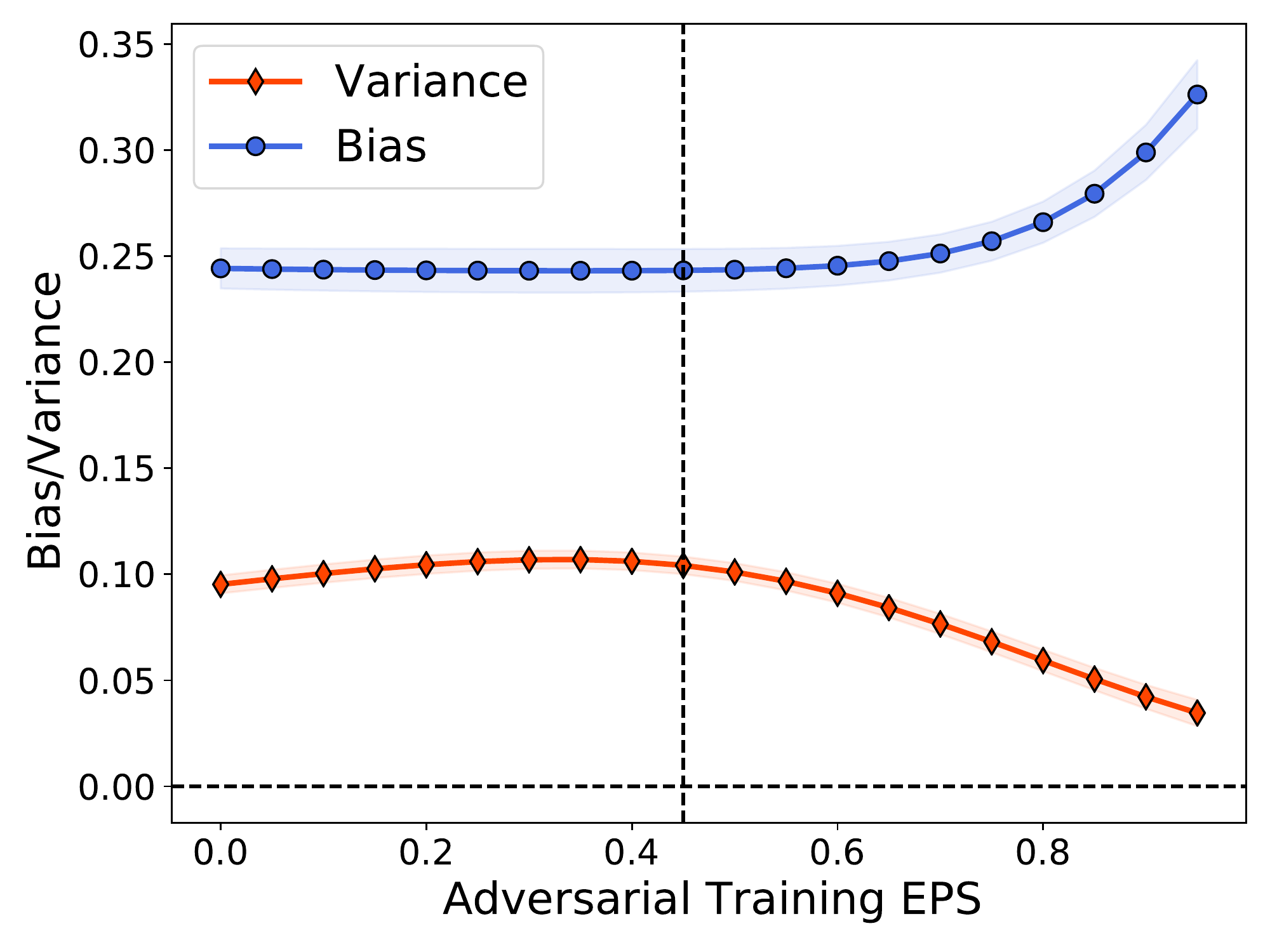}
\includegraphics[width=.235\textwidth]{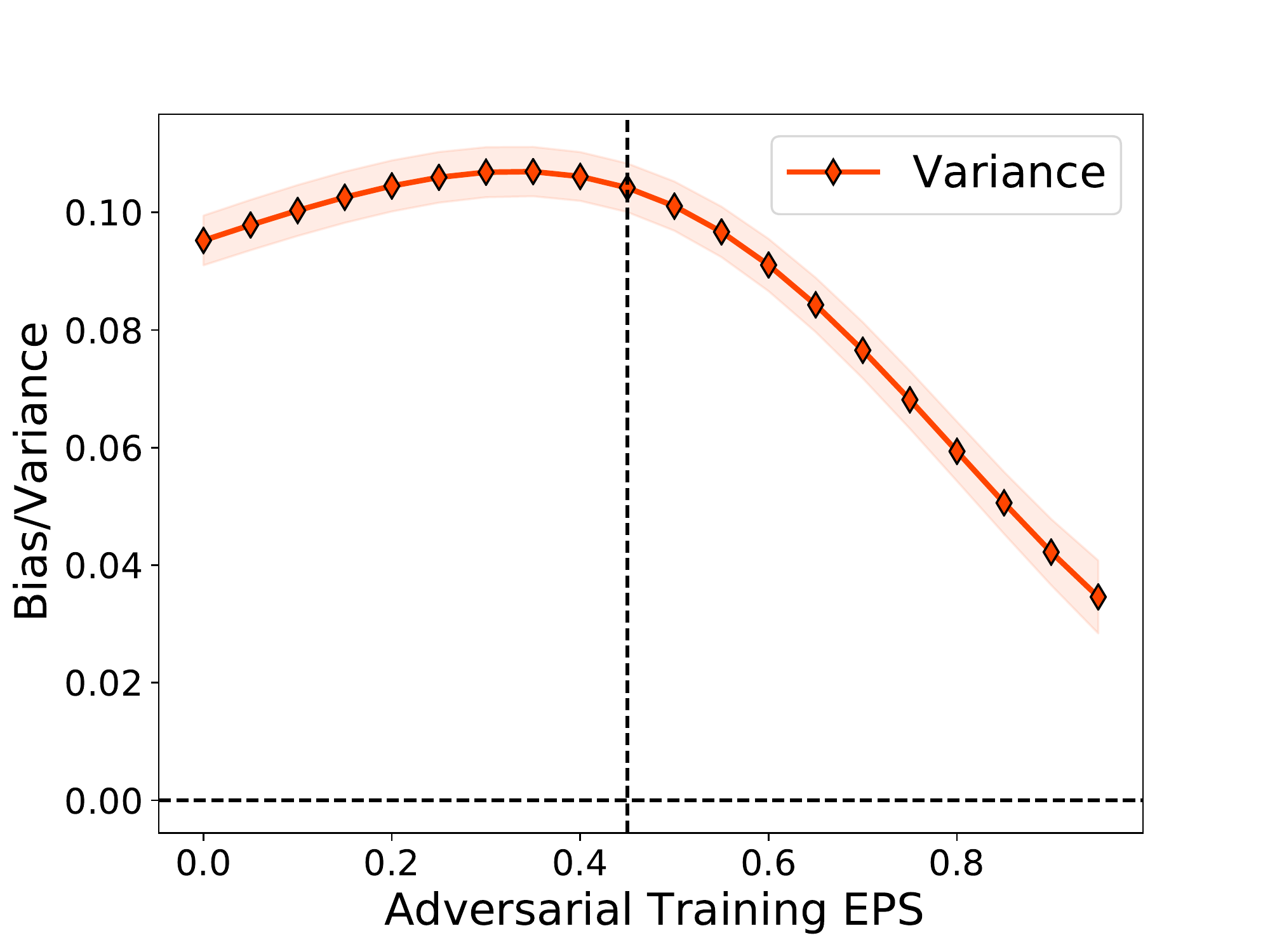}
}
\subfigure[$n=200, d=200, \sigma=0.7$]{
\includegraphics[width=.21\textwidth]{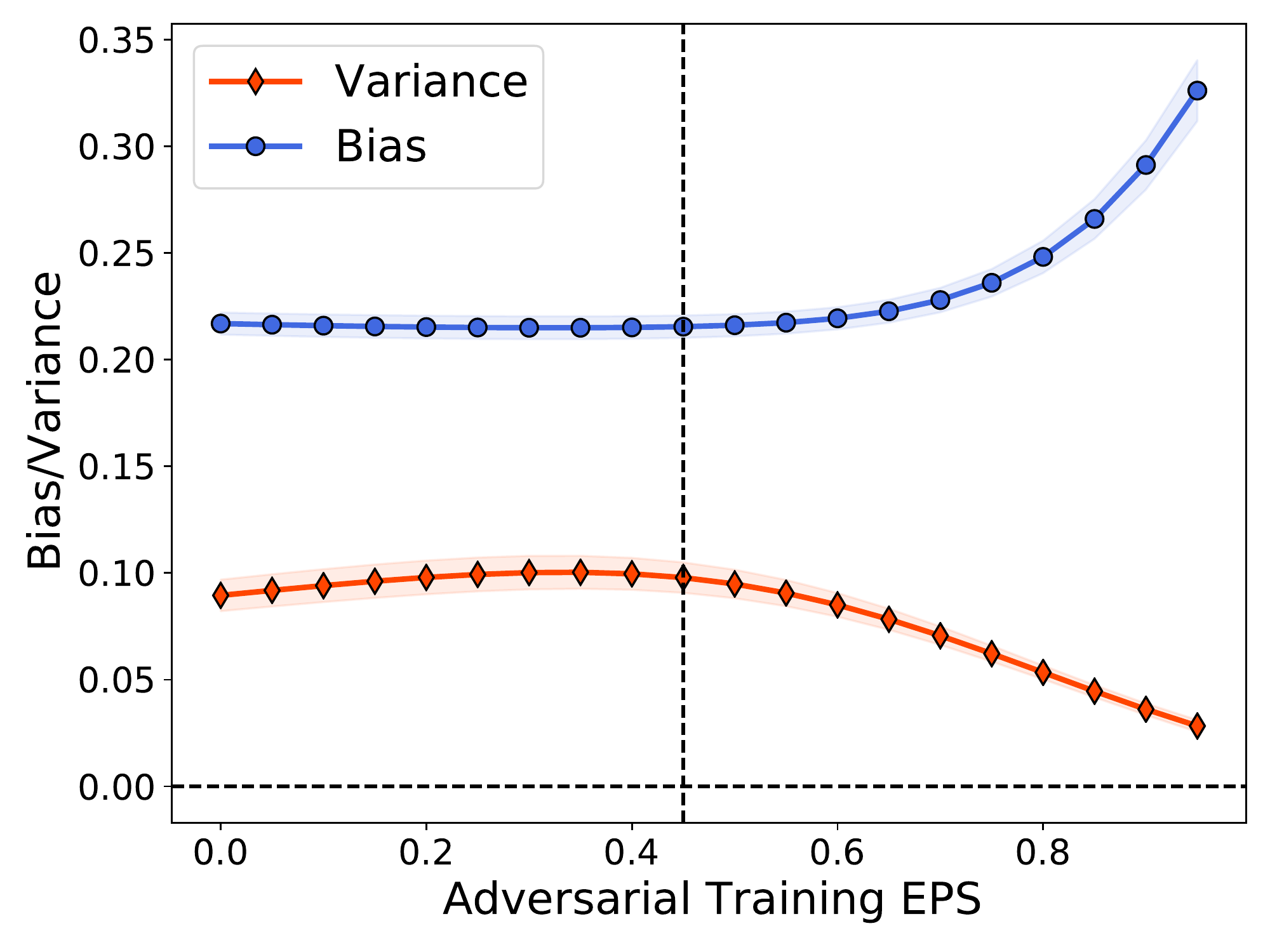}
\includegraphics[width=.235\textwidth]{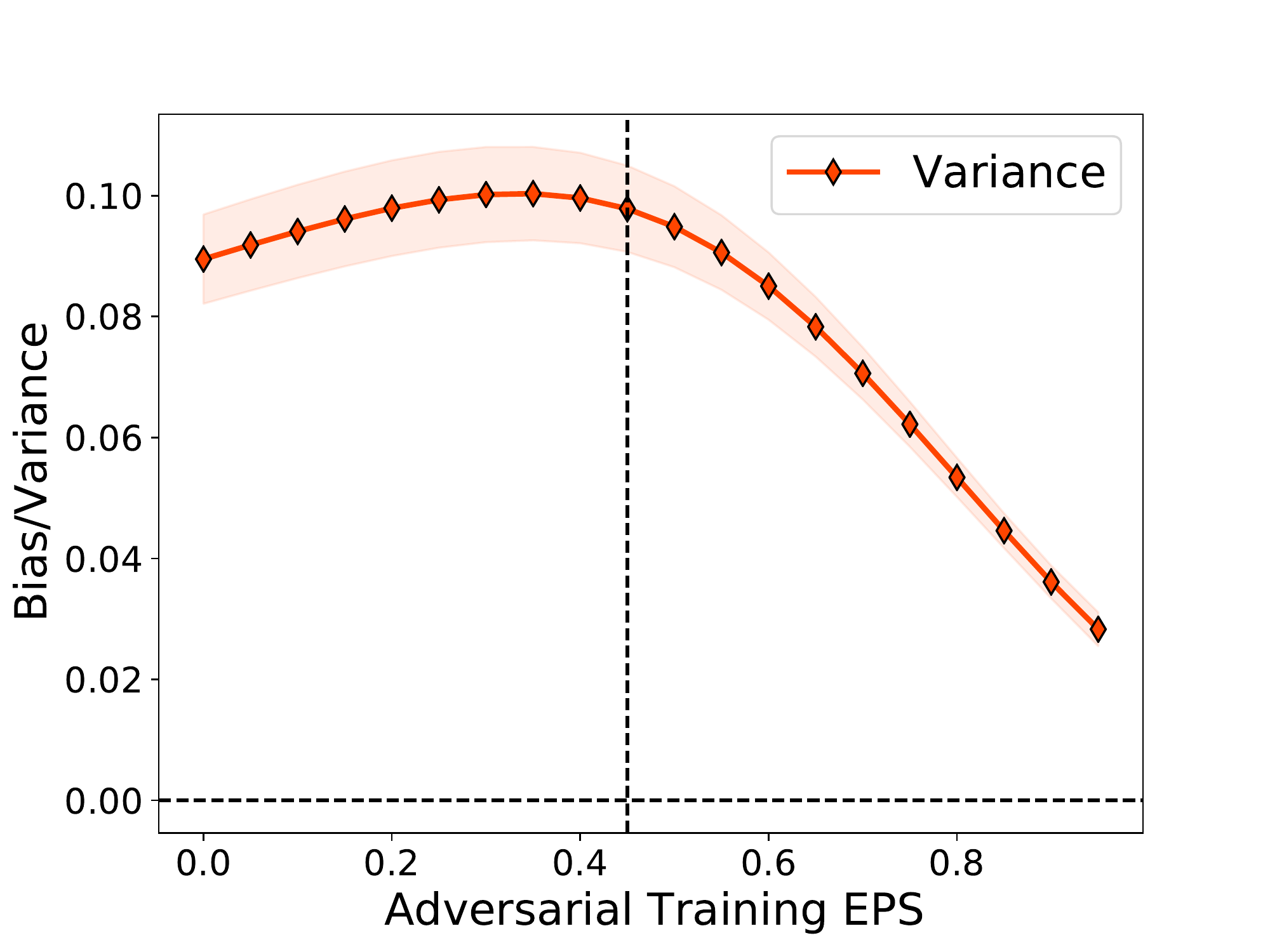}
}
\subfigure[$n=500, d=500, \sigma=0.7$]{
\includegraphics[width=.21\textwidth]{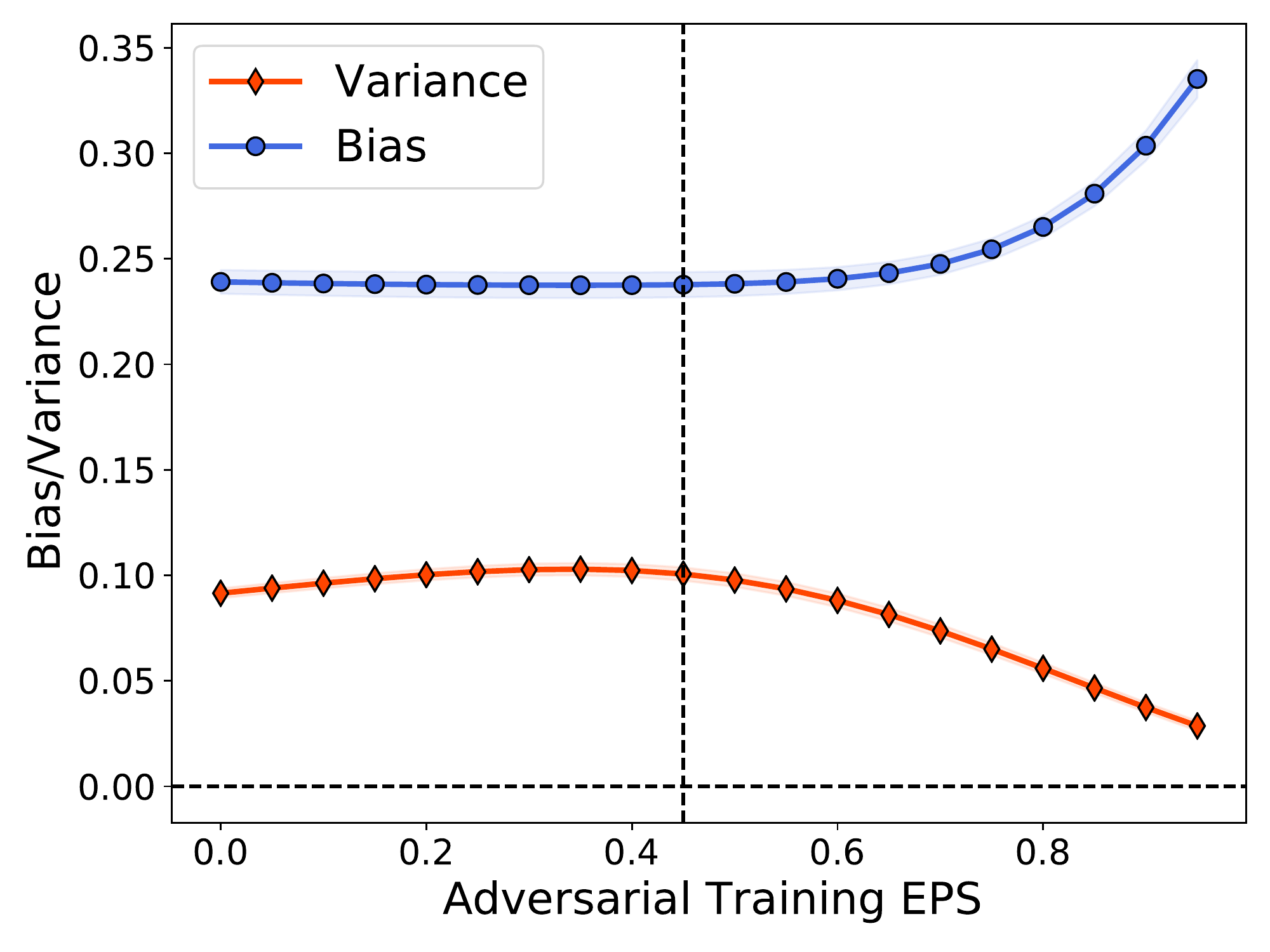}
\includegraphics[width=.235\textwidth]{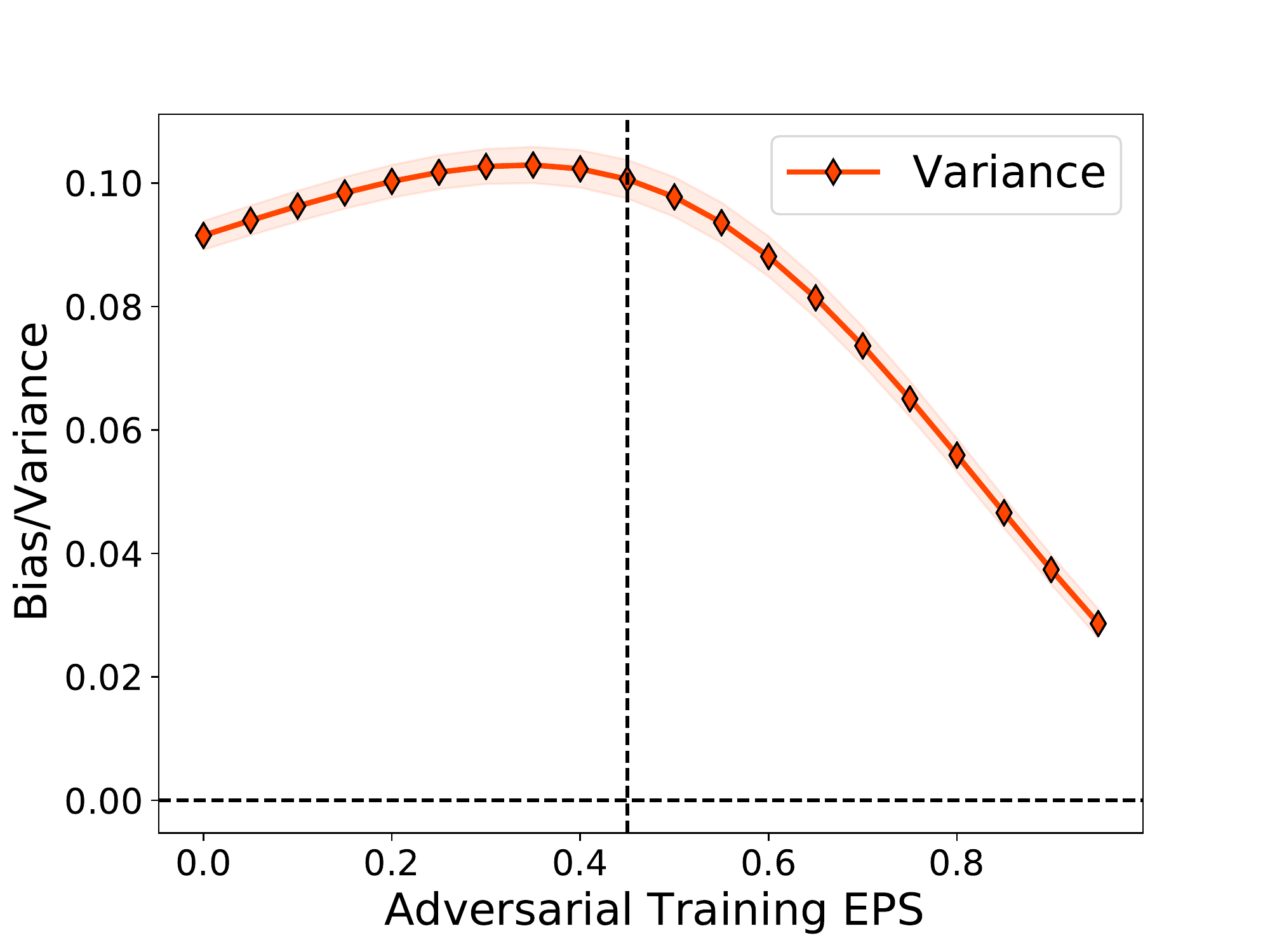}
}
\vskip -0.1in
\caption{More results on the relation between bias/variance/robust interpolation threshold for adversarial logistic regression (with different dimension $d\in\{10, 20, 50, 100, 200, 500\}$). 
We can see that the variance peak corresponds closely to the interpolation threshold.}
\label{fig:appendix-logistic-regression}
\end{center}
\vskip -0.2in
\end{figure*}

Secondly, we present additional experiments on adversarial logistic regression. The precise setup of the experiments is defined in Section \ref{sec:testing_toy}. We consider $d\in\{10, 20, 50, 100, 200, 500\}$ and $n=d$ for the adversarial logistic regression. As shown in Figure~\ref{fig:appendix-logistic-regression}, the variance is unimodal and the variance peak is close to the interpolation threshold. 

\clearpage
\section{Bias and Variance for Logistic Regression}\label{sec:logistic_bv}
\vspace{-0.1in}
In this section, we present the bias-variance decomposition for logistic loss. We also prove that the logistic variance satisfies the usual properties of the variance.
\vspace{-0.1in}
\subsection{Definitions}\label{sec:logistic_bv_def}
\vspace{-0.1in}
Let $\cC =\{\pm 1\}$ be the set of classes. Then, the label $y\in \cC$ defines a probability distribution $\pi$ on $\cC$  as
\[
  \pi_y(c) = \frac{1}{2}(1+cy).
\]
Similarly, given test data $\bx$ and learned parameter $\htn$, the classification rule defines a distribution $\hpi$ on $\cC$ as
\[
  \hpi_{\bx,\htn}(c) = \left(1+e^{-c\<\bx,\htn\>}\right)^{-1}.
\]
Since $\htn$ is random, $\hpi$ is also a random distribution. The logistic loss can equivalently be written in terms of the cross-entropy between $\pi$ and $\hpi$: $H(\pi_y, \hpi_{x,\htn})$
\[
\begin{aligned}
  \sR(\htn) &= \E_{\bx, y}\ell(y\<\htn, \bx\>)\\
  & = \E_{\bx, y} \left[\log\left(1+e^{-y\< \htn, \bx \>}\right)\right],\\ 
  & =  \E_{\bx, y} \left[ -\sum_{c} \pi_y(c) \log \hpi_{x,\htn}(c) \right]\\
  &= \E_{\bx, y} \left[ H(\pi_y, \hpi_{x,\htn}) \right],\\
  & = \E_{\bx, y} \left[ H(\pi_y) + D(\pi_y||\hpi_{x,\htn}) \right],\\
  & = \E_{\bx, y} \left[ D(\pi_y\|\hpi_{x,\htn}) \right],
\end{aligned}
\]
where $D(\cdot \|\cdot)$ denotes the KL divergence. In \cite{YYBV2020}, the following bias-variance decomposition is given for the cross-entropy loss:
\[
\begin{aligned}
  &\E_{\bx, y}\E_{\htn} \left[ D(\pi_y||\hpi_{x,\htn}) \right]\\
  &= \E_{\bx, y}D(\pi_y\|\bpi_x ) + \E_{\bx, y}\E_{\htn} D(\bpi_x  \| \hpi_{x,\htn}),
\end{aligned}
\]
where $\bpi_\bx $ is the average of log probability:
\[
  \bpi_\bx (c) = \frac{e^{\E_{\htn} \log \hpi_{\bx,\htn}(c)}}{Z_\bx } = \frac{\exp\left[-\E_{\htn} \ell(c\<\htn, \bx\>)\right]}{Z_\bx },
\]
where 
\[Z_\bx = \exp\left[-\E_{\htn} \ell(\<\htn, \bx\>)\right]  + \exp\left[-\E_{\htn} \ell(-\<\htn, \bx\>)\right]
\]
is the normalization factor. Then, the bias term is
\[
\begin{aligned}
  \textsf{Bias} &= \E_{\bx, y} \left[ \sum_{c} \pi_y(c) \log \frac{\pi_y(c)}{\bpi_x (c)} \right],\\
  &= \E_{\bx, y} \left[\log Z_\bx + \E_{\htn} \ell (y\<\htn, \bx\>)\right].
\end{aligned}
\]
Notice that 
\begin{equation}\label{eqn:log_bias}
  \textsf{Bias} = \E_{\bx, y}\log Z_\bx + \E_{\htn} \sR(\htn).
\end{equation}
Thus, the above calculation also identifies 
\begin{equation}\label{eqn:log_variance}
  \textsf{Variance} = -\E_{\bx}\log {Z_\bx }.
\end{equation}

\subsection{Properties of Logistic Variance}
The proposition below shows that the variance for logistic regression has the usual non-negativity property of the variance; as expected. 
\begin{proposition}[Non-Negativity of Logistic Variance] The variance for logistic regression defined in \eqref{eqn:log_variance} is non-negative, and equals 0 when the learned parameters $\hat{\btheta}_n$ is non-random.
\label{prop:logistic_var} 
\begin{proof}
First, we evaluate the formula for the limiting case when the sample size diverges to infinity. This is used later to prove non-negativity. Suppose that as $n\to\infty$, the estimate $\htn$ concentrates around some value $\ts$ (which of course can depend on the objective function, and in particular on $\ep$ for adversarial robustness).  Then, $Z_\bx$ is approximated by  
\begin{align*}
Z_\bx^* & = \exp\left[-\ell(\<\ts, \bx\>)\right]  + \exp\left[- \ell(-\<\ts, \bx\>)\right]\\
&= \exp\left[-\log\left(1+e^{-\<\ts, \bx\>}\right)\right]  
+ \exp\left[- \log\left(1+e^{\<\ts, \bx\>}\right)\right]\\
&= \left(1+e^{-\<\ts, \bx\>}\right)^{-1}  
+ \left(1+e^{\<\ts, \bx\>}\right)^{-1} = 1.
\end{align*}
Therefore, we have $\textsf{Variance} \to 0$ asymptotically. Since the log-sum-exp function $L(z_1,\ldots,z_m) = \log(\sum \exp z_i)$ is convex, it follows from Jensen's inequality that 
\begin{align*}
\log Z_\bx & =\log(\exp\left[-\E_{\htn} \ell(\<\htn, \bx\>)\right]  + \exp\left[-\E_{\htn} \ell(-\<\htn, \bx\>)\right])\\
& =L(-\E_{\htn} \ell(\<\htn, \bx\>),-\E_{\htn} \ell(-\<\htn, \bx\>))\\
& \ge \E_{\htn} L(- \ell(\<\htn, \bx\>),-\ell(-\<\htn, \bx\>))\\
& = \E_{\htn} 0 = 0.
\end{align*}
Hence, $Z_\bx \ge 1$ and the above notion of variance is non-negative, $\textsf{Variance}\geq 0$.
\end{proof}
\end{proposition}

\end{document}